\newcommand{\comment}[1]{}
\comment{

Page limit: 8

제출 11.15 5시 기준 추가된 내용

intro 에서 teaser figure 언급

3.4
Obj-Part sim
특정 obj-part 와 특정 obj-part 와의 similarity 라는 점 언급

2024.12.12
TODO: 그림 수정 및 arXiv (?)

2025.01.17 02:45 아카이브 제출 v1

2025.06.10 00:0 아카이브 제출 v2

}

\documentclass[10pt,twocolumn,letterpaper]{article}

\usepackage[pagenumbers]{cvpr} 

\newcommand{\TODO}[1]{\noindent\textbf{\color{red}[TODO: #1]}}

\usepackage{threeparttable}        
\usepackage{wrapfig}               
\usepackage{booktabs}
\usepackage{xcolor}                
\usepackage{bbding}                
\usepackage{amsmath}               
\usepackage{graphicx}              
\usepackage{accents}               
\usepackage{multirow}              
\usepackage{mathtools}             
\usepackage{footnote}
\usepackage{tablefootnote}
\usepackage{tabularx}              
\usepackage{subfiles}
\usepackage{colortbl}              
\usepackage{amssymb}               
\usepackage{comment}               
\usepackage{titletoc}              
\usepackage{tocbibind}             
\usepackage{appendix}              
\usepackage{url}
\usepackage{cuted}                 
\usepackage{stfloats}              
\usepackage{capt-of}               
\usepackage{subcaption}            

\usepackage{tikz}
\usetikzlibrary{calc}

\usepackage{lipsum}                
\usepackage{kotex}                 
\usepackage{pifont}
\usepackage{makecell}

\definecolor{darkergreen}{RGB}{21, 152, 56}
\definecolor{red2}{RGB}{252, 54, 65}

\definecolor{tealgreen}{rgb}{0.0, 0.51, 0.5}
\newcommand{\gainp}[1]{\textcolor{tealgreen}{$^{\texttt{(#1)}}$}}

\newcommand{\MODELNAME}[1]{{PartCATSeg}}

\definecolor{cvprblue}{rgb}{0.21,0.49,0.74}
\usepackage[pagebackref,breaklinks,colorlinks,allcolors=cvprblue]{hyperref}

\def\paperID{1327} 

\def\confName{CVPR}
\def\confYear{2025}

\title{Fine-Grained Image-Text Correspondence with Cost Aggregation\\for Open-Vocabulary Part Segmentation}

\author{
    Jiho~Choi\textsuperscript{\textrm{1}}\thanks{Equal contribution} ,~
    Seonho~Lee\textsuperscript{\textrm{1}}\footnotemark[1] ,~
    Minhyun~Lee\textsuperscript{\textrm{2}},~
    Seungho~Lee\textsuperscript{\textrm{2}},~
    Hyunjung~Shim\textsuperscript{\textrm{1}}\thanks{Corresponding author} \\
    \textsuperscript{\textrm{1}}KAIST, Republic of Korea \\
    \textsuperscript{\textrm{2}}Samsung Electronics, Republic of Korea \\
    {\tt\small {\{jihochoi, glanceyes, kateshim\}@kaist.ac.kr, \{mh315.lee, sh622.lee\}@samsung.com}} \\
}

\begin{document}

\maketitle

\begin{strip}
    \begin{minipage}{\textwidth}
        \centering
        \vspace{-5em}
        \includegraphics[width=0.95\linewidth]{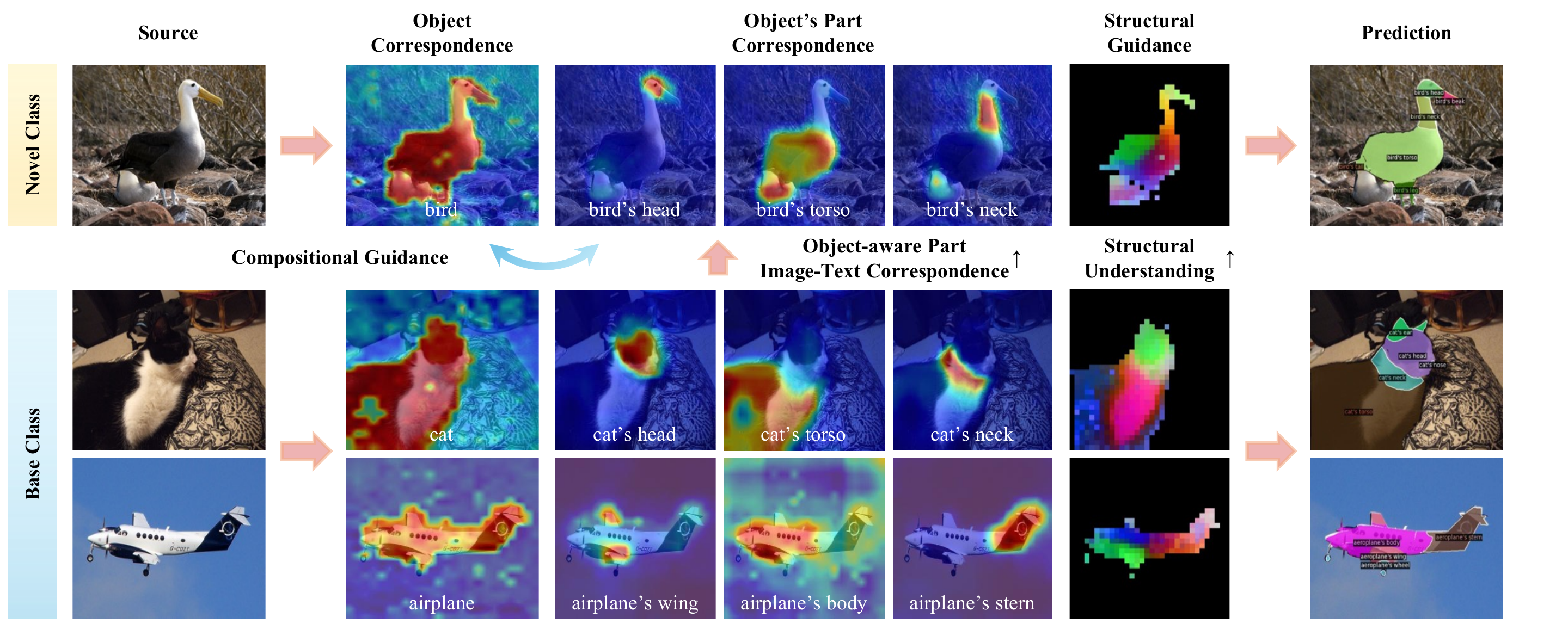}
        \vspace{-0.5em}
        \captionof{figure}{
            The proposed PartCATSeg exploits part-level and object-level image-text correspondence using cost aggregation, enhancing object-aware part image-text matching. Additionally, it utilizes structural guidance to achieve successful part segmentation even for classes not encountered during training.
        }
        \label{fig:teaser}
    \end{minipage}
\end{strip}

\begin{abstract}
\noindent
Open-Vocabulary Part Segmentation (OVPS) is an emerging field for recognizing fine-grained parts in unseen categories.
We identify two primary challenges in OVPS: (1) the difficulty in aligning part-level image-text correspondence, and (2) the lack of structural understanding in segmenting object parts.
To address these issues, we propose PartCATSeg, a novel framework that integrates object-aware part-level cost aggregation, compositional loss, and structural guidance from DINO.
Our approach employs a disentangled cost aggregation strategy that handles object and part-level costs separately, enhancing the precision of part-level segmentation.
We also introduce a compositional loss to better capture part-object relationships, compensating for the limited part annotations.
Additionally, structural guidance from DINO features improves boundary delineation and inter-part understanding.
Extensive experiments on Pascal-Part-116, ADE20K-Part-234, and PartImageNet datasets demonstrate that our method significantly outperforms state-of-the-art approaches, setting a new baseline for robust generalization to unseen part categories.

\end{abstract}


\vspace{-0.5em}

\section{Introduction}
\label{sec:intro}

\begin{figure}[t]
    \centering
    \small
    \begin{subfigure}{0.15\textwidth} \centering
        \includegraphics[width=1\linewidth]{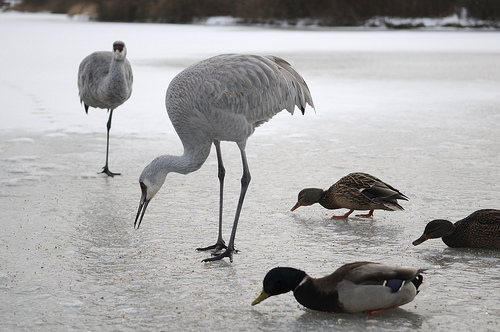}
        \caption{Original Image}
    \end{subfigure}
    \hfill
    \begin{subfigure}{0.15\textwidth} \centering
        \includegraphics[width=1\linewidth]{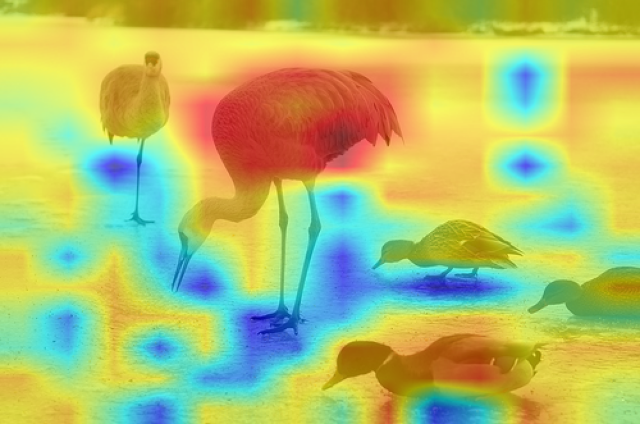}
        \caption{``head''}
    \end{subfigure}
    \hfill
    \begin{subfigure}{0.15\textwidth} \centering
        \includegraphics[width=1\linewidth]{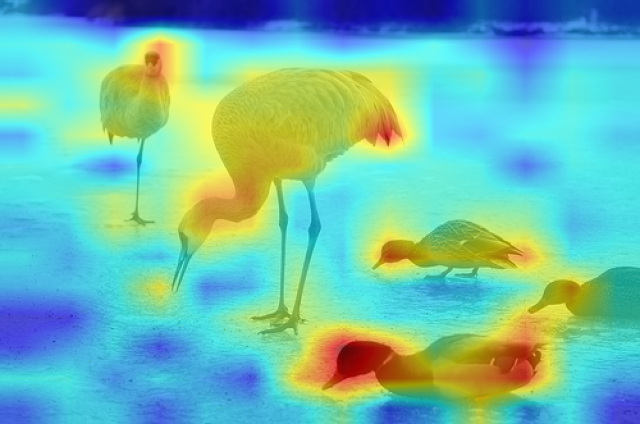}
        \caption{``wing''}
    \end{subfigure}
    \hfill
    \begin{subfigure}{0.155\textwidth} \centering
        \includegraphics[width=1\linewidth]{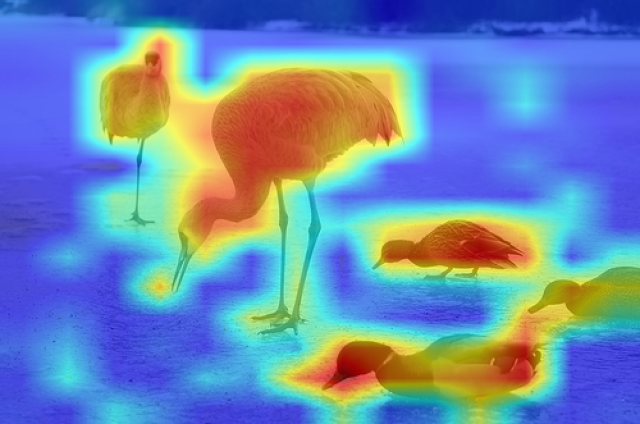}
        \caption{``bird''}
    \end{subfigure}
    \hfill
    \begin{subfigure}{0.155\textwidth} \centering
        \includegraphics[width=1\linewidth]{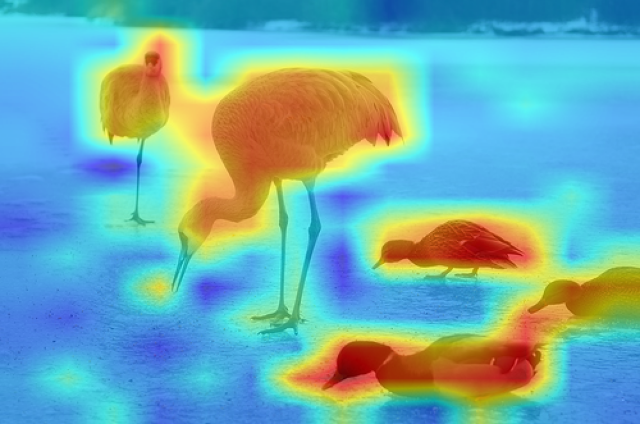}
        \caption{``bird's head''}
    \end{subfigure}
    \hfill
    \begin{subfigure}{0.155\textwidth} \centering
        \includegraphics[width=1\linewidth]{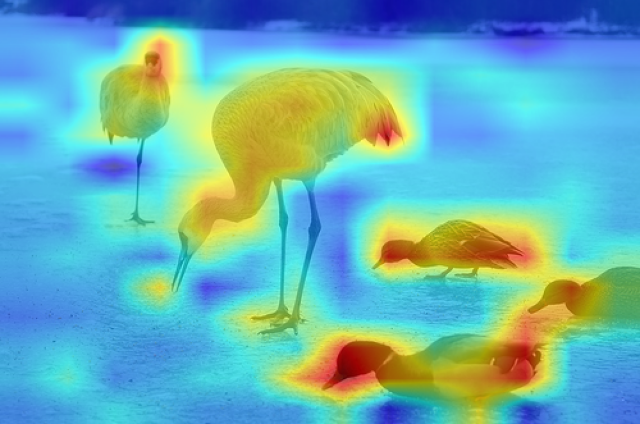}
        \caption{``bird's wing''}
    \end{subfigure}
    \hfill
    \vspace{-2em}
    \caption{
        \textbf{CLIP Image-Text Similarity Visualization for Object-Level and Part-Level Text.}
        The visualization~\cite{li2023clip_surgery} compares the frozen CLIP~\cite{radford2021learning_CLIP} image-text similarity between object-level and part-level text descriptions.
        (a) shows the original images; (b), (c) depict the part-level similarities for terms such as "head" and "wing" while (e), (f) show object-specific parts.
        The stronger activation for object-level text in (d) suggests a dominant focus on the entire object rather than individual parts in the image-text correspondence.
    }
    \label{fig:intro_limitations_clip_sim}
    \vspace{-1.5em}
\end{figure}



\noindent
The rapid advancements in model architectures and improved training techniques have allowed cutting-edge models to achieve exceptional performance on closed-set datasets and object-level segmentation~\cite{long2015fully_FCN, ronneberger2015_unet, he2017mask_rcnn, cheng2021per_MaskFormer, cheng2022masked_Mask2Former, chen2017deeplabv3, li2023mask_dino}.
However, they still exhibit suboptimal performance in zero-shot scenarios~\cite{akata2015label_zero_shot, frome2013devise_zero_shot_learning, akata2015evaluation_zero_shot_learning, bucher2019zero_ZS3Net} and open-vocabulary setups~\cite{zhao2017open, zhou2022extract_MaskCLIP, xu2022simple_ZSSeg, gu2021open_ViLD, liang2023open_OVSeg}.
Moreover, part-level fine-grained segmentation~\cite{chen2014detect_PascalPart, he2022partimagenet_PartImageNet, balbuena2013paco_PACO} remains challenging, highlighting the ongoing difficulties in these research areas.

Recently, {Open-Vocabulary Part Segmentation (OVPS)} has emerged as a novel approach aimed at recognizing novel objects at a fine-grained, part level~\cite{sun2023going_VLPart, wei2024ov_OV_PARTS, PartCLIPSeg2024, li2024partglee}.
OVPS enables the segmentation of detailed object parts, even for categories not seen during training, positioning it as both a research challenge and a high-demand application area.
This framework has numerous practical applications, including robotic control~\cite{wan2024instructpart_InstructPart, yenamandra2023homerobot}, medical imaging~\cite{yu2020c2fnas_medical}, image editing~\cite{ling2021editgan_editing, liu20203d_3D_part_editing}, and image generation~\cite{wang2024instancediffusion_InstanceDiffusion, koo2023salad}, serving as a significant milestone for the advancement of visual systems.



OVPS research has recently accelerated, leveraging pretrained vision-language models (VLMs)~\cite{radford2021learning_CLIP,jia2021scaling_ALIGN_google,li2022blip} to transfer part-level knowledge from base to novel classes.
Recently,
VLPart~\cite{sun2023going_VLPart} utilizes DINO~\cite{caron2021emerging_DINO,oquab2023dinov2} features to establish part-level correspondence between base and novel categories, while OV-PARTS~\cite{wei2024ov_OV_PARTS} focuses on leveraging object-level contextual information to enhance part-level segmentation.
More recent approaches, such as PartGLEE~\cite{li2024partglee} and PartCLIPSeg~\cite{PartCLIPSeg2024}, further advance the field by employing attention-based approaches with joint training of object and part to refine part segmentation.
Despite recent advancements, OVPS still suffers from poor performance, as key challenges remain unresolved.
First, aligning part-level text with its corresponding visual features is more challenging than at the object level.
This stems from the significantly lower ratio of part-level text-image pairs compared to object-level text during the pretraining of VLMs, as noted in previous studies~\cite{sun2023going_VLPart,pan2023towards_OPS_OWPS,he2022partimagenet_PartImageNet}.
Consequently, part-level data often receives weaker or noisier guidance compared to object-level data due to its diversity.
Additionally, the limited availability of part-level ground truth (GT) labels in base datasets further degrades performance.
\Cref{fig:intro_limitations_clip_sim} (b, c, e, f) illustrate the lower similarity and weaker alignment of part-level text compared to object-level text~\cite{radford2021learning_CLIP, li2023clip_surgery} in pretrained CLIP.


\begin{figure}[t]
    \centering
    \begin{subfigure}[t]{0.49\linewidth}
        \centering
        \includegraphics[width=\linewidth, keepaspectratio, trim=0 0 20 15, clip]{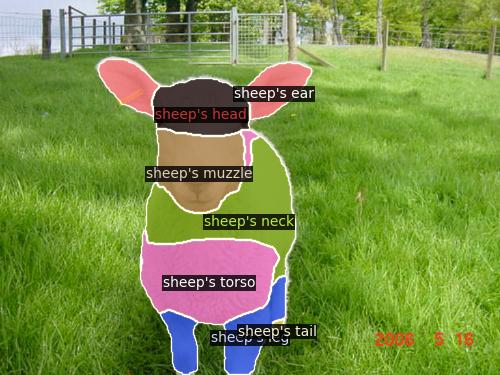}
        \caption{\normalsize CAT-Seg~\cite{cho2023cat_CATSeg}}
    \end{subfigure}
    \hfill
    \begin{subfigure}[t]{0.49\linewidth}
        \centering
        \includegraphics[width=\linewidth, keepaspectratio, trim=85 20 0 30, clip]{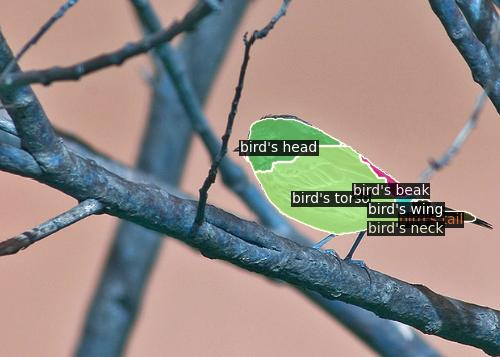}
        \caption{\normalsize PartCLIPSeg~\cite{PartCLIPSeg2024}}
    \end{subfigure}
    \vspace{-1em}
    \caption{
        \textbf{Challenges of Current OVPS.}
        (a, b) Prediction results of state-of-the-art OVSS~\cite{cho2023cat_CATSeg} and OVPS~\cite{PartCLIPSeg2024} methods.
        Due to a lack of structural understanding, these methods often produce incorrect OVPS predictions, such as predicting a ``sheep's neck'' as being larger than its ``torso'' or placing a ``bird's beak'' and a ``bird's neck'' at the tip of its ``tail''.
    }
    \vspace{-1.5em}
    \label{fig:intro_limitations_structural}
\end{figure}

Second, current OVPS approaches struggle to achieve a comprehensive structural understanding between parts and objects.
As shown in~\Cref{fig:intro_limitations_structural}, existing state-of-the-art models may misclassify a ``leg'' as a ``tail'' due to an incomplete understanding of part-object relationships and structural information.
This happens because the visual features of the ``leg'' resemble those of the ``tail'', leading to misrecognition.
This issue is similar to the part-whole illusion~\cite{tanaka1993parts_and_whiles} observed in humans when focusing only on parts without considering the whole.

We propose a novel framework, PartCATSeg, to address the critical challenges of OVPS and significantly improve model performance, as summarized in~\Cref{fig:teaser}.
First, we focus on the issue of \textit{part-level image guidance} being biased toward object-level guidance and propose a solution to mitigate this bias.
Specifically, we adopt a cost aggregation strategy for explicit handling of image-text correspondence, as outlined in~\cite{cho2021cats, cho2023cat_CATSeg}.
Then, we extend this model architecture to separate and operate with cost volumes tailored to both object and part levels.
This extension enables our model to disentangle and process object-level and part-level costs independently.
This change provides more precise guidance and enhances image-text alignment, particularly at the part level.
To further augment training signals, we integrate an unsupervised compositional loss that enforces the compositional relationship where parts collectively represent the object.
This approach ensures robust part representation even with the lack of explicit part-level supervision.

Second, we leverage DINO~\cite{li2023mask_dino,oquab2023dinov2} features for \textit{structural guidance} to improve the model's ability to understand inter-part relationships and effectively separate objects from the background.
Previous works, such as VLPart, highlighted DINO's rich semantic representations and utilized them for matching novel classes to their closest base classes.
Unlike their perspective, our study focuses on DINO’s capability to convey spatial structure and geometric proximity information between features.
To fully exploit this rich geometric prior, we incorporate pixel-level features as guidance when conducting the cost aggregation.
This pixel-level similarity between features proves particularly beneficial for distinguishing parts.

Our approach sets a new standard baseline for OVPS, demonstrating remarkable performance gains across diverse unseen categories in multiple datasets. Extensive experiments on benchmark datasets such as Pascal-Part-116, ADE20K-Part-234, and PartImageNet show that our method significantly outperforms existing state-of-the-art methods. We achieve notable improvements in segmentation accuracy and generalization capabilities, with increases in h-IoU scores by over 10\% on all datasets in the Pred-All setting. 
Our analyses confirm that the combination of object-aware part-level cost aggregation techniques, the compositional loss, and structural guidance from DINO contribute to these performance gains, validating the effectiveness of our proposed framework in addressing the critical challenges of OVPS.







\section{Related Work}
\label{sec:related_work}



\noindent \textbf{Open-Vocabulary Semantic Segmentation (OVSS).}
OVSS~\cite{gu2021open_ViLD, zareian2021open_OVR_CNN_OV_RCNN, li2022language_LSeg, luddecke2022image_CLIPSeg, zhao2017open, xian2019semantic, zhou2022extract_MaskCLIP, liu2023open_SCAN, zhou2023zegclip, xie2023sed_SED, yu2024convolutions_FC_CLIP} advances beyond traditional semantic segmentation's~\cite{he2017mask_rcnn,cheng2022masked_Mask2Former,chen2017deeplabv3} predefined class limitation by enabling predictions for unseen classes.
OVSS methods generally rely on the shared embedding space of vision-language models (VLMs)~\cite{jia2021scaling_ALIGN_google,li2022blip,radford2021learning_CLIP} such as CLIP, allowing both visual and text features to be aligned in a unified semantic space.
There are two main approaches.
The first approach is a two-stage approach~\cite{zareian2021open_OVR_CNN_OV_RCNN, ding2022open, ghiasi2022scaling_OpenSeg, han2023global_GKC, liang2023open_OVSeg,liu2023open_SCAN,xu2023open_ODISE,xu2022simple_ZSSeg}, such as ZSSeg and ODISE, that first generates class-agnostic masks before VLM-based classification.
The second approach is single-stage frameworks~\cite{cho2023cat_CATSeg,li2022language_LSeg,luddecke2022image_CLIPSeg,xie2023sed_SED,yu2024convolutions_FC_CLIP,zhou2023zegclip} such as CLIPSeg and CAT-Seg that directly align pixel-level features with text features.

 \noindent \textbf{Dense Correspondence and Cost Aggregation.}
In computer vision, \textit{cost} computation is used to evaluate \textit{dense visual correspondence}~\cite{hosni2012fast_cost,sun2018pwc_cost,truong2020glu_flow_cost,hong2022cost_4d,chen2023costformer,cho2021cats,hong2024unifying_cost,min2019hyperpixel,min2020learning,liu2020semantic} by assessing the differences between matching points.
From the perspective of Optimal Transport~\cite{villani2009optimal_book, peyre2019computational}, this cost reflects the difficulty of matching or high disparity.
Minimizing disparity aligns with maximizing similarity.
Hence, we interpret cost as a measure of correspondence, similar to prior notations~\cite{cho2021cats,chen2023costformer,hong2022cost_4d,hong2024unifying_cost}.
\noindent \textit{Cost Aggregation}~\cite{hosni2012fast_cost,sun2018pwc_cost,truong2020glu_flow_cost,hong2022cost_4d,chen2023costformer,cho2021cats,cho2022cats_pp} is a key framework to reduce matching errors and enhance generalization in dense correspondence tasks.
Early methods using handcrafted approaches~\cite{liu2010sift,dalal2005histograms} transitioned to learnable kernels with 2D and 4D convolutions~\cite{hong2022cost_4d,rocco2018neighbourhood_NC}. With the advent of Transformer architectures, CATs and CATs++~\cite{cho2021cats,cho2022cats_pp} demonstrated the effectiveness of Transformer-based cost aggregation. Recently, CAT-Seg~\cite{cho2023cat_CATSeg} extended this concept to multi-modal settings, using cosine similarity between image and text embeddings as a matching cost volume. In this study, we apply cost aggregation to open-vocabulary part segmentation, enabling recognition of finer-grained entities.

\noindent \textbf{Part Segmentation}~\cite{chen2014detect_PascalPart,he2022partimagenet_PartImageNet,balbuena2013paco_PACO,chen2014detect_PascalPart,choudhury2021unsupervised_UnsupervisedPartDiscovery,he2023compositor_Compositor,hung2019scops_SCOPS,van2023pdisconet_PDiscoNet} aims to recognize objects by dividing them into finer units (parts).
Early works focused on contrastive and self-supervised co-part segmentation~\cite{hung2019scops_SCOPS,he2023compositor_Compositor,choudhury2021unsupervised_UnsupervisedPartDiscovery,van2023pdisconet_PDiscoNet}, establishing foundational methods for identifying object parts without extensive labeled data.
Recent developments have advancing toward open-vocabulary approaches, with notable research including VLPart~\cite{sun2023going_VLPart}, OV-PARTS~\cite{wei2024ov_OV_PARTS}, PartGLEE~\cite{li2024partglee}, and PartCLIPSeg~\cite{PartCLIPSeg2024}. VLPart aimed to generalize to novel classes using DINO correspondence~\cite{li2023mask_dino,oquab2023dinov2}, OV-PARTS introduced object mask prompts, while PartGLEE and PartCLIPSeg combined part information with object-level context.
These approaches have enhanced part segmentation performance, particularly in handling previously unseen categories.
In this work, we further advance these studies by improving the image-text correspondence for both objects and parts.

\section{Method}
\label{sec:method}

\begin{figure*}[!t]
    \centering
    \includegraphics[width=0.850 \textwidth]{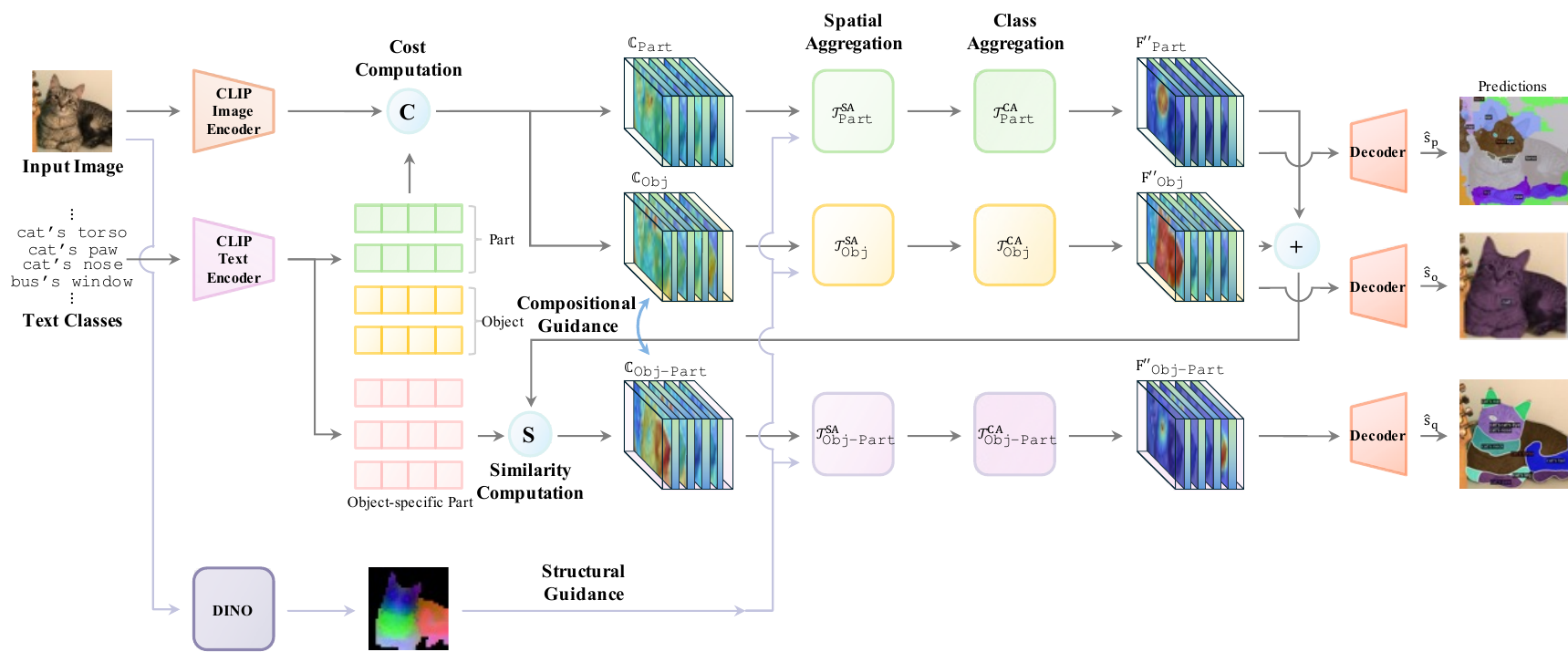}
    \vspace{-1.0em}
    \caption{
        The Overall Architecture of PartCATSeg.
    }
    \label{fig:overview}
\vspace{-1.0em}
\end{figure*}


\subsection{Problem Definition}
\label{subsec:Problem Definition}


\noindent
Open-Vocabulary Part Segmentation (OVPS)~\cite{sun2023going_VLPart,wei2024ov_OV_PARTS,li2024partglee,PartCLIPSeg2024} aims to segment an image into distinct \texttt{object-specific parts}, $\mathbf{C}_{\texttt{Obj-Part}}$, (e.g., ``cat's paw,'' ``bus's headlight'') without relying on predefined part categories.
During training, the model uses image and ground-truth mask pairs, where the masks represent only the $\mathbf{C}_{\texttt{Obj-Part}}$ of {base} categories, ${\mathbf{C}}_{\texttt{Obj-Part}}^{\texttt{base}}$.
At inference, the model segments {novel} categories, ${\mathbf{C}}_{\texttt{Obj-Part}}^{\texttt{novel}}$, beyond these {base} categories, leveraging learned representations to generalize without additional annotations.
OVPS performance is assessed in two scenarios~\cite{sun2023going_VLPart,wei2024ov_OV_PARTS,PartCLIPSeg2024}:
In \textbf{Zero-Shot} setup, the training and test categories do not overlap (${\mathbf{C}}_{\texttt{Obj-Part}}^{\texttt{base}} \cap {\mathbf{C}}_{\texttt{Obj-Part}}^{\texttt{novel}} = \emptyset)$.
While in \textbf{Cross-Dataset} setting, the model evaluated on a different dataset from the training set without fine-tuning, introducing challenges from domain shifts and differences in part granularity.



\noindent


\subsection{Preliminary}
\label{subsec:preliminary}


\noindent
\textbf{Image-Text Cost Computation.}
As described in \Cref{sec:related_work}, \textit{cost} refers to the correspondence between image-to-image or image-to-text features.
CAT-Seg~\cite{cho2023cat_CATSeg} extends the cost computation approach to effectively model image-text correspondence, showing promising performance in object-level open-vocabulary segmentation tasks~\cite{zhao2017open,ding2022open,gu2021open_ViLD}.
The initial cost $\mathbb{C} \in \mathbb{R}^{(H \times W) \times |T|}$ between the $i^{\text{th}}$ image patch and the $n^{\text{th}}$ text token is defined as:
\begin{equation}
    \textstyle
    \mathbb{C}(i, n)
        = \frac{
            D^V(i) \cdot D^L(n)
        }{
            \|D^V(i)\| \|D^L(n)\|
        },
\end{equation}
where $D^V(\cdot)$ and $D^L(\cdot)$ are dense visual and language embeddings, respectively.
Here, $i \in \mathbb{R}^{(H \times W)}$ is the spatial index of the image patch and $n \in \mathbb{R}^{T}$ denotes the index of the text token corresponding to category classes $T$.

\noindent \textbf{Cost Aggregation.}
The computed cost volume is refined through a process known as cost aggregation~\cite{cho2021cats, chen2023costformer, cho2023cat_CATSeg, hong2022cost_4d, hong2024unifying_cost}, which employs self-attention modules~\cite{vaswani2017attention_self_attention_transformer}.
This process consists of two main components: the spatial aggregation transformer ($\mathcal{T}^{\texttt{SA}}$) and the class aggregation transformer ($\mathcal{T}^{\texttt{CA}}$), both of which are designed to effectively model image-text correspondence.
\textbf{Spatial aggregation} utilizes the local continuity of image embeddings within the cost volume to capture spatially consistent features.
Specifically, Swin Transformer~\cite{liu2021swin} blocks are used to enhance semantic coherence by capturing both local and semi-global features, refining class representations, and reducing background noise. The initial cost volume feature $F \in \mathbb{R}^{(H \times W) \times |T| \times d}$ is derived from $\mathbb{C}$ using convolutional layers, represented as: $F = \texttt{Conv}(\mathbb{C})$.
The spatial aggregation operation for each class is then defined as:
\begin{equation}
    \textstyle
    F'(:, n) = \mathcal{T}^{\texttt{SA}}(F(:, n)), \quad F(:, n) \in \mathbb{R}^{(H \times W) \times d},
\end{equation}
where $\mathcal{T}^{\texttt{SA}}$ is applied independently for each class in the cost volume.
\textbf{Class aggregation} models inter-class relationships using a Transformer block without positional embeddings, ensuring permutation invariance.
This operation accounts for the spatial context of each class and its interactions with other classes, enabling a holistic cost calculation for each class:
\begin{equation}
    \textstyle
    F''(i, :) = \mathcal{T}^{\texttt{CA}}(F'(i, :)), \quad F' \in \mathbb{R}^{(H \times W) \times |T| \times d}.
\end{equation}
This two-stage aggregation framework refines the cost volume, enhancing prediction accuracy across diverse object categories.


\subsection{Disentangled Cost Aggregation}
\label{subsec:Disentangled Cost Aggregation}

Although CAT-Seg~\cite{cho2023cat_CATSeg} has shown promising results in open-vocabulary semantic segmentation (OVSS), its performance is limited when it comes to fine-grained, part-level categories.
This limitation arises from the predictions being biased towards object-level categories, caused by the lack of \textbf{part-level correspondence}, as shown in~\Cref{fig:intro_limitations_clip_sim}.

To address this issue, we propose a method of parsing object-level and part-level class names $\mathbf{C}_{\texttt{Obj}}$ and $\mathbf{C}_{\texttt{Part}}$, computing separate cost volumes for each, similar to approaches in OVPS frameworks like PartGLEE~\cite{li2024partglee} and PartCLIPSeg~\cite{PartCLIPSeg2024}.
However, our approach differentiates itself by explicitly constructing separate cost volumes from the image-text correspondence perspective.
Specifically, we compute the initial cost volumes for object-level categories and part-level categories, $\mathbb{C}_{\{\texttt{Obj}|\texttt{Part}\}} \in \mathbb{R}^{(H \times W) \times |\mathbf{C}_{\{\texttt{Obj}|\texttt{Part}\}}|}$ as follows:
\begin{equation}
    \textstyle
    \mathbb{C}_{\{\texttt{Obj}|\texttt{Part}\}}(i, x)
        = \frac{
            D^V(i) \cdot D^L(x)
        }{
            \|D^V(i)\| \|D^L(x)\|
        }, \, x \in \{o, p\},
\end{equation}
where $o$ and $p$ are the object-level and part-level class tokens, respectively, from the categories $\mathbf{C}_{\texttt{Obj}}$ and $\mathbf{C}_{\texttt{Part}}$.

We then obtain initial cost volume features $F_{\texttt{Obj}} \in \mathbb{R}^{(H\times W) \times |\mathbf{C}_{\texttt{Obj}}| \times d}$ and $F_{\texttt{Part}} \in \mathbb{R}^{(H\times W) \times |\mathbf{C}_{\texttt{Part}}| \times d}$ through convolutional layers, respectively.
These features are further refined using spatial and class aggregation transformers to enhance the correspondence between image regions and class tokens.
This allows us to compute cost volumes that are better aligned with the hierarchical concepts, providing more precise correspondence for both object-level and part-level class tokens.
As shown in~\Cref{fig:overview}, the features are processed through a sequence of spatial and class aggregation transformers as:
\begin{equation}
    \textstyle
    F''_{\texttt{Obj}}(i, :) = \mathcal{T}^{\texttt{CA}}_{\texttt{Obj}}(\mathcal{T}^{\texttt{SA}}_{\texttt{Obj}}(F_{\texttt{Obj}}(i, :))), 
\end{equation}
\begin{equation}
    \textstyle
    F''_{\texttt{Part}}(i, :) = \mathcal{T}^{\texttt{CA}}_{\texttt{Part}}(\mathcal{T}^{\texttt{SA}}_{\texttt{Part}}(F_{\texttt{Part}}(i, :))), 
\end{equation}
where $i$ is a spatial location index.
The refined cost volume embeddings $F''_{\texttt{Obj}}$ and $F''_{\texttt{Part}}$ are then fed into decoders to generate prediction masks $\hat{s}_{o}$ for objects and $\hat{s}_{p}$ for parts. These predictions are supervised using the corresponding ground truth masks $s_o$, $s_p$ with \textbf{disentanglement loss} as:
\begin{equation}
    \textstyle
    \mathcal{L}_{\texttt{disen}}
        = \lambda_{\texttt{Obj}} \sum\limits_{o \in {\mathbf{C}_{\texttt{Obj}}}}
            \text{\sc{Bce}}(s_{o}, \hat{s}_{o})
        + \lambda_{\texttt{Part}} \sum\limits_{p \in {\mathbf{C}_{\texttt{Part}}}}
            \text{\sc{Bce}}(s_{p}, \hat{s}_{p}),
\label{eq:obj_part_level_guidance}
\end{equation}
where $\text{\sc{Bce}}$ denotes the binary cross-entropy loss, and $\lambda_{\texttt{Obj}}$ and $\lambda_{\texttt{Part}}$ are hyperparameters balancing two losses.
By leveraging this decoupled and disentangled cost aggregation strategy, our approach effectively enhances object-aware part-level image-text correspondence.

\begin{figure}[t]
    \centering
    \includegraphics[width=1.0\linewidth]{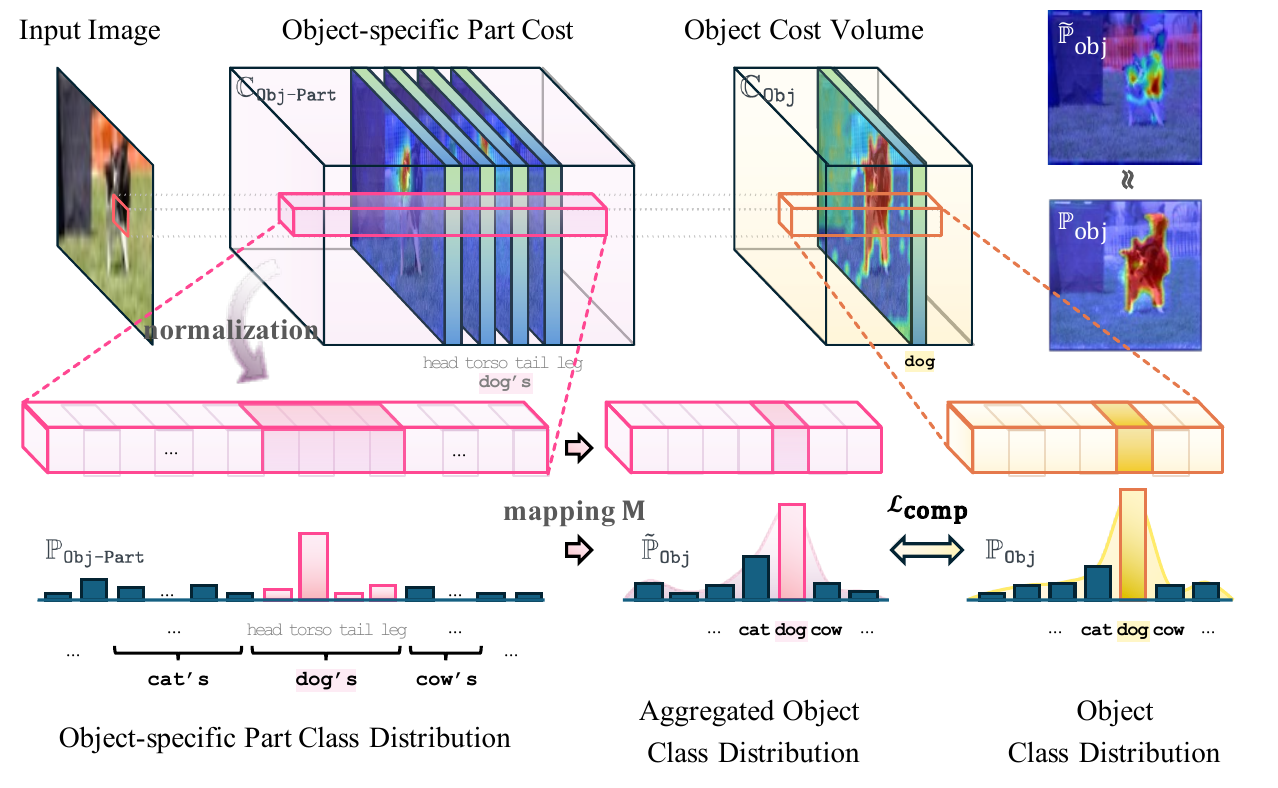}
    \vspace{-1em}
    \caption{
        \textbf{Compositional Loss.}
        The loss function $\mathcal{L}_{\text{comp}}$ guides learning by ensuring that the aggregated class distribution of object parts aligns closely with the overall class distribution of the object, enhancing the consistency between part-level and object-level representations.
    }
    \label{fig:method_compositional_component}
    \vspace{-1em}

\end{figure}


\subsection{Object-aware Part Cost Aggregation}
\label{subsec:Object-aware Part Cost Aggregation}
To further enhance image-text correspondence for object-specific parts, we integrate the object-level cost feature $F''_{\texttt{Obj}}$ and the part-level cost feature $F''_{\texttt{Part}}$.
This integration forms an object-part combination, denoted as $F_{\texttt{Obj-Part}} \in \mathbb{R}^{(H \times W) \times |\mathbf{C}_{\texttt{Obj-Part}}| \times d}$, using a simple projection as follows:
\begin{equation} 
    \textstyle
    F_{\texttt{Obj-Part}}(i)
        = \texttt{Linear} \left( \left[
            F''_{\texttt{Obj}}(i) ; F''_{\texttt{Part}}(i)
        \right] \right),
\end{equation}
where $[\cdot;\cdot]$ denotes concatenation and $\texttt{Linear}$ is a projection layer. By combining object and part features, the model integrates object-level context with part-level details. 
We then align the integrated features with the object-specific part category names, $\mathbf{C}_{\texttt{Obj-Part}}$, by computing the similarity between $F_{\texttt{Obj-Part}}$ and the object-specific part text embeddings $q \in \mathbf{C}_{\texttt{Obj-Part}}$, thereby obtaining $\mathbb{C}_{\texttt{Obj-Part}}$ as:
\begin{equation} 
    \textstyle
    \mathbb{C}_{\texttt{Obj-Part}}(i, q) = \frac{ F_{\texttt{Obj-Part}}(i, q) \cdot D^L(q) }{ \| F_{\texttt{Obj-Part}}(i, q) \|  \| D^L(q) \| }.
\end{equation}
The updated object-aware part cost undergoes a convolution operation to transform it into a feature representation, $F_{\texttt{Obj-Part}}$, before refinement through spatial and class aggregation transformers.
\begin{equation} 
F''_{\texttt{Obj-Part}}(i, :) = \mathcal{T}^{\texttt{CA}}_{\texttt{Obj-Part}}\left( \mathcal{T}^{\texttt{SA}}_{\texttt{Obj-Part}}\left( F_{\texttt{Obj-Part}}(i, :) \right) \right), \end{equation}
where $F_{\texttt{Obj-Part}}$ is a convoluted object-aware cost feature from $\mathbb{C}_{\texttt{Obj-Part}}$.
By propagating object-level and part-level semantics between image and text, this approach improves image-text correspondence alignment, enhancing accurate boundary recognition at both object and part levels.
Finally, the refined object-aware part embeddings $F_{\texttt{Obj-Part}}$ are fed into the shared decoder to generate the prediction masks $\hat{s}_{q}$ and supervised using the ground truth masks $s_q$ with the following loss: $\mathcal{L}_{\texttt{Obj-Part}} = \sum_{q \in \mathbf{C}_{\texttt{Obj-Part}}} \text{\sc{Bce}}\left( s_{q}, \hat{s}_{q} \right)$.




\subsection{Cost as Compositional Components}
\label{subsec:Cost as Compositional Components}

Parts are \textit{compositional components} that constitute an object.
However, the ground truth annotations for parts are relatively scarce compared to those for objects, making supervision from part annotations alone insufficient.


\noindent \textbf{Compositional Bias.}
To address this problem, we propose a compositional loss, $\mathcal{L}_{\texttt{comp}}$, that enforces inductive bias of ensuring that the parts are the compositional component of its object.
This loss function enhances the model's capacity to capture spatial relationships among parts and improves the part boundary delineation, particularly for smaller components.
Specifically, at each spatial location $i \in \mathbb{R}^{(H \times W)}$ in the cost volumes, we compute the normalized distributions ${\mathbb{P}}_{\texttt{Obj}}$ and ${\mathbb{P}}_{\texttt{Obj-Part}}$ over object and part classes, respectively, by applying a softmax function to $\mathbb{C}_{\texttt{Obj}}$ and $\mathbb{C}_{\texttt{Obj-Part}}$ along the class dimension as:
\begin{equation}
    \textstyle
    {\mathbb{P}}_{\texttt{Obj}}(i, o) = \frac{\exp\left( \mathbb{C}_{\texttt{Obj}}(i, o) \right)}{\sum_{k \in \mathbf{C}_{\texttt{Obj}}} \exp\left( \mathbb{C}_{\texttt{Obj}}(i, k) \right)},
\end{equation} \vspace{-1.5mm}
\begin{equation}
    \textstyle
    {\mathbb{P}}_{\texttt{Obj-Part}}(i, p)
        = \frac{\exp\left( \mathbb{C}_{\texttt{Obj-Part}}(i, p) \right)}{\sum_{k \in \mathbf{C}_{\texttt{Obj-Part}}} \exp\left( \mathbb{C}_{\texttt{Obj-Part}}(i, k) \right)}.
\end{equation}
We then map the part distributions to their corresponding object classes using a predefined mapping $M:\left\{1, \dots, |\mathbf{C}_{\texttt{Obj-Part}}| \right\} \rightarrow \left\{1, \dots, |\mathbf{C}_{\texttt{Obj}}| \right\}$, which is deterministically defined based on $\mathbf{C}_{\texttt{Obj-Part}}$.
The aggregated object class distribution ${\tilde{\mathbb{P}}}_{\texttt{Obj}} \in \mathbb{R}^{(H \times W) \times |\mathbf{C}_{\texttt{Obj}}|}$ is then obtained by summing the mapped part probabilities as:
\begin{equation}
    \textstyle
    \tilde{\mathbb{P}}_{\texttt{Obj}}(i, o) = \sum_{p \in M^{-1}(o)} {\mathbb{P}}_{\texttt{Obj-Part}}(i, p).
\end{equation}
We finally compute the Jensen-Shannon~\cite{lin1991divergence_jensen, englesson2021generalized_jensen} divergence between the object's distribution and the aggregated object distribution at each spatial location:
\begin{equation} 
    \textstyle
    \mathcal{L}_{\texttt{comp}}
    = \frac{1}{2} \left(D_\text{KL}( {\mathbb{P}}_{\texttt{Obj}} \| \tilde{\mathbb{P}}_{\texttt{Obj}}) + D_\text{KL}( \tilde{\mathbb{P}}_{\texttt{Obj}} \| {\mathbb{P}}_{\texttt{Obj}})\right),
\end{equation}
By enforcing the similarity between the object's distribution and the aggregated object distribution, we inject the inductive bias that parts collectively compose the object.
This encourages the model to recognize the spatial relationships among parts and ensures that smaller parts are appropriately delineated.


Ultimately, our training objective aims to minimize the following loss function:
\begin{equation} 
    \textstyle
    \mathcal{L} = \mathcal{L}_{\texttt{Obj-Part}} + \mathcal{L}_{\texttt{disen}} + \lambda_{\texttt{comp}} \mathcal{L}_{\texttt{comp}},
\label{eq:final_loss}
\end{equation}
where $\lambda_{\texttt{comp}}$ is a hyperparameter for balancing the loss.

\begin{figure}[t]
    \begin{subfigure}[b]{0.49\linewidth}
        \includegraphics[width=0.49\linewidth]{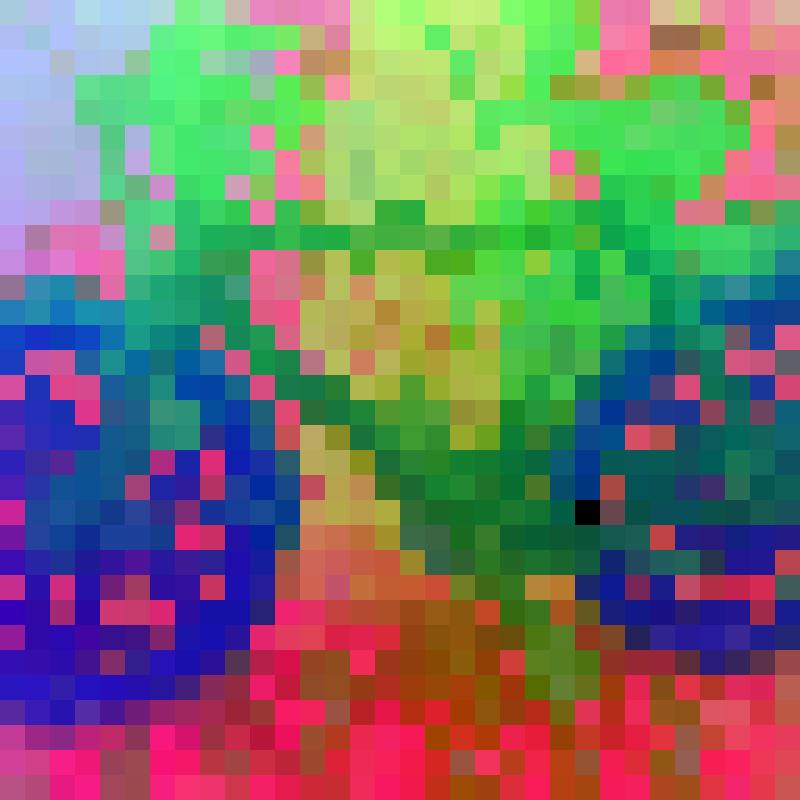}
        \includegraphics[width=0.49\linewidth]{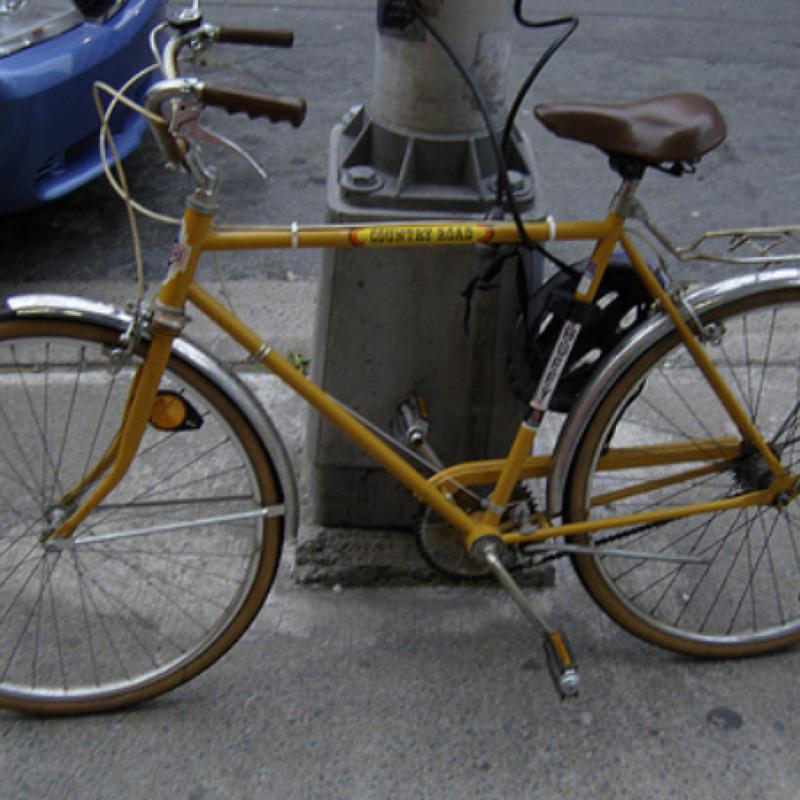}
        \caption{\normalsize w/o mask}
        \label{fig:dino_subfig2_without_mask}
    \end{subfigure}
    \hfill
    \begin{subfigure}[b]{0.49\linewidth}
        \includegraphics[width=0.49\linewidth]{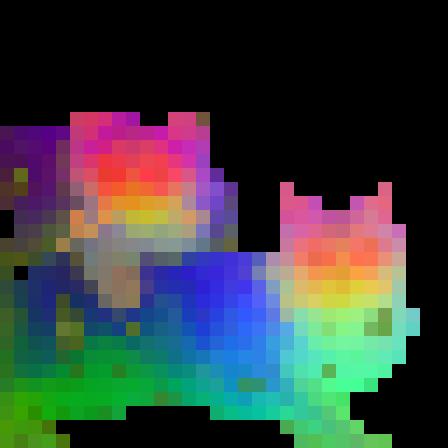}
        \includegraphics[width=0.49\linewidth]{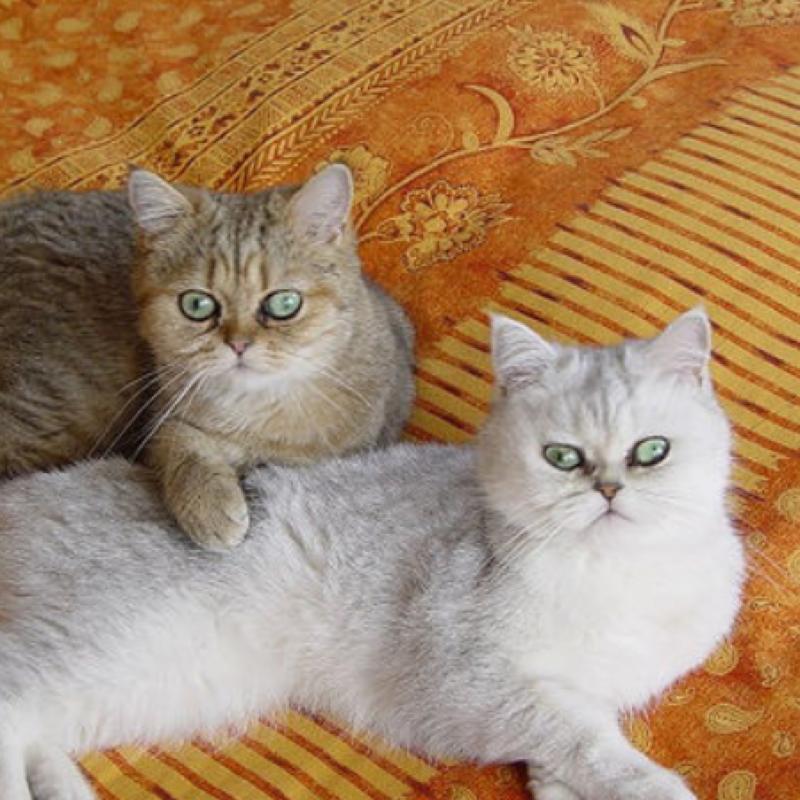}
        \caption{\normalsize w/ mask}
        \label{fig:dino_subfig1_withmask}
    \end{subfigure}
    \vspace{-1.0em}
    \caption{
        \textbf{Visualization of Frozen DINO PCA results.}
        (a) Without mask: The PCA visualization shows that DINO can separate object features from the background.
        (b) With mask: highlights consistent feature values across structurally similar parts of the objects, indicating strong part correspondences.
        This showcases DINO’s ability to capture detailed object representations.
    }
    \label{fig:method_dino}
    \vspace{-1.0em}

\end{figure}


\subsection{Cost with Structural Guidance}
\label{subsec:Cost with Structural Guidance}

\noindent
%
%
%
We utilize DINO's~\cite{caron2021emerging_DINO,oquab2023dinov2} self-supervised learning capabilities, which are known to exhibit strong semantic and geometric priors.
DINO features have previously demonstrated their effectiveness in VLPart~\cite{sun2023going_VLPart} by aiding in the correspondence computation between base and novel classes.
While VLPart focused on image-to-image correspondences, we extend this approach by forming an image-text cost volume, enabling the use of DINO features for image-text correspondence directly related to specific categories.

We employ DINO features to provide structural guidance during spatial aggregation.
As shown in Figure~\ref{fig:method_dino} (a), the PCA visualization illustrates that DINO features effectively distinguish between objects and background components.
Additionally, in Figure~\ref{fig:method_dino} (b), after background masking and PCA application, we observed detailed intra-object information, revealing continuous PCA results within objects.
This indicates that DINO features effectively capture structural information useful for predicting part relationships.
The visualization further shows consistent representations for structurally similar parts (e.g., ``heads'' in red, ``torso'' in blue, and ``legs'' in green for two ``cats''), demonstrating the alignment capability of DINO features.

\noindent \textbf{Structural Guidance.} To maximize the use of these structural features, we introduce structural guidance during the spatial aggregation process by incorporating it into the $Query$ and $Key$ matrices along with the cost volume as:
\begin{equation}
    \textstyle
    F'(:, n) = \mathcal{T}^{\texttt{SA}}(F(:, n), F_\texttt{dino}(:, n)) \in \mathbb{R}^{(H \times W) \times (d + d_\texttt{dino})}.
\end{equation}
The effectiveness of our approach is comprehensively evaluated in the ablation study (see~\Cref{subsec:Ablation Study}).
The study demonstrates the impact of incorporating structural guidance into the cost volume, leading to improved part and object image-text correspondences.


\section{Experiments}
\label{sec:experiments}

\subsection{Experimental Setup}

\noindent \textbf{Datasets.} We perform our evaluation on three OVPS benchmarks: Pascal-Part-116~\cite{chen2014detect_PascalPart,wei2024ov_OV_PARTS}, ADE20K-Part-234~\cite{wei2024ov_OV_PARTS,zhou2017scene_ADE20K}, and PartImageNet~\cite{he2022partimagenet_PartImageNet}.
Pascal-Part-116 is an adapted version of PascalPart~\cite{chen2014detect_PascalPart} of OVPS, containing 8k training images and 850 test images, with 116 part classes across 17 object categories.
ADE20K-Part-234, which is a more challenging and fine-grained dataset, includes 7k training images and 1k validation images with 234 part classes in 44 object categories.
PartImageNet provides 16k training images and 2.9k validation images across 158 object classes from ImageNet~\cite{deng2009imagenet}, with 40 representative classes used for cross-dataset evaluation in this study.
Additionally, we use PartImageNet (OOD), a split originally designed for few-shot learning, consisting of 109 training, 19 validation, and 30 test classes.
This split is included for cross-dataset evaluation.
Additional details of the datasets are provided in the supplementary materials.

\noindent \textbf{Evaluation Metrics.}
To comprehensively evaluate OVPS performance, we use two protocols: (1) \textbf{Pred-All} introduced by PartCLIPSeg~\cite{PartCLIPSeg2024}, where the model performs segmentation without access to object masks or class labels, relying entirely on its own predictions, and (2) \textbf{Oracle-Obj}, where ground truth object-level masks and class labels are provided as in setting of OV-PARTS~\cite{wei2024ov_OV_PARTS}.
This setup assumes the scenario of part segmentation with the reliance on an object-level segmentation results.
We measure segmentation performance using mIoU and assess generalization via the harmonic mean (h-IoU) of seen and unseen categories.

\noindent \textbf{Implementation Details.}
We adopted the training approach of CAT-Seg~\cite{cho2023cat_CATSeg} and fine-tuned the Query and Value heads of CLIP's encoders.
Details on the information on the experimental setup, hyperparameters, and training can be found in the supplementary materials.

\subsection{Main Results}

\noindent \textbf{Zero-Shot Part Segmentation.} We compare PartCATSeg with previous methods on three popular OVPS benchmarks, demonstrating its significant performance and strong generalization capabilities. As shown in \Cref{tab:main_pascal}, PartCATSeg consistently outperforms existing methods on Pascal-Part-116, with significant improvements in h-IoU: a 15.10\% increase in Pred-All and 11.62\% increase in Oracle-Obj over the second-best method, PartCLIPSeg~\cite{PartCLIPSeg2024}. 
On the more challenging ADE20K-Part-234 dataset, as demonstrated in \Cref{tab:main_ade}, PartCATSeg again outperforms previous SoTA methods, achieving a remarkable 12.81\% improvement in Pred-All and 8.13\% in Oracle-Obj. 
On PartImageNet, as in \Cref{tab:main_partimagenet}, PartCATSeg sets new performance benchmarks, breaking through the h-IoU thresholds with 55.12\% in Pred-All and 72.66\% in Oracle-Obj. These results highlight PartCATSeg's advanced capabilities, particularly in performance on unseen categories, positioning it as a leading solution in OVPS with exceptional accuracy in both seen and unseen classes. By addressing the challenges in OVPS, our method achieves a significant advancement in OVPS, highlighting PartCATSeg's robustness and superior generalization abilities across diverse datasets.

\begin{table}[t!]
    \centering
    \small
    \resizebox{\linewidth}{!}
    {

    \begin{threeparttable}
        \begin{tabular}{@{}l ccc ccc@{}}
            \toprule
            \multicolumn{1}{c}{\multirow{2}{*}{Method}}                              & \multicolumn{3}{c}{Pred-All} & \multicolumn{3}{c}{Oracle-Obj} \\
            \cmidrule(l){2-4} \cmidrule(l){5-7}
            & Seen & Unseen & h-IoU & Seen & Unseen & h-IoU  \\ \midrule
            \multirow{1}{*}{ZSSeg+ \cite{xu2022simple_ZSSeg}}                        &  38.05  &   3.38  &   6.20  &  54.43  &  19.04  &  28.21 \\
            \multirow{1}{*}{VLPart \cite{sun2023going_VLPart}}                       &  35.21  &   9.04  &  14.39  &  42.61  &  18.70  &  25.99 \\
            CLIPSeg \cite{luddecke2022image_CLIPSeg,wei2024ov_OV_PARTS}              &  27.79  &  13.27  &  17.96  &  48.91  &  27.54  &  35.24 \\
            \multirow{1}{*}{CAT-Seg \cite{cho2023cat_CATSeg,wei2024ov_OV_PARTS}}     &  36.80  &  23.39  &  28.60  &  43.81  &  27.66  &  33.91 \\
            \multirow{1}{*}{PartGLEE \cite{li2024partglee}}                          &  -  &  -  &  -  &  \underline{57.43}  &  27.41  &  37.11 \\
            PartCLIPSeg \cite{PartCLIPSeg2024}                                       &  \underline{43.91}  &  \underline{23.56}  &  \underline{30.67}  &  50.02  &  \underline{31.67}  &  \underline{38.79} \\
            \midrule
            PartCATSeg (Ours)                                                        &  \textbf{52.62} & \textbf{40.51} & \textbf{45.77} & \textbf{57.49} & \textbf{44.88} & \textbf{50.41} \\
            \vspace{-4pt}                                                            &  &  &  \gainp{+15.10}  &  &  &  \gainp{+11.62} \\
            \bottomrule
        \end{tabular}
        \begin{tablenotes}
            \item[1] The best score is \textbf{bold} and the second-best score is \underline{underlined}.
        \end{tablenotes}
        \end{threeparttable}
    }
    \vspace{-1.0em}
    \caption{
        Comparison of zero-shot performance with state-of-the-art methods on Pascal-Part-116.
    }
    \label{tab:main_pascal}        
    \vspace{-1.0em}
\end{table}

\small

\begin{table}[t!]
    \centering
    \small
    \resizebox{\linewidth}{!}
    {
        \begin{tabular}{@{}l ccc ccc@{}}
            \toprule
            \multicolumn{1}{c}{\multirow{2}{*}{Method}} & \multicolumn{3}{c}{Pred-All} & \multicolumn{3}{c}{Oracle-Obj}
            \\ \cmidrule(l){2-4} \cmidrule(l){5-7}
            & Seen & Unseen & h-IoU & Seen & Unseen & h-IoU  \\ \midrule
            \multirow{1}{*}{ZSSeg+ \cite{xu2022simple_ZSSeg}}                    &  \underline{32.20}  &   0.89  &   1.74  &  43.19  &  27.84  &  33.85 \\
            CLIPSeg \cite{luddecke2022image_CLIPSeg,wei2024ov_OV_PARTS}          &  3.14  &  0.55  &  0.93  &  38.15  &  30.92  &  34.15 \\
            \multirow{1}{*}{CAT-Seg \cite{cho2023cat_CATSeg,wei2024ov_OV_PARTS}} &  7.02  &  2.36  &  3.53  &  33.80  &  25.93  &  29.34 \\ 
            PartGLEE \cite{li2024partglee}                                       &  -  &  -  &  -  &  \underline{51.29}  &  35.33  &  \underline{41.83} \\
            PartCLIPSeg \cite{PartCLIPSeg2024}                                   & 14.15 & \underline{9.52} & \underline{11.38} & 38.37 & \underline{38.82} & 38.60 \\
            \midrule
            PartCATSeg (Ours)                                                    & \textbf{38.87} & \textbf{17.56} & \textbf{24.19} & \textbf{53.13} & \textbf{47.16} & \textbf{49.96} \\
            \vspace{-4pt}                                                            &  &  &  \gainp{+12.81}  &  &  &  \gainp{+8.13} \\

            \bottomrule
        \end{tabular}
    }
    \vspace{-1.0em}
    \caption{Comparison of zero-shot performance with state-of-the-art methods on ADE20K-Part-234.}
    \label{tab:main_ade}
    \vspace{-1.0em}
\end{table}

\begin{table}[t!]
    \centering
    \small
    \resizebox{\linewidth}{!}
    {
        \begin{tabular}{@{}l ccc ccc @{}}
            \toprule
            \multicolumn{1}{c}{\multirow{2}{*}{Method}}                   & \multicolumn{3}{c}{Pred-All} & \multicolumn{3}{c}{Oracle-Obj}
            \\ \cmidrule(l){2-4} \cmidrule(l){5-7} 
            & Seen & Unseen & h-IoU & Seen & Unseen & h-IoU  \\ \midrule
            CLIPSeg \cite{luddecke2022image_CLIPSeg,wei2024ov_OV_PARTS}   &  32.39  &  12.27  &  17.80  &  53.91  &  37.17  &  44.00 \\
            CAT-Seg \cite{cho2023cat_CATSeg,wei2024ov_OV_PARTS}           &  33.58  &  \underline{23.04}  &  \underline{27.33}  &  47.34  &  35.14  &  40.33 \\
            PartCLIPSeg \cite{PartCLIPSeg2024}                            & \underline{38.82} & 19.47 & 25.94 & \underline{56.26} & \underline{51.65} & \underline{53.85} \\
            \midrule
            PartCATSeg (Ours)                                             & \textbf{57.33} & \textbf{53.07} & \textbf{55.12} & \textbf{73.83} & \textbf{71.52} & \textbf{72.66} \\
            \vspace{-4pt}                                                            &  &  &  \gainp{+27.79}  &  &  &  \gainp{+18.81} \\

            \bottomrule
        \end{tabular}
    }
    \vspace{-1.0em}
    \caption{
        Comparison of zero-shot performance with state-of-the-art methods on PartImageNet.
    }
    \label{tab:main_partimagenet}
    \vspace{-2.0em}
\end{table}

\noindent \textbf{Cross-Dataset Part Segmentation.}
Our experiments evaluate the performance of our method in two cross-dataset settings: (1) PartImageNet (OOD) and (2) PartImageNet $\rightarrow$ Pascal-Part-116. 
PartImageNet (OOD) includes classes with complex, irregular part shapes that differ from the training data, for evaluating adaptability to novel configurations within a familiar structure. As shown in \Cref{tab:main_cross_dataset}, PartCATSeg outperforms previous methods in both Pred-All and Oracle-Obj settings, demonstrating its robustness in handling unseen categories in an OOD context. On PartImageNet $\rightarrow$ Pascal-Part-116, we evaluate the model's ability to transfer to a different dataset, Pascal-Part-116 \cite{chen2014detect_PascalPart,wei2024ov_OV_PARTS}. 
Trained on PartImageNet, our model is evaluated on Pascal-Part-116. Herein, PartCATSeg again achieves the highest performance across both Pred-All and Oracle-Obj settings. 
These results confirm PartCATSeg’s effectiveness in diverse cross-dataset scenarios, highlighting its versatility and robustness in part segmentation tasks across various dataset configurations.

\begin{table}[ht]
    \centering
    \small
    \resizebox{\linewidth}{!}
    {
        \begin{tabular}{@{}l cc cc @{}}
            \toprule
            \multicolumn{1}{c}{\multirow{3}{*}{Method}}                   & \multicolumn{2}{c}{PartImageNet (OOD)} & \multicolumn{2}{c}{\scriptsize PartImageNet $\rightarrow$ Pascal-Part-116} 
            \\ \cmidrule(l){2-3} \cmidrule(l){4-5} 
            & \multicolumn{1}{c}{Pred-All} & \multicolumn{1}{c}{Oracle-Obj} & \multicolumn{1}{c}{Pred-All} & \multicolumn{1}{c}{Oracle-Obj}
            \\ \cmidrule(l){2-2} \cmidrule(l){3-3} \cmidrule(l){4-4} \cmidrule(l){5-5} 
            & Unseen & Unseen & Unseen & Unseen  \\ \midrule
            CLIPSeg \cite{luddecke2022image_CLIPSeg,wei2024ov_OV_PARTS}   &  6.69  &  55.86  &  11.72  &  14.87 \\
            CAT-Seg \cite{cho2023cat_CATSeg,wei2024ov_OV_PARTS}           &  \underline{19.83}  &  39.12  &  11.55  &  12.50 \\
            PartCLIPSeg \cite{PartCLIPSeg2024}                            & 10.33   & \underline{59.16} & \underline{14.74} & \underline{19.86} \\
            \midrule
            PartCATSeg (Ours)                                             & \textbf{40.17} & \textbf{66.15} & \textbf{18.21} & \textbf{22.88} \\
            \vspace{-4pt} & \gainp{+20.34} & \gainp{+6.99} & \gainp{+3.47} & \gainp{+3.02} \\

            \bottomrule
        \end{tabular}
    }
    \vspace{-1.0em}
    \caption{Cross-dataset performance with PartImageNet (OOD) and PartImageNet to Pascal-Part-116.}
    \label{tab:main_cross_dataset}
\end{table}

\begin{table}[ht]
\centering
\small
\resizebox{\linewidth}{!}
{
\begin{tabular}{lcccccc}
\toprule
\multicolumn{1}{c}{\multirow{2}{*}{\makecell{Compositional\\Loss}}} & \multicolumn{3}{c}{Pred-All} & \multicolumn{3}{c}{Oracle-Obj} \\ \cmidrule(l){2-4} \cmidrule(l){5-7} 
                        & Seen & Unseen & h-IoU       & Seen  & Unseen & h-IoU       \\ \midrule
Cost Agg                              & 44.05 & 25.06 & 31.94      & 56.40 & 28.26  & 37.66      \\
+ DINO & \underline{54.48} & 35.89 & 43.28      & \underline{60.56} & 40.51  & 48.55      \\
+ DINO + $\mathcal{L}_{\texttt{comp-L1}}$   & \textbf{54.70} & \underline{35.92} & \underline{43.36}      & \textbf{61.29} & \underline{41.62}  & \underline{49.57}      \\
+ DINO + $\mathcal{L}_{\texttt{comp-SM}}$                       & 52.62 & \textbf{40.51} & \textbf{45.77}      & 57.49 & \textbf{44.88}  & \textbf{50.41}      \\ 
\bottomrule
\end{tabular}
}
\vspace{-1.0em}
\caption{Impact of Compositional Loss on Pascal-Part-116}
\label{tab:ablation_loss}
\end{table}

\begin{table}[ht]
\centering
\small
\resizebox{\linewidth}{!}
{
\begin{threeparttable}
\begin{tabular}{lcccccc}
\toprule
\multicolumn{1}{c}{\multirow{2}{*}{\makecell{Structural\\Guidance}}} & \multicolumn{3}{c}{Pred-All} & \multicolumn{3}{c}{Oracle-Obj} \\ \cmidrule(l){2-4} \cmidrule(l){5-7} 
                        & Seen & Unseen & h-IoU       & Seen  & Unseen & h-IoU       \\ \midrule
w/o Structural Guidance  & 42.29 & 27.94 & 33.65      & 46.44 & 31.59  & 37.60      \\
DINO ($\mathcal{T}^{\texttt{SA}}_{\texttt{Obj}}$) & 50.24 & 31.35 & 38.61      & 55.03 & 34.29  & 42.25      \\
    DINO ($\mathcal{T}^{\texttt{SA}}_{\texttt{Part}}$)   & \textbf{56.28} & \underline{36.67} & \underline{44.41}      & \textbf{62.73} & \underline{43.47}  & \textbf{51.35}      \\
DINO ($\mathcal{T}^{\texttt{SA}}_{\texttt{Obj}}$, $\mathcal{T}^{\texttt{SA}}_{\texttt{Part}}$)                       & \underline{52.62} & \textbf{40.51} & \textbf{45.77}      & \underline{57.49} & \textbf{44.88}  & \underline{50.41}      \\ 
\bottomrule
\end{tabular}

\begin{tablenotes}
    \item[1] All configurations use the compositional loss $\mathcal{L}_{\text{comp}}$.
    \item[2] Configurations utilizing DINO also include object-specific part-level guidance.
\end{tablenotes}

\end{threeparttable}
}
\vspace{-1.0em}

\caption{Impact of Structural Guidance on Pascal-Part-116.}
\vspace{-1.0em}
\vspace{-1.0em}
\label{tab:ablation_dino}
\end{table}

\begin{figure*}[!t]
    \centering
    \begin{subfigure}[t]{0.160\textwidth} \includegraphics[width=\textwidth]{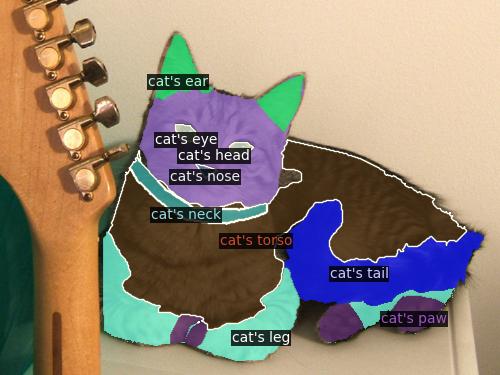} \end{subfigure}
    \begin{subfigure}[t]{0.160\textwidth} \includegraphics[width=\textwidth]{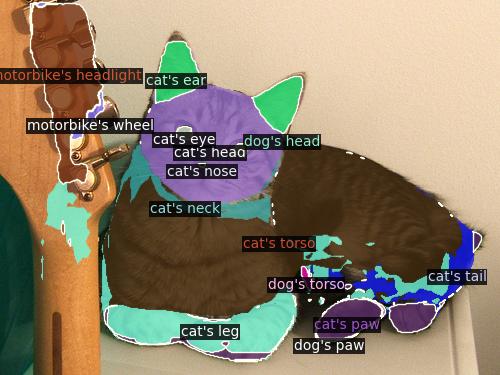} \end{subfigure}
    \begin{subfigure}[t]{0.160\textwidth} \includegraphics[width=\textwidth]{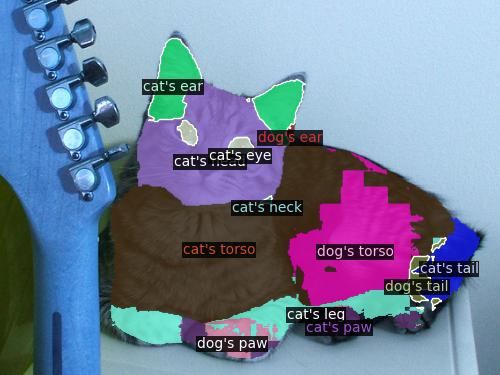} \end{subfigure}
    \begin{subfigure}[t]{0.160\textwidth} \includegraphics[width=\textwidth]{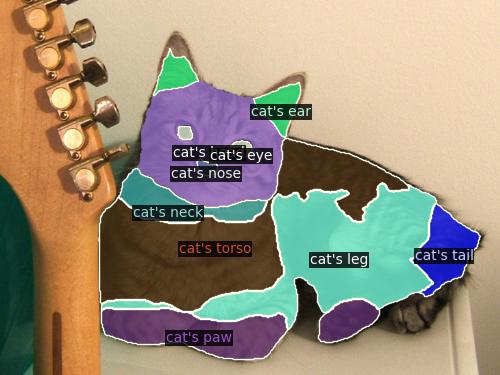} \end{subfigure}
    \begin{subfigure}[t]{0.160\textwidth} \includegraphics[width=\textwidth]{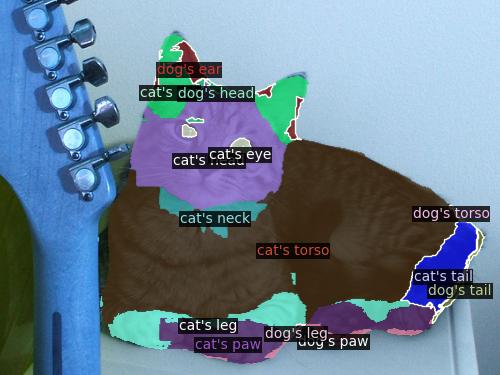} \end{subfigure}
    \begin{subfigure}[t]{0.160\textwidth} \includegraphics[width=\textwidth]{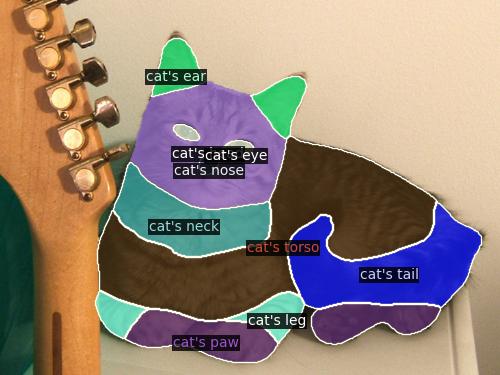} \end{subfigure}
    \begin{subfigure}[t]{0.160\textwidth} \includegraphics[width=\textwidth]{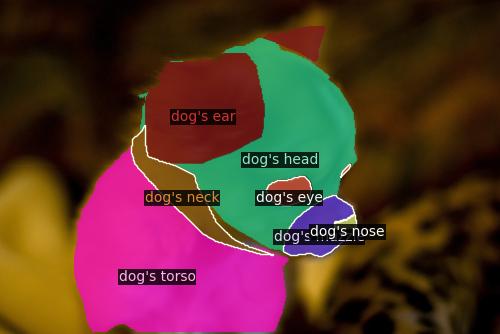} \end{subfigure}
    \begin{subfigure}[t]{0.160\textwidth} \includegraphics[width=\textwidth]{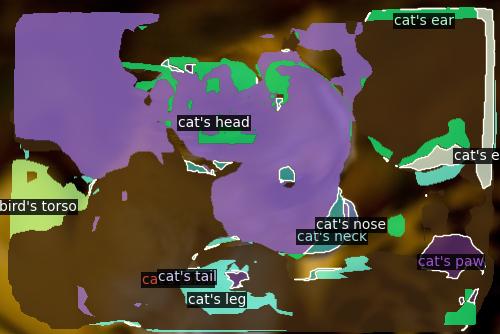} \end{subfigure}
    \begin{subfigure}[t]{0.160\textwidth} \includegraphics[width=\textwidth]{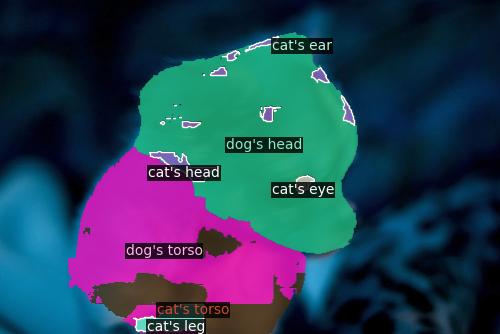} \end{subfigure}
    \begin{subfigure}[t]{0.160\textwidth} \includegraphics[width=\textwidth]{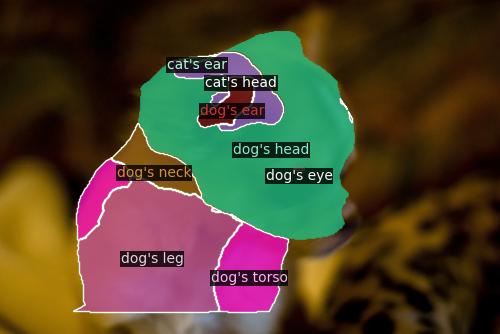} \end{subfigure}
    \begin{subfigure}[t]{0.160\textwidth} \includegraphics[width=\textwidth]{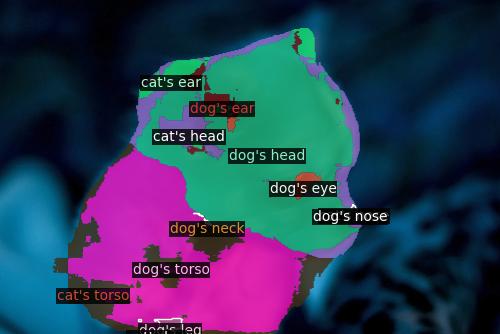} \end{subfigure}
    \begin{subfigure}[t]{0.160\textwidth} \includegraphics[width=\textwidth]{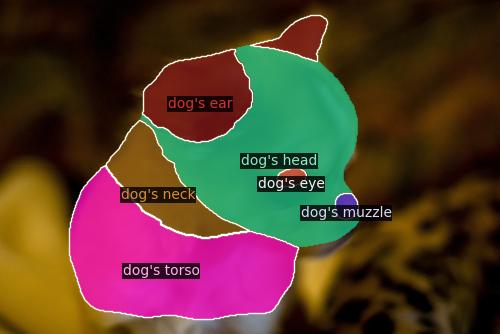} \end{subfigure}
    
    \begin{subfigure}[t]{0.160\textwidth}
        \includegraphics[width=\textwidth, trim=0 0 0 20, clip]{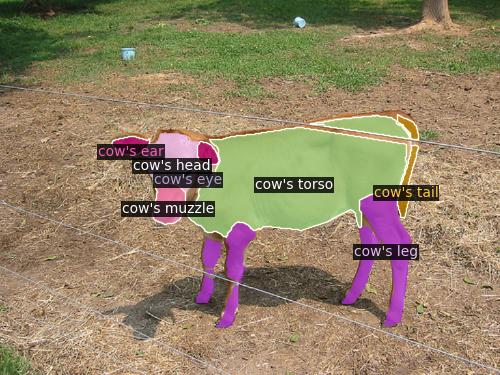}
    \end{subfigure}
    \begin{subfigure}[t]{0.160\textwidth}
        \includegraphics[width=\textwidth, trim=0 0 0 20, clip]{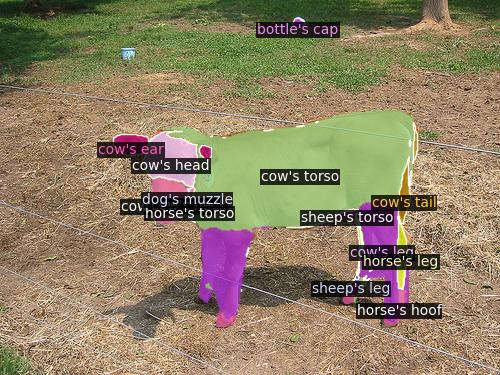}
    \end{subfigure}
    \begin{subfigure}[t]{0.160\textwidth}
        \includegraphics[width=\textwidth, trim=0 0 0 20, clip]{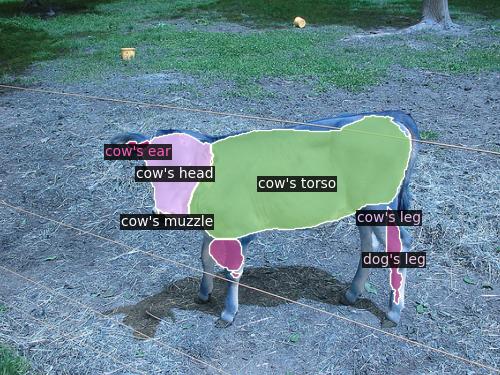}
    \end{subfigure}
    \begin{subfigure}[t]{0.160\textwidth}
        \includegraphics[width=\textwidth, trim=0 0 0 20, clip]{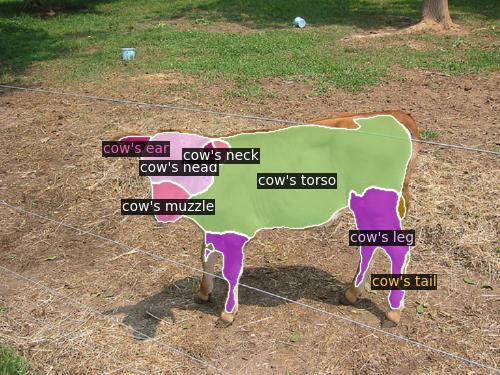}
    \end{subfigure}
    \begin{subfigure}[t]{0.160\textwidth}
        \includegraphics[width=\textwidth, trim=0 0 0 20, clip]{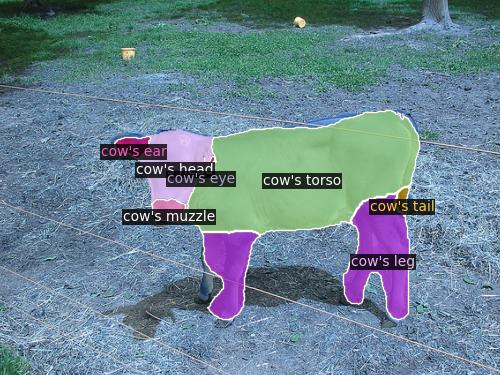}
    \end{subfigure}
    \begin{subfigure}[t]{0.160\textwidth}
        \includegraphics[width=\textwidth, trim=0 0 0 20, clip]{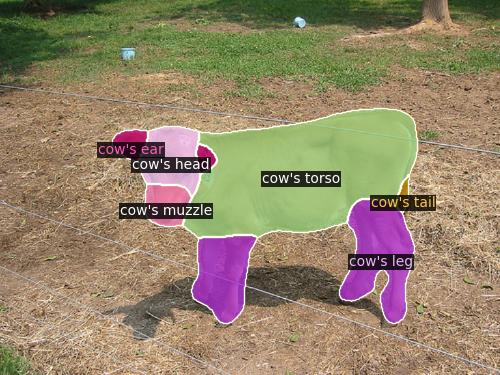}
    \end{subfigure}
    
    \begin{subfigure}[t]{0.160\textwidth} \includegraphics[width=\textwidth]{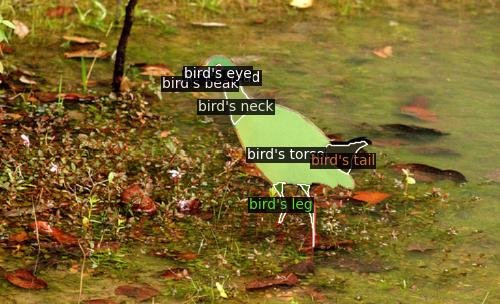} \end{subfigure}
    \begin{subfigure}[t]{0.160\textwidth} \includegraphics[width=\textwidth]{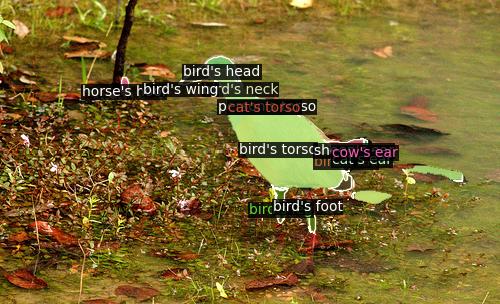} \end{subfigure}
    \begin{subfigure}[t]{0.160\textwidth} \includegraphics[width=\textwidth]{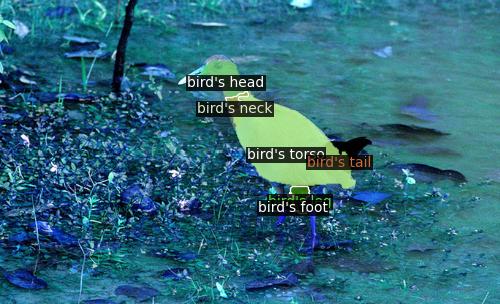} \end{subfigure}
    \begin{subfigure}[t]{0.160\textwidth} \includegraphics[width=\textwidth]{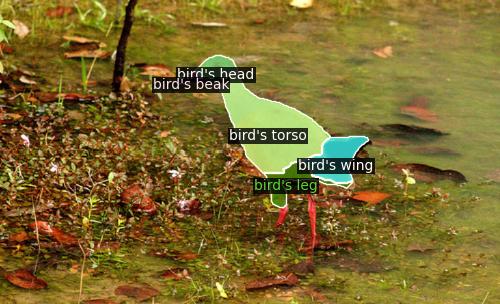} \end{subfigure}
    \begin{subfigure}[t]{0.160\textwidth} \includegraphics[width=\textwidth]{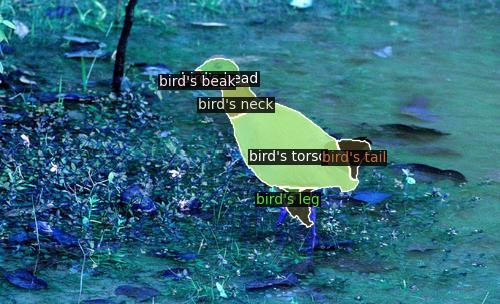} \end{subfigure}
    \begin{subfigure}[t]{0.160\textwidth} \includegraphics[width=\textwidth]{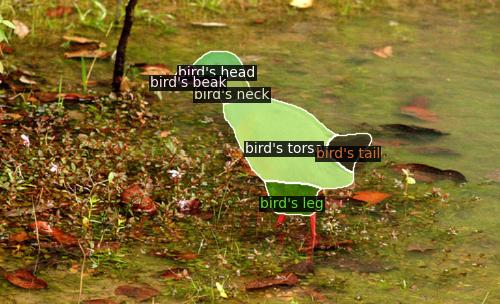} \end{subfigure}

    \begin{subfigure}[t]{0.160\textwidth}
        \includegraphics[width=\textwidth, trim=0 0 0 50, clip]{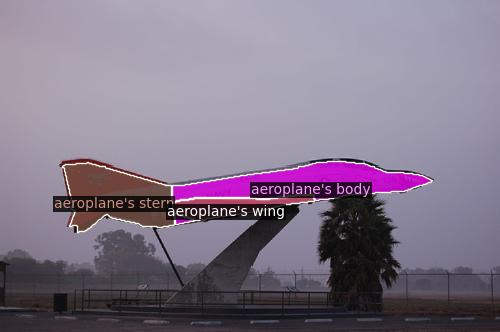}
    \end{subfigure}
    \begin{subfigure}[t]{0.160\textwidth}
        \includegraphics[width=\textwidth, trim=0 0 0 50, clip]{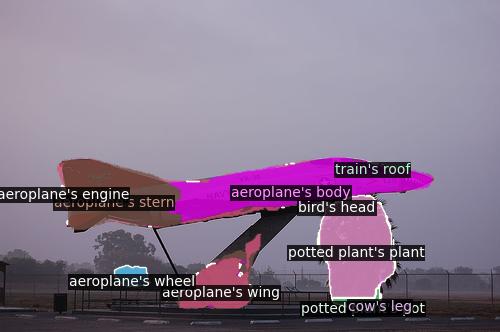}
    \end{subfigure}
    \begin{subfigure}[t]{0.160\textwidth}
        \includegraphics[width=\textwidth, trim=0 0 0 50, clip]{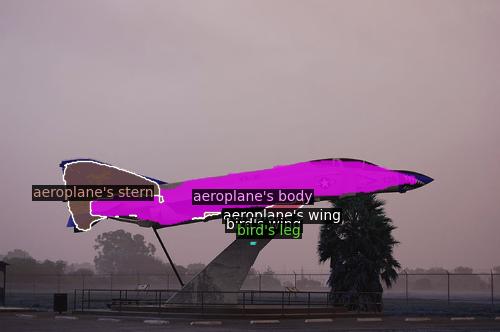}
    \end{subfigure}
    \begin{subfigure}[t]{0.160\textwidth}
        \includegraphics[width=\textwidth, trim=0 0 0 50, clip]{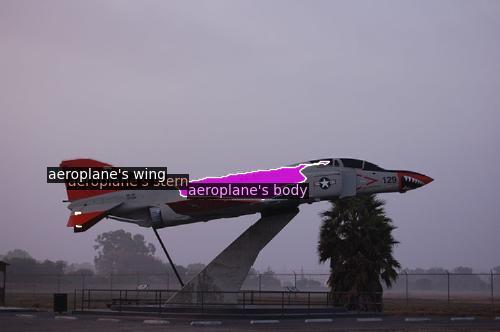}
    \end{subfigure}
    \begin{subfigure}[t]{0.160\textwidth}
        \includegraphics[width=\textwidth, trim=0 0 0 50, clip]{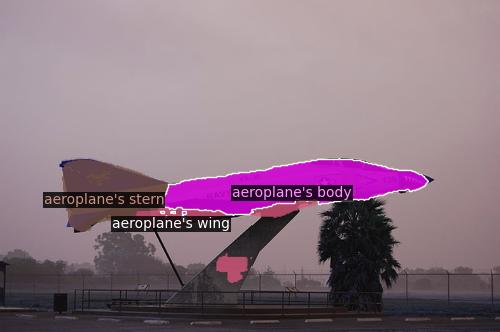}
    \end{subfigure}
    \begin{subfigure}[t]{0.160\textwidth}
        \includegraphics[width=\textwidth, trim=0 0 0 50, clip]{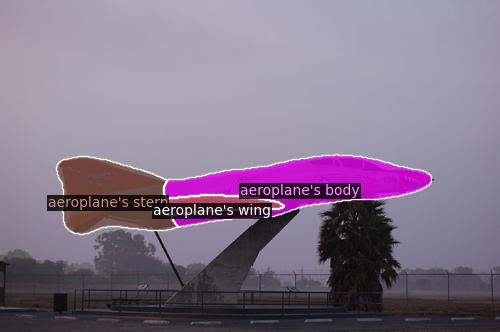}
    \end{subfigure}

    \begin{subfigure}[t]{0.160\textwidth} \includegraphics[width=\textwidth]{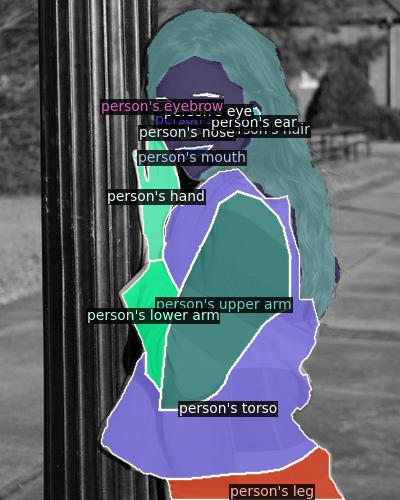} \end{subfigure}
    \begin{subfigure}[t]{0.160\textwidth} \includegraphics[width=\textwidth]{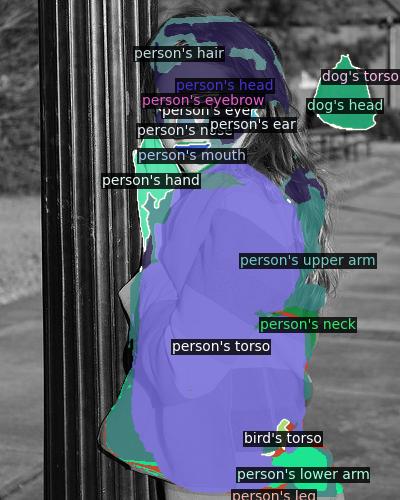} \end{subfigure}
    \begin{subfigure}[t]{0.160\textwidth} \includegraphics[width=\textwidth]{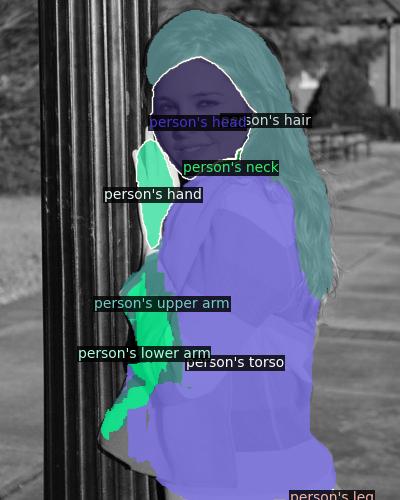} \end{subfigure}
    \begin{subfigure}[t]{0.160\textwidth} \includegraphics[width=\textwidth]{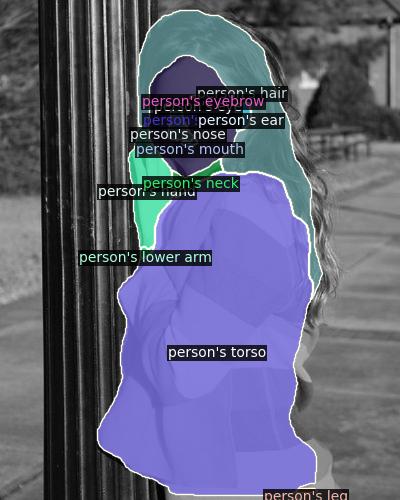} \end{subfigure}
    \begin{subfigure}[t]{0.160\textwidth} \includegraphics[width=\textwidth]{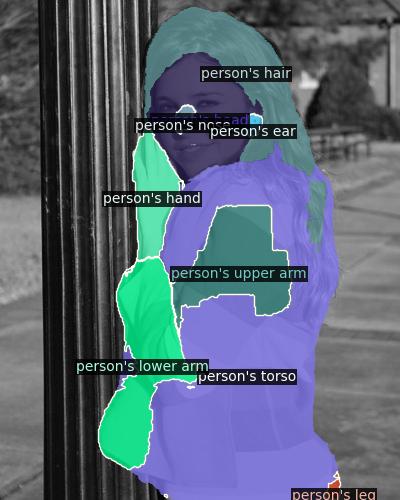} \end{subfigure}
    \begin{subfigure}[t]{0.160\textwidth} \includegraphics[width=\textwidth]{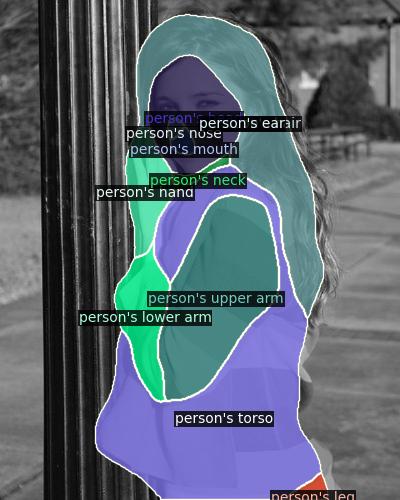} \end{subfigure}
    
    \vspace{-1em}

    \begin{subfigure}[t]{0.160\textwidth}
        \caption{Ground-truth}
    \end{subfigure}
    \begin{subfigure}[t]{0.160\textwidth}
        \caption{VLPart~\cite{sun2023going_VLPart}}
    \end{subfigure}
    \begin{subfigure}[t]{0.160\textwidth}
        \caption{CLIPSeg~\cite{radford2021learning_CLIP,wei2024ov_OV_PARTS}}
    \end{subfigure}
    \begin{subfigure}[t]{0.160\textwidth}
        \caption{CAT-Seg~\cite{cho2023cat_CATSeg,wei2024ov_OV_PARTS}}
    \end{subfigure}
    \begin{subfigure}[t]{0.160\textwidth}
        \caption{PartCLIPSeg~\cite{PartCLIPSeg2024}}
    \end{subfigure}
    \begin{subfigure}[t]{0.160\textwidth}
        \caption{PartCATSeg}
    \end{subfigure}
    \caption{
        Qualitative evaluation of zero-shot part segmentation on Pascal-Part-116 in the \textbf{Pred-All} configuration.
        Note that annotations for unseen categories (e.g., bird, cow, dog) are excluded from the training set.
    }
    \label{fig:vis_pred_all_qualitative}
    \vspace{-1.5em}
\end{figure*}

\noindent \textbf{Qualitative Evaluation.} We present qualitative results on the Pascal-Part-116 dataset in the Pred-All setting. 
\Cref{fig:vis_pred_all_qualitative} compares our method with other baselines, demonstrating significant improvements by addressing several challenges in previous OVPS methods. For instance, in \textbf{rows 1 and 2}, compared to the other baselines showing unnatural segmentations, our model successfully segments the cat's leg and the dog's neck, preserving the object's structural integrity with the structural guidance. In \textbf{row 6}, PartCATSeg correctly segments the person's upper and lower arm as distinct parts.
Additionally, our method almost completely segments the parts in the object without leaving regions unsegmented and effectively captures smaller parts that other models often miss. For example, in \textbf{rows 2 -- 4}, finer details like the dog's ear, the cow's tail, and the bird's beak are accurately segmented in our method. In \textbf{rows 5 and 6}, our method segments almost all the parts in the object without sacrificing accuracy compared to the other baselines. This improvement is largely due to the compositional loss, which encourages the competitive assignment of parts that collectively constitute the whole object.
Moreover, our method resolves confusion between similar object categories. In \textbf{rows 1 and 2}, other models misclassify parts of the cat and dog, sometimes confusing the cat for a dog and vice versa. Our object-aware part-level guidance helps the model correctly segment the parts of the correct object.


\subsection{Ablation Study}
\label{subsec:Ablation Study}

\noindent \textbf{Impact of Compositional Loss.}
To assess the effectiveness of the proposed compositional loss $\mathcal{L}_{\texttt{comp}}$, we conduct an ablation study on its impact on Pascal-Part-116 in~\Cref{tab:ablation_loss}. We compare the effects of using compositional loss with a softmax-based normalization ($\mathcal{L}_{\texttt{comp-SM}}$), which encourages each spatial location to be predominantly assigned to a single part class, versus L1 normalization ($\mathcal{L}_{\texttt{comp-L1}}$) before combining part costs. We observe that adding $\mathcal{L}{\texttt{comp}}$ gains about 2\% h-IoU increase in the Pred-All setting compared to the model without using  $\mathcal{L}{\texttt{comp}}$.
Furthermore, the model using $\mathcal{L}_{\texttt{comp-SM}}$ outperforms the one using $\mathcal{L}_{\texttt{comp-L1}}$.
These results confirm that $\mathcal{L}_{\texttt{comp}}$ effectively improves performance by leveraging the inductive bias that parts collectively compose the object. Further qualitative results of ablation on $\mathcal{L}_{\texttt{comp}}$ are provided in the supplementary materials.




\noindent \textbf{Impact of Structural Guidance from DINO.} To investigate the effect of structural guidance from DINO, we conduct an ablation study by selectively applying it in our cost aggregation framework. 
In PartCATSeg, structural guidance from DINO is incorporated into three levels of spatial aggregation transformer modules: objects ($\mathcal{T}^{\texttt{SA}}_{\texttt{Obj}}$), parts ($\mathcal{T}^{\texttt{SA}}_{\texttt{Part}}$), and object-specific parts ($\mathcal{T}^{\texttt{SA}}_{\texttt{Obj-Part}}$). We measure performance changes on Pascal-Part-116 when structural guidance is applied only at the object level, only at the part level, and at both levels. As shown in \Cref{tab:ablation_dino}, applying structural guidance at the part level yields more h-IoU increases in both Pred-All and Oracle-Obj settings than at the object level. When structural guidance is applied at both levels, we observe further improvements, especially in unseen classes. This suggests that, within the cost aggregation framework, structural guidance from DINO is more effective when leveraging detailed structural information within individual objects, rather than relying solely on DINO's background-object separation ability.

\section{Conclusion}
\label{sec:conclusion}


In this paper, we present a novel approach to open-vocabulary part segmentation, PartCATSeg, which leverages cost aggregation to address the inherent challenges of OVPS.
We make strategic modifications to cost aggregation to enhance object-aware part-level guidance, incorporate a compositional loss to compensate for limited part-level guidance from the base set, and utilize DINO features for robust structural guidance.
These combined efforts enable PartCATSeg to significantly outperform existing state-of-the-art methods.
We believe that our approach can serve as a powerful new baseline for practitioners and researchers, fostering greater interest and advancement within the community.

\section*{Acknowledgements}
\label{sec:Acknowledgements}


\noindent
This research was supported by the Basic Science Research Program through the National Research Foundation of Korea (NRF) funded by the MSIP (RS-2025-00520207, RS-2023-00219019), KEIT grant funded by the Korean government (MOTIE) (No. 2022-0-00680, No. 2022-0-01045), the IITP grant funded by the Korean government (MSIT) (No. RS-2024-00457882, National AI Research Lab Project, No. 2021-0-02068 Artificial Intelligence Innovation Hub, RS-2019-II190075 Artificial Intelligence Graduate School Program (KAIST)), Samsung Electronics Co., Ltd (IO230508-06190-01) and SAMSUNG Research, Samsung Electronics Co., Ltd.


{
    \small
    \bibliographystyle{ieeenat_fullname}
    \bibliography{main}
}


\clearpage
\newpage
\setcounter{page}{1}
\appendix
\setcounter{enumiv}{0} 



\startcontents[supplement] 
\printcontents[supplement]{}{1}{\section*{\contentsname}}  

\setcounter{section}{0}

\setcounter{figure}{0}
\renewcommand{\thetable}{A\arabic{table}}

\setcounter{table}{0}
\renewcommand{\thefigure}{A\arabic{figure}}

\setcounter{equation}{0}
\renewcommand{\theequation}{A\arabic{equation}}





\newpage

\section{Limitations \& Future Work}

Our PartCATSeg framework has shown strong performance but also has some limitations and potential directions for future work.
Specifically, it struggles with handling extremely fine-grained part distinctions, where subtle differences between parts are challenging to capture accurately.
Additionally, the framework is designed based on semantic segmentation, as it continues to demonstrate superior performance compared to instance segmentation in many scenarios.
However, this design choice inherently limits the ability to distinguish individual parts as separate instances.
For example, distinguishing between the left and right handles of a bicycle requires an instance-level understanding that the current framework lacks.
Achieving this would necessitate the generation or inclusion of instance object masks, which are not currently supported.
Expanding the framework to incorporate instance segmentation could address this limitation, enabling more precise and versatile part segmentation for applications that demand detailed instance-level information.





Future work aims to address these limitations by improving fine-grained differentiation through advanced attention mechanisms~\cite{liu2021swin,wang2021pyramid,ham20203_a3} and adaptive structural priors tailored to specific datasets.
Enhancing the framework's ability to distinguish subtle part-level differences could significantly improve its performance in complex scenarios.
Additionally, integrating the framework with off-the-shelf open-vocabulary instance segmentation modules offers a promising solution for overcoming the inability to individual parts as separate instances.
This integration could enable the model to assign unique labels to similar parts across different instances, further extending its applicability and effectiveness.

\section{Experimental Details}

\subsection{Code \& Reproduction}

Details can be found in the publicly available code.
For additional details, refer to the GitHub repository available at \href{https://github.com/kaist-cvml/part-catseg}{https://github.com/kaist-cvml/part-catseg}

\subsection{Device Information}

All experiments were conducted using eight NVIDIA A6000 GPUs and PyTorch 2.2 for training and evaluation.

\subsection{Implementation Details}

Our model is based on CAT-Seg~\cite{cho2023cat_CATSeg}, a state-of-the-art open-vocabulary semantic segmentation (OVSS) method, redefined to suit open-vocabulary part segmentation (OVPS) tasks in OV-PARTS~\cite{wei2024ov_OV_PARTS}.
It employs a CLIP~\cite{radford2021learning_CLIP} encoder built on CLIP ViT-B/16 and leverages DINOv2~\cite{oquab2023dinov2}, a pre-trained model, for structural guidance.

We begin by utilizing the pre-trained object-level OVSS models from CAT-Seg and fine-tune them with the datasets described in~\Cref{suppl_sec:datasets_details}. The model undergoes training with the AdamW~\cite{loshchilov2017decoupled_adamw} optimizer, starting with an initial learning rate of 0.0001, over 20,000 iterations, and a batch size of 8.
During training, model checkpoints are saved every 1,000 iterations.
The final model is selected based on the highest validation performance.
For instance, the best validation score on the Pascal-Part-116 dataset in the Oracle-Obj setting comes from the checkpoint saved at 12,000 iterations.

\subsection{Evaluation Details}

For the evaluation protocol, the Pred-All setup of PartCLIPSeg~\cite{PartCLIPSeg2024} and the Oracle-Obj setup of OV-PARTS~\cite{wei2024ov_OV_PARTS} were utilized.
The Pred-All setup assumes a more challenging scenario in which predictions are made without any prior information.
In contrast, the Oracle-Obj setup assumes the availability of object-level masks.
As noted in OV-PARTS, the Oracle-Obj setup simulates results achievable when using off-the-shelf open-vocabulary semantic segmentation models.

\subsection{Hyperparameters}

The model architecture incorporates layers and parameters from CAT-Seg~\cite{cho2023cat_CATSeg}. Furthermore, as outlined in~\Cref{eq:final_loss}, our method defines three key hyperparameters, $\lambda_{\texttt{Obj}}$, $\lambda_{\texttt{Part}}$, and $\lambda_{\texttt{comp}}$, which are associated with two primary loss functions: the disentanglement loss $\mathcal{L}_{\texttt{disen}}$ and the compositional loss $\mathcal{L}_{\texttt{comp}}$. These lambda parameters were fine-tuned through experimental validation on the training set to balance the contributions of the proposed loss functions. The final values were determined as $\lambda_{\texttt{Obj}} = 1.0$, $\lambda_{\texttt{Part}} = 1.0$, and $\lambda_{\texttt{comp}} = 1.0$.

\subsection{Baselines}

\begin{itemize}
    \item ZSSeg+ \cite{xu2022simple_ZSSeg,wei2024ov_OV_PARTS}: ZSSeg is a two-stage framework for open-vocabulary semantic segmentation that uses CLIP to classify class-agnostic mask proposals, enabling segmentation of seen and unseen classes. ZSSeg+ extends ZSSeg to support part-level segmentation. We evaluate ZSSeg+ using a ResNet-50 \cite{he2016deep_resnet} baseline, fine-tuned with Compositional Prompt Tuning based on CoOp \cite{zhou2022learning_cptcoop}. 
    \item VLPart \cite{sun2023going_VLPart}: VLPart enables open-vocabulary part segmentation by training on data across multiple granularities (part-level, object-level, and image-level) and segments novel objects into parts through dense correspondences with base objects.
    \item CLIPSeg \cite{luddecke2022image_CLIPSeg,wei2024ov_OV_PARTS}: CLIPSeg extends CLIP for segmentation tasks, using a transformer-based decoder to generate segmentation maps conditioned on text or image prompts, supporting tasks like referring expression segmentation and zero-shot segmentation. For evaluation, we fine-tune the FiLM layer, decoder, visual encoder, and language embedding layer in the text encoder, following the approach in~\cite{wei2024ov_OV_PARTS}. 
    \item CAT-Seg \cite{cho2023cat_CATSeg,wei2024ov_OV_PARTS}: CAT-Seg adapts vision-language models like CLIP by aggregating cosine similarity between image and text embeddings to create cost volumes, enabling segmentation of seen and unseen classes.
    Additionally, CAT-Seg proposes learning the self-attention heads of CLIP’s encoders, achieving effective results.
    \item PartGLEE \cite{li2024partglee}: PartGLEE is a part-level segmentation model that uses a unified framework and the Q-Former to model hierarchical relationships between objects and parts, allowing segmentation at any granularity in open-world scenarios.
    \item PartCLIPSeg \cite{PartCLIPSeg2024}: PartCLIPSeg leverages generalized parts and object-level contexts to enhance fine-grained part segmentation, incorporating competitive part relationships and attention mechanisms to improve segmentation accuracy and generalization to unseen vocabularies.
\end{itemize}



\section{Additional Quantitative Evaluation}

\subsection{Evaluation on Recall}

We evaluated various baselines across multiple datasets, as detailed in~\Cref{tab:main_pascal},~\Cref{tab:main_ade}, and~\Cref{tab:main_partimagenet}. Here, we provide an additional analysis of recall performance in zero-shot evaluation on Pascal-Part-116 and ADE20K-Part-234. The recall metric measures a model's ability to correctly identify less frequent or smaller objects, which are often more difficult to segment accurately. As shown in~\Cref{suppl_tab:main_recall}, our method achieves state-of-the-art recall scores, highlighting its superior capability in identifying and segmenting these challenging parts.

\begin{table}[ht]
    \centering
    \small
    \resizebox{\linewidth}{!}
    {

    \begin{threeparttable}
        \begin{tabular}{@{}l ccc ccc@{}}
            \toprule
            \multicolumn{1}{c}{\multirow{2}{*}{Method}}                              & \multicolumn{3}{c}{Pascal-Part-116} & \multicolumn{3}{c}{ADE20K-Part-234} \\
            \cmidrule(l){2-4} \cmidrule(l){5-7}
            & Seen & Unseen & h-Recall & Seen & Unseen & h-Recall  \\ \midrule
            \multirow{1}{*}{ZSSeg+ \cite{xu2022simple_ZSSeg}}                        &  \underline{65.47}  &   32.13  &   43.10  &  \underline{55.78}  &  40.71  &  47.07 \\
            CLIPSeg \cite{luddecke2022image_CLIPSeg,wei2024ov_OV_PARTS}              &  55.71  &  43.35 &  48.76  &  49.59  &  48.11  &  48.84 \\
            \multirow{1}{*}{CAT-Seg \cite{cho2023cat_CATSeg,wei2024ov_OV_PARTS}}     &  56.00  &  43.20  &  48.77  &  43.48  &  39.87  &  41.60 \\
            PartCLIPSeg \cite{PartCLIPSeg2024}                                       &  {58.46}  &  \underline{47.93}  &  \underline{52.67}  &  53.31  &  \underline{51.52}  &  \underline{52.40} \\
            \midrule
            PartCATSeg (Ours)                                                        &  \textbf{67.15} & \textbf{61.02} & \textbf{63.94} & \textbf{64.81} & \textbf{64.22} & \textbf{64.52} \\
            \vspace{-4pt}                                                            &  &  &  \gainp{+11.27}  &  &  &  \gainp{+12.12} \\
            \bottomrule
        \end{tabular}
        \begin{tablenotes}
            \item[1] The best score is \textbf{bold} and the second-best score is \underline{underlined}.
  
        \end{tablenotes}
        \end{threeparttable}
    }
    \vspace{-1.0em}
    \caption{
        Comparison of zero-shot performance with state-of-the-art methods in terms of \textbf{Recall} for \textbf{Oracle-Obj} setting on Pascal-Part-116.
    }
    \label{suppl_tab:main_recall}        
    \vspace{-1.0em}
\end{table}

\subsection{DINO Structural Guidance}
We confirmed the effectiveness of DINO's structural guidance for PartCATSeg in~\Cref{tab:ablation_dino} of the main text.
Additionally, the supplementary material examines how DINO's structural guidance impacts the original CAT-Seg.
As shown in~\Cref{suppl_tab:main_pascal_catseg_dino}, while DINO's structural guidance proves effective, its performance improvement is relatively modest compared to the proposed PartCATSeg.
This highlights that the proposed framework for disentangling parts is more effective overall.

\begin{table}[ht]
    \centering
    \small
    \resizebox{\linewidth}{!}
    {
    \begin{threeparttable}
        \begin{tabular}{@{}l ccc ccc@{}}
            \toprule
            \multicolumn{1}{c}{\multirow{2}{*}{Method}}                              & \multicolumn{3}{c}{Pred-All} & \multicolumn{3}{c}{Oracle-Obj} \\
            \cmidrule(l){2-4} \cmidrule(l){5-7}
            & Seen & Unseen & h-IoU & Seen & Unseen & h-IoU  \\ \midrule
            \multirow{1}{*}{CAT-Seg }     &  36.80  &  23.39  &  28.60  &  43.81  &  27.66  &  33.91 \\
            CAT-Seg w/ Structural Guidance              &  38.84 & 28.99 & 33.20 & 50.37 & 36.86 & 42.57 \\
            \midrule
            PartCATSeg w/o Structural Guidance  & 42.29 & 27.94 & 33.65      & 46.44 & 31.59  & 37.60      \\
            PartCATSeg                                                        &  \textbf{52.62} & \textbf{40.51} & \textbf{45.77} & \textbf{57.49} & \textbf{44.88} & \textbf{50.41} \\
            \bottomrule
        \end{tabular}
        \begin{tablenotes}
            \item[1] The best score is \textbf{bold}.
        \end{tablenotes}
        \end{threeparttable}
    }
    \vspace{-1.0em}
    \caption{
        Comparison of zero-shot performance with state-of-the-art methods on Pascal-Part-116.
    }
    \label{suppl_tab:main_pascal_catseg_dino}
    \vspace{-1.0em}
\end{table}



\begin{figure*}[!h]
    \centering
    \small
    \begin{subfigure}{0.162\textwidth} \centering \includegraphics[width=1\linewidth, height=1\linewidth]{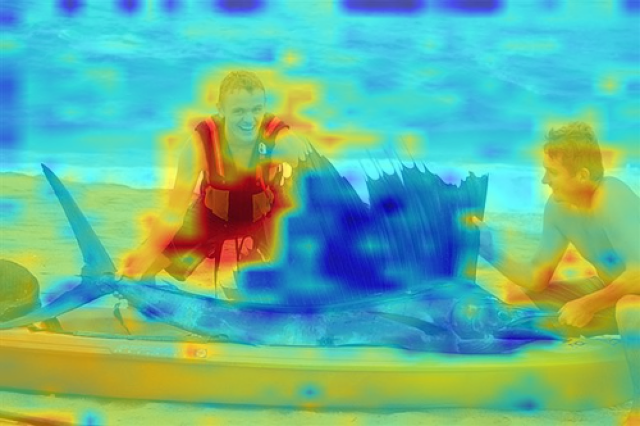} \caption{``person''} \end{subfigure} \hfill
    \begin{subfigure}{0.162\textwidth} \centering \includegraphics[width=1\linewidth, height=1\linewidth]{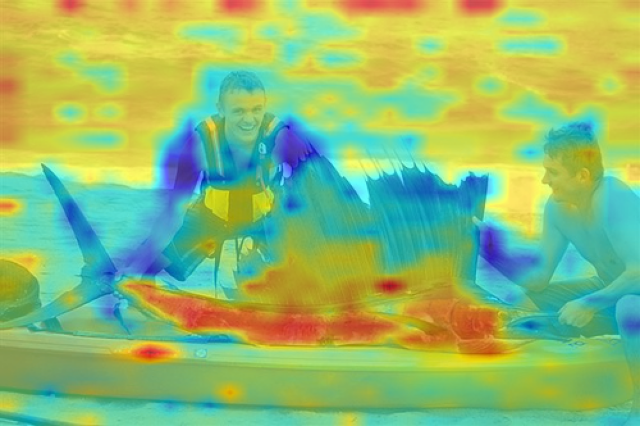} \caption{``person's eye''} \end{subfigure} \hfill
    \begin{subfigure}{0.162\textwidth} \centering \includegraphics[width=1\linewidth, height=1\linewidth]{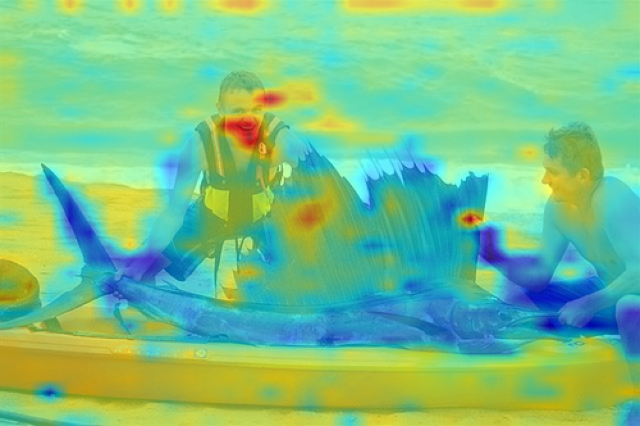} \caption{``person's nose''} \end{subfigure} \hfill
    \begin{subfigure}{0.162\textwidth} \centering \includegraphics[width=1\linewidth, height=1\linewidth]{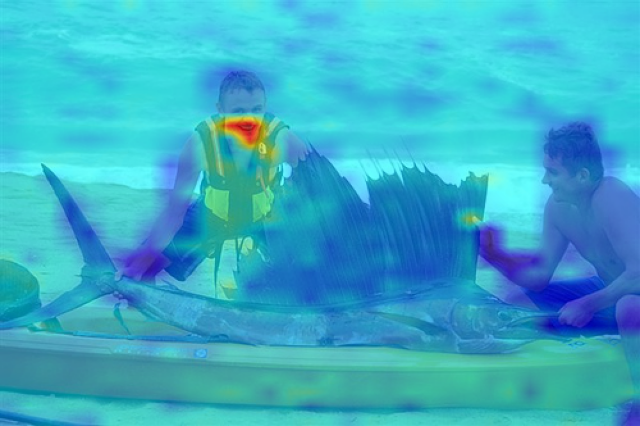} \caption{``person's mouth''} \end{subfigure} \hfill
    \begin{subfigure}{0.162\textwidth} \centering \includegraphics[width=1\linewidth, height=1\linewidth]{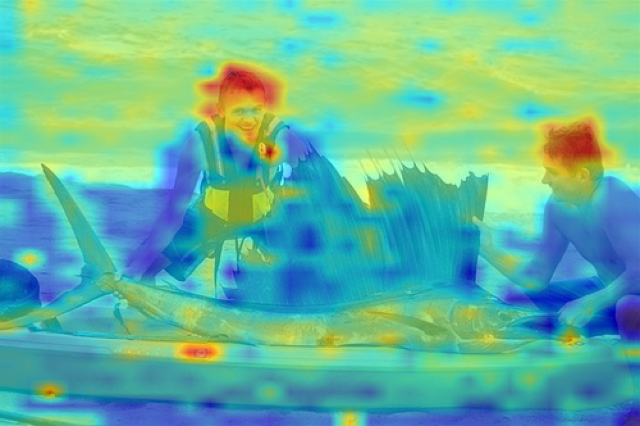} \caption{``person's ear''} \end{subfigure} \hfill
    \begin{subfigure}{0.162\textwidth} \centering \includegraphics[width=1\linewidth, height=1\linewidth]{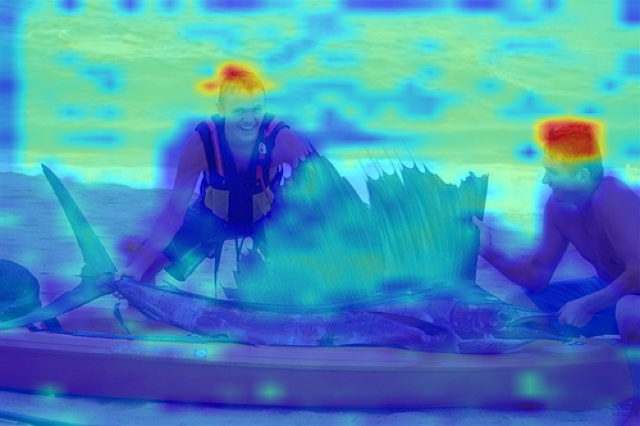} \caption{``person's hair''} \end{subfigure} \hfill
    \begin{subfigure}{0.162\textwidth} \centering \includegraphics[width=1\linewidth, height=1\linewidth]{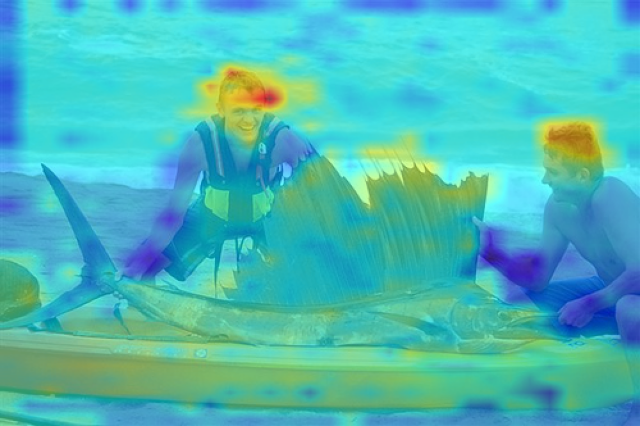} \caption{``person's head''} \end{subfigure} \hfill
    \begin{subfigure}{0.162\textwidth} \centering \includegraphics[width=1\linewidth, height=1\linewidth]{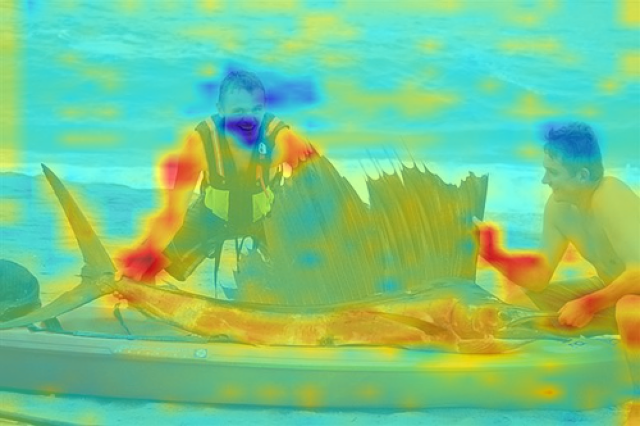} \caption{``person's lower arm''} \end{subfigure} \hfill
    \begin{subfigure}{0.162\textwidth} \centering \includegraphics[width=1\linewidth, height=1\linewidth]{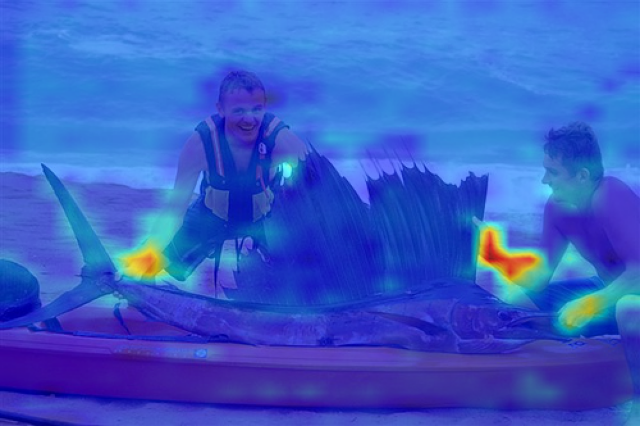} \caption{``person's hand''} \end{subfigure} \hfill
    \begin{subfigure}{0.162\textwidth} \centering \includegraphics[width=1\linewidth, height=1\linewidth]{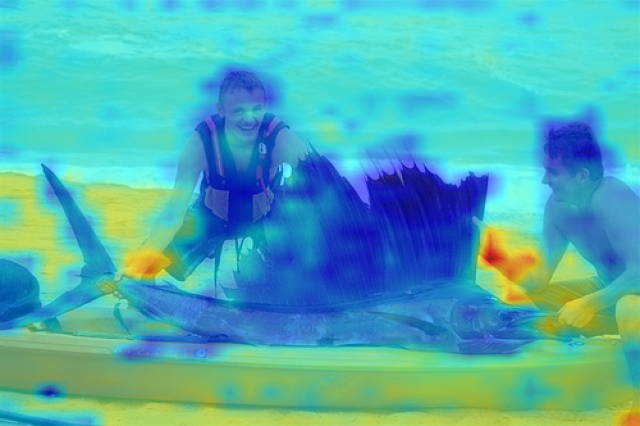} \caption{``person's foot''} \end{subfigure} \hfill
    \begin{subfigure}{0.162\textwidth} \centering \includegraphics[width=1\linewidth, height=1\linewidth]{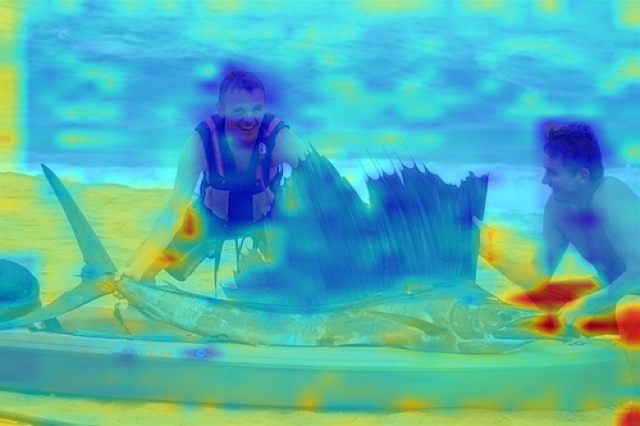} \caption{``person's leg''} \end{subfigure} \hfill
    \begin{subfigure}{0.162\textwidth} \centering \includegraphics[width=1\linewidth, height=1\linewidth]{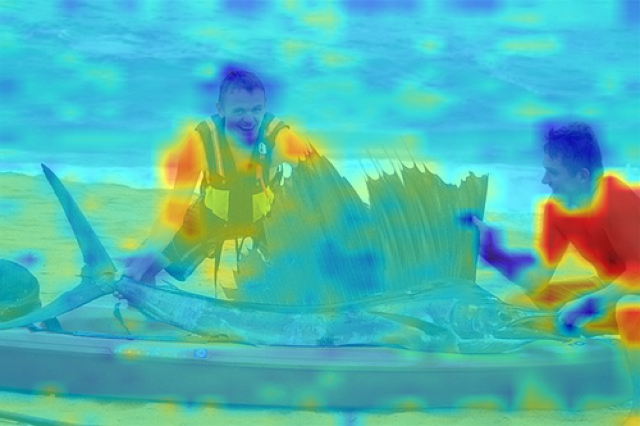} \caption{``person's torso''} \end{subfigure}
    \caption{
        Image-Text Correspondence Visualization \textbf{Before} Training
    }
    \label{suppl_fig:cost_volume_1_person_before_512}
    
\end{figure*}

\begin{figure*}[!h]
    \centering
    \small
    \begin{subfigure}{0.162\textwidth} \centering \includegraphics[width=1\linewidth]{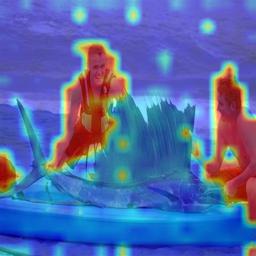} \caption{``person''} \end{subfigure} \hfill
    \begin{subfigure}{0.162\textwidth} \centering \includegraphics[width=1\linewidth]{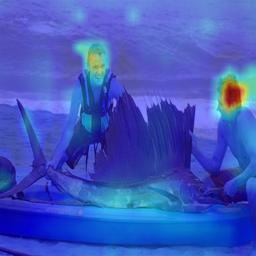} \caption{``person's eye''} \end{subfigure} \hfill
    \begin{subfigure}{0.162\textwidth} \centering \includegraphics[width=1\linewidth]{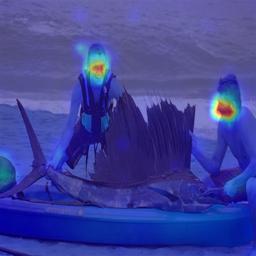} \caption{``person's nose''} \end{subfigure} \hfill
    \begin{subfigure}{0.162\textwidth} \centering \includegraphics[width=1\linewidth]{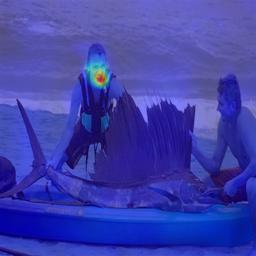} \caption{``person's mouth''} \end{subfigure} \hfill
    \begin{subfigure}{0.162\textwidth} \centering \includegraphics[width=1\linewidth]{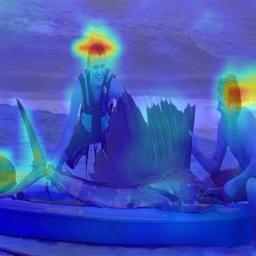} \caption{``person's ear''} \end{subfigure} \hfill
    \begin{subfigure}{0.162\textwidth} \centering \includegraphics[width=1\linewidth]{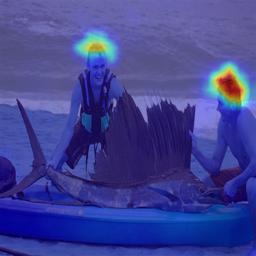} \caption{``person's hair''} \end{subfigure} \hfill
    \begin{subfigure}{0.162\textwidth} \centering \includegraphics[width=1\linewidth]{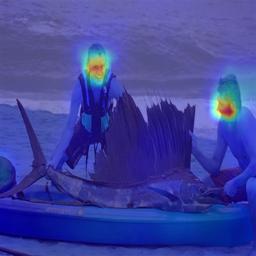} \caption{``person's head''} \end{subfigure} \hfill
    \begin{subfigure}{0.162\textwidth} \centering \includegraphics[width=1\linewidth]{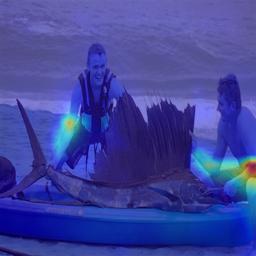} \caption{``person's lower arm''} \end{subfigure} \hfill
    \begin{subfigure}{0.162\textwidth} \centering \includegraphics[width=1\linewidth]{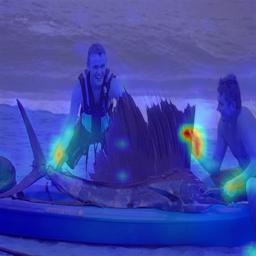} \caption{``person's hand''} \end{subfigure} \hfill
    \begin{subfigure}{0.162\textwidth} \centering \includegraphics[width=1\linewidth]{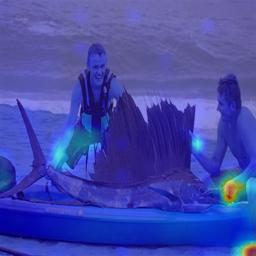} \caption{``person's foot''} \end{subfigure} \hfill
    \begin{subfigure}{0.162\textwidth} \centering \includegraphics[width=1\linewidth]{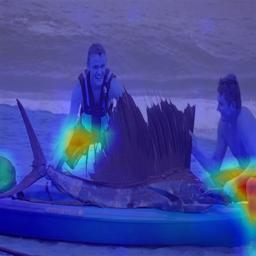} \caption{``person's leg''} \end{subfigure} \hfill
    \begin{subfigure}{0.162\textwidth} \centering \includegraphics[width=1\linewidth]{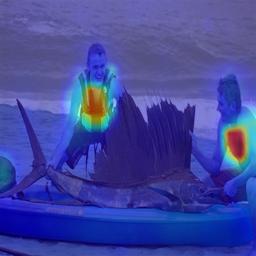} \caption{``person's torso''} \end{subfigure} \hfill
    \caption{
        Image-Text Correspondence Visualization \textbf{After} Training
    }
    \label{suppl_fig:cost_volume_1_person}

\end{figure*}

\section{Additional Qualitative Results}

\subsection{Pascal-Part-116 (Oracle-Obj Setting)}

\begin{figure*}[ht]
    \centering
    \begin{subfigure}[t]{0.195\textwidth} \includegraphics[width=\textwidth]{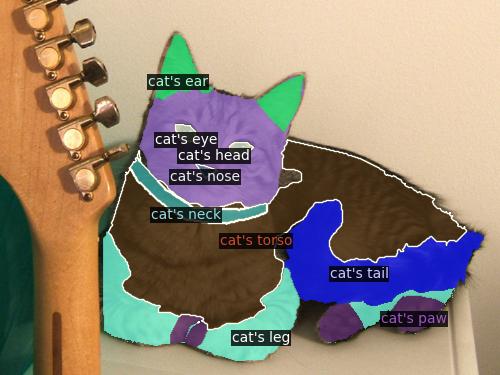} \end{subfigure}
    \begin{subfigure}[t]{0.195\textwidth} \includegraphics[width=\textwidth]{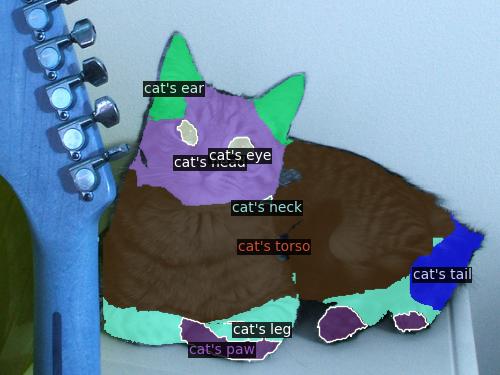} \end{subfigure}
    \begin{subfigure}[t]{0.195\textwidth} \includegraphics[width=\textwidth]{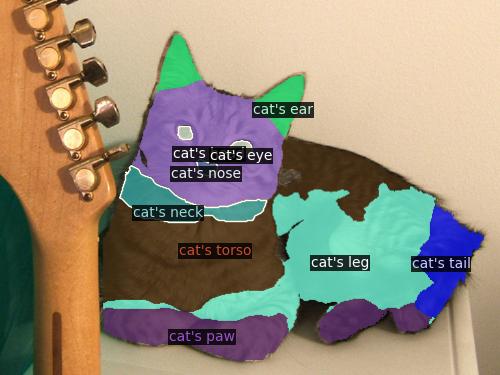} \end{subfigure}
    \begin{subfigure}[t]{0.195\textwidth} \includegraphics[width=\textwidth]{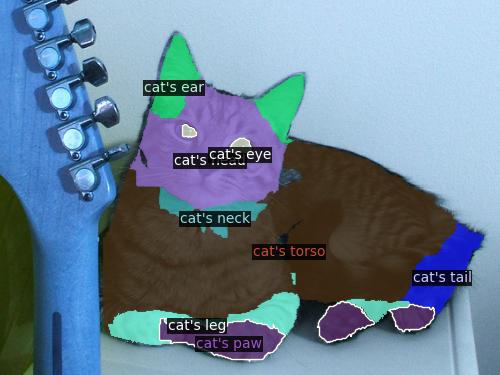} \end{subfigure}
    \begin{subfigure}[t]{0.195\textwidth} \includegraphics[width=\textwidth]{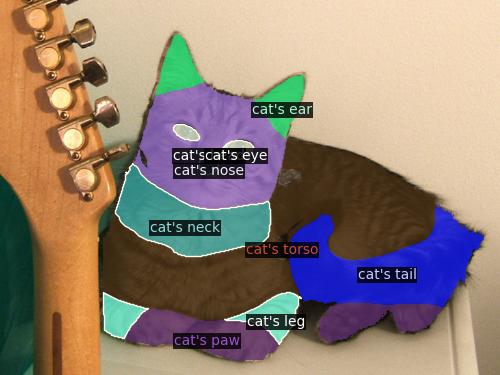} \end{subfigure}
    \begin{subfigure}[t]{0.195\textwidth} \includegraphics[width=\textwidth]{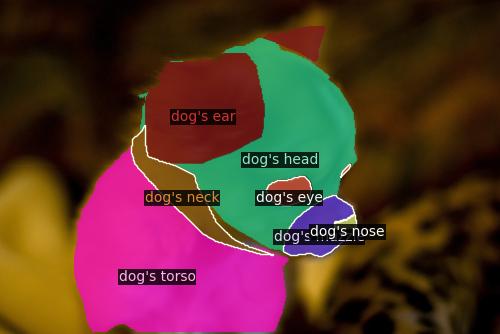} \end{subfigure}
    \begin{subfigure}[t]{0.195\textwidth} \includegraphics[width=\textwidth]{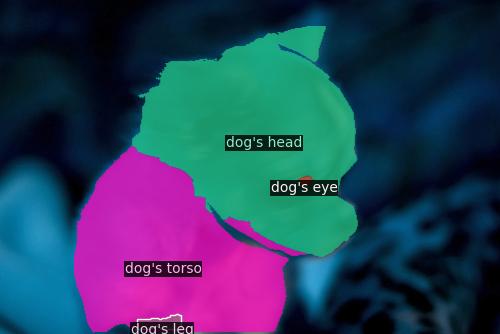} \end{subfigure}
    \begin{subfigure}[t]{0.195\textwidth} \includegraphics[width=\textwidth]{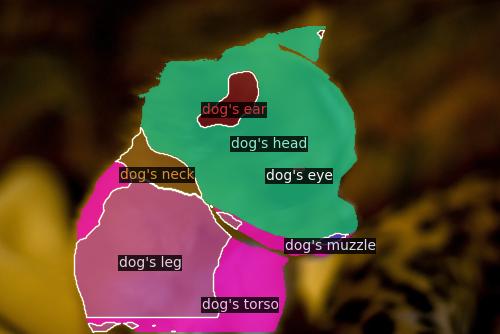} \end{subfigure}
    \begin{subfigure}[t]{0.195\textwidth} \includegraphics[width=\textwidth]{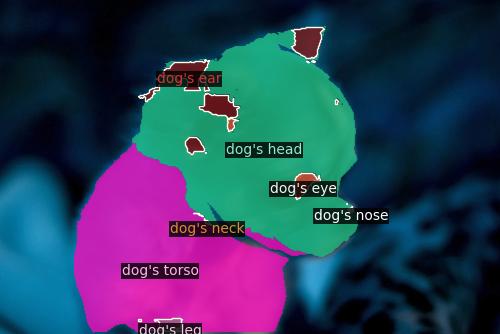} \end{subfigure}
    \begin{subfigure}[t]{0.195\textwidth} \includegraphics[width=\textwidth]{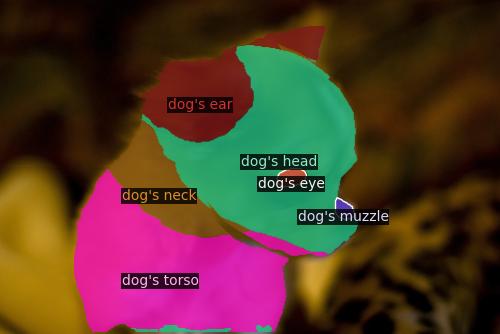} \end{subfigure}
    \begin{subfigure}[t]{0.195\textwidth} \includegraphics[width=\textwidth, trim=0 0 0 20, clip]{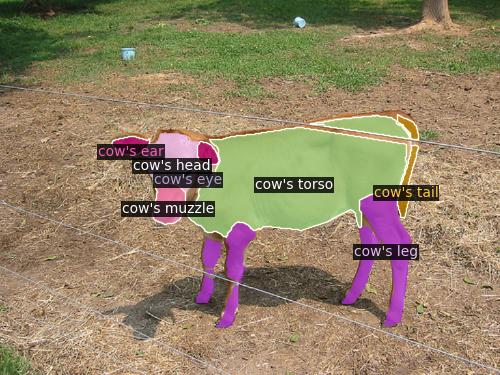} \end{subfigure}
    \begin{subfigure}[t]{0.195\textwidth} \includegraphics[width=\textwidth, trim=0 0 0 20, clip]{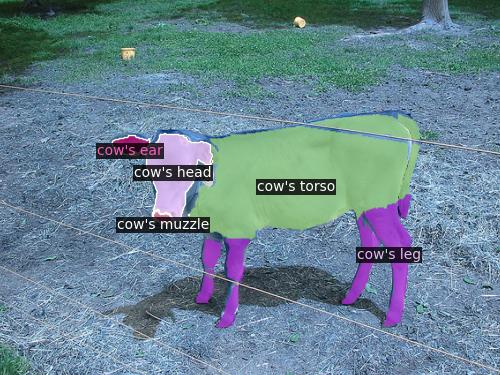}
    \end{subfigure}
    \begin{subfigure}[t]{0.195\textwidth} \includegraphics[width=\textwidth, trim=0 0 0 20, clip]{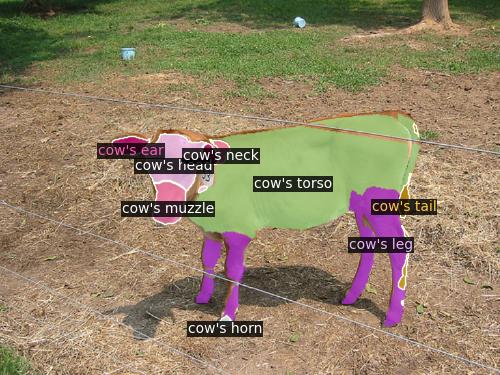} \end{subfigure}
    \begin{subfigure}[t]{0.195\textwidth} \includegraphics[width=\textwidth, trim=0 0 0 20, clip]{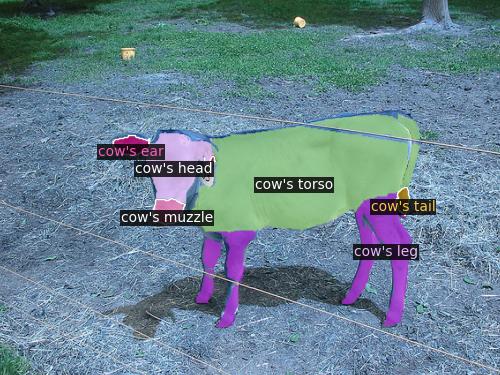} \end{subfigure}
    \begin{subfigure}[t]{0.195\textwidth} \includegraphics[width=\textwidth, trim=0 0 0 20, clip]{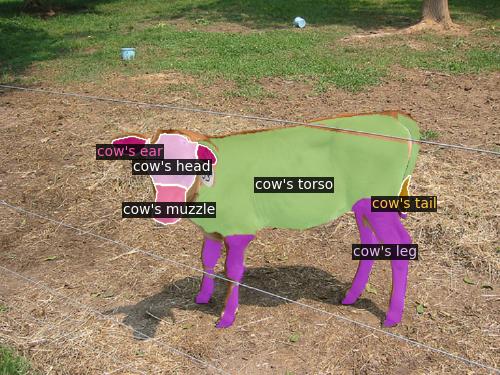} \end{subfigure}
    \begin{subfigure}[t]{0.195\textwidth} \includegraphics[width=\textwidth]{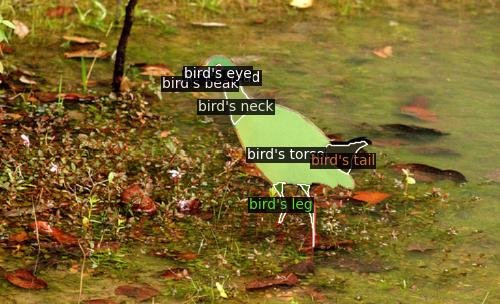} \end{subfigure}
    \begin{subfigure}[t]{0.195\textwidth} \includegraphics[width=\textwidth]{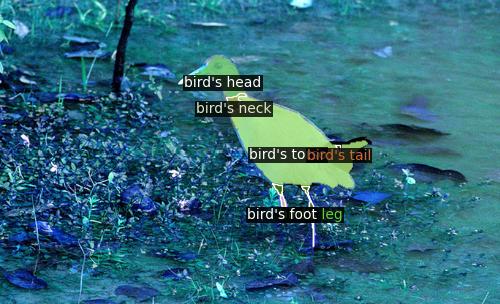} \end{subfigure}
    \begin{subfigure}[t]{0.195\textwidth} \includegraphics[width=\textwidth]{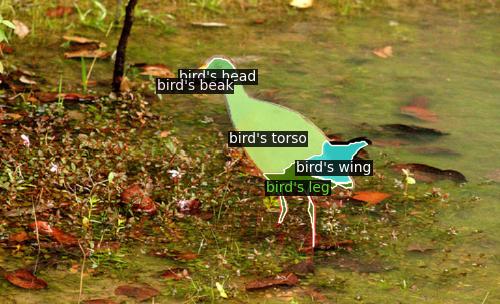} \end{subfigure}
    \begin{subfigure}[t]{0.195\textwidth} \includegraphics[width=\textwidth]{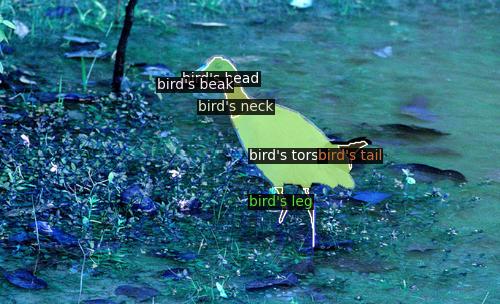} \end{subfigure}
    \begin{subfigure}[t]{0.195\textwidth} \includegraphics[width=\textwidth]{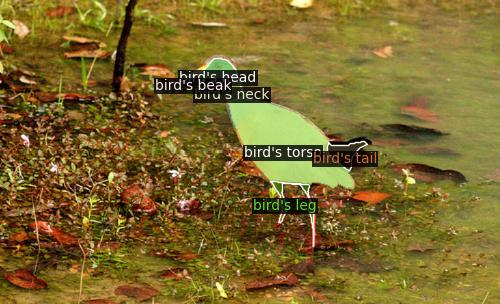} \end{subfigure}
    \begin{subfigure}[t]{0.195\textwidth} \includegraphics[width=\textwidth, trim=0 0 0 50, clip]{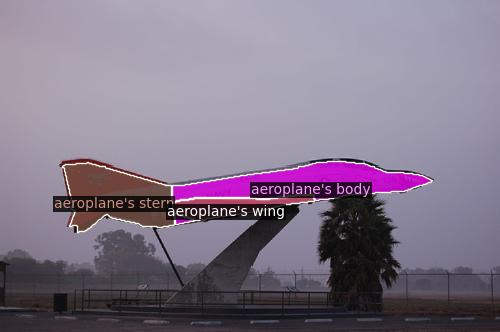} \end{subfigure}
    \begin{subfigure}[t]{0.195\textwidth} \includegraphics[width=\textwidth, trim=0 0 0 50, clip]{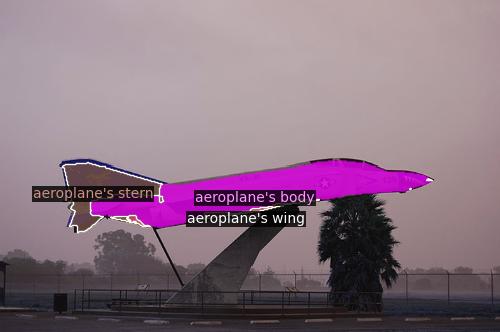} \end{subfigure}
    \begin{subfigure}[t]{0.195\textwidth} \includegraphics[width=\textwidth, trim=0 0 0 50, clip]{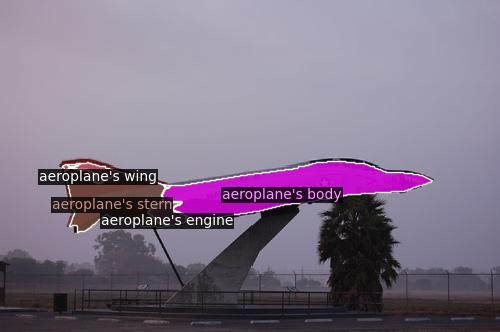} \end{subfigure}
    \begin{subfigure}[t]{0.195\textwidth} \includegraphics[width=\textwidth, trim=0 0 0 50, clip]{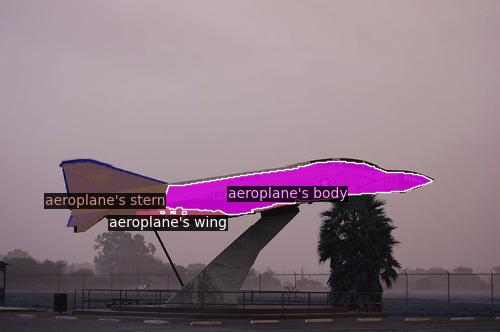} \end{subfigure}
    \begin{subfigure}[t]{0.195\textwidth} \includegraphics[width=\textwidth, trim=0 0 0 50, clip]{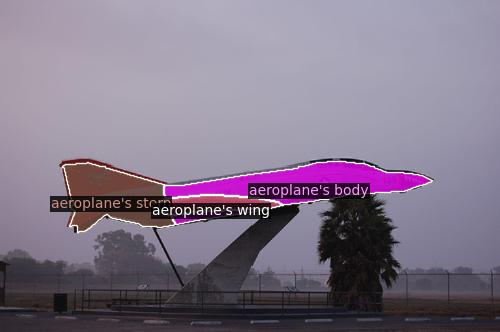} \end{subfigure}
    \vspace{-1em}
    \begin{subfigure}[t]{0.195\textwidth} \includegraphics[width=\textwidth]{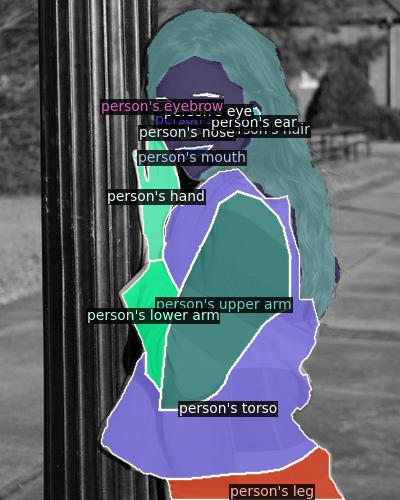} \end{subfigure}
    \begin{subfigure}[t]{0.195\textwidth} \includegraphics[width=\textwidth]{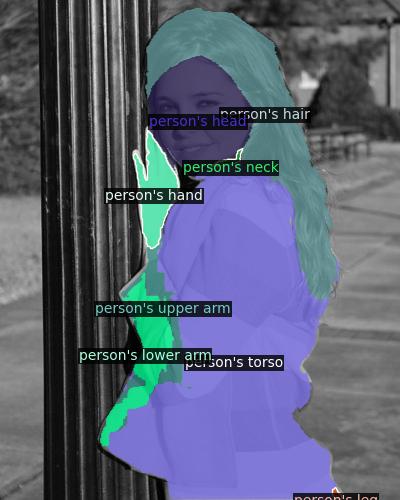} \end{subfigure}
    \begin{subfigure}[t]{0.195\textwidth} \includegraphics[width=\textwidth]{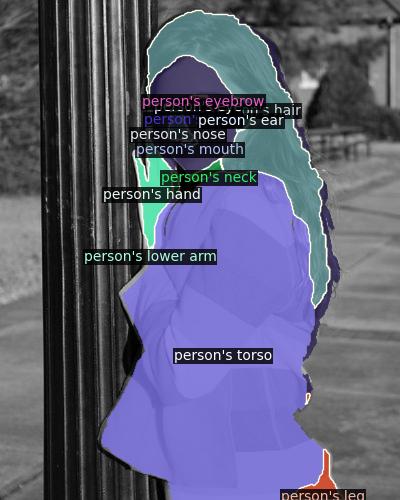} \end{subfigure}
    \begin{subfigure}[t]{0.195\textwidth} \includegraphics[width=\textwidth]{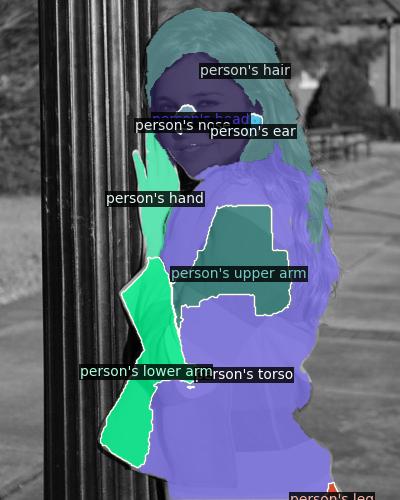} \end{subfigure}
    \begin{subfigure}[t]{0.195\textwidth} \includegraphics[width=\textwidth]{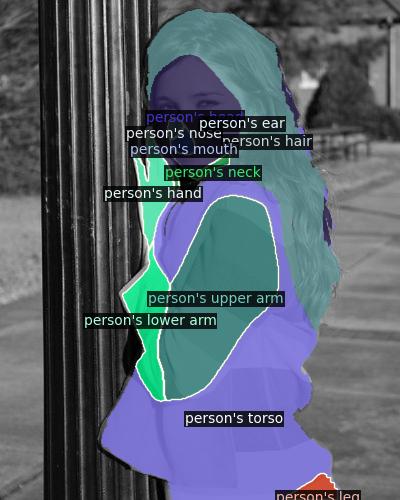} \end{subfigure}
    \vspace{-1em}
    \begin{subfigure}[t]{0.195\textwidth} \caption{Ground-truth} \end{subfigure}
    \begin{subfigure}[t]{0.195\textwidth} \caption{CLIPSeg~\cite{radford2021learning_CLIP,wei2024ov_OV_PARTS}} \end{subfigure}
    \begin{subfigure}[t]{0.195\textwidth} \caption{CAT-Seg~\cite{cho2023cat_CATSeg,wei2024ov_OV_PARTS}} \end{subfigure}
    \begin{subfigure}[t]{0.195\textwidth} \caption{PartCLIPSeg~\cite{PartCLIPSeg2024}} \end{subfigure}
    \begin{subfigure}[t]{0.160\textwidth} \caption{PartCATSeg} \end{subfigure}
    \vspace{1em}
    \caption{
        Qualitative evaluation of zero-shot part segmentation on Pascal-Part-116 in the \textbf{Oracle-Obj} configuration.
        Note that annotations for unseen categories (e.g., bird, cow, dog) are excluded from the training set.
    }
    \label{fig:vis_oracle_qualitative}
\end{figure*}

The~\Cref{fig:vis_oracle_qualitative} illustrates qualitative evaluations under the Oracle-Obj setting, which assumes that object masks are provided.
This setting evaluates the fine-grained part segmentation results using object-level segmentation generated by other off-the-shelf OVSS models.

PartCATSeg consistently demonstrates superior segmentation performance compared to other baselines.
Notably, it effectively segments small parts, such as arms and eyes, which are often missed by other models.

\subsection{PartImageNet (Pred-All Setting)}

\begin{figure*}[ht]
    \centering
    \begin{subfigure}[t]{0.195\textwidth} \includegraphics[width=\textwidth]{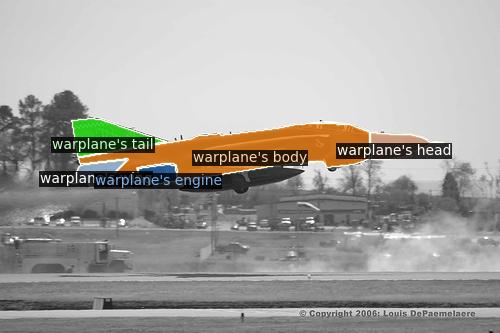} \end{subfigure}
    \begin{subfigure}[t]{0.195\textwidth} \includegraphics[width=\textwidth]{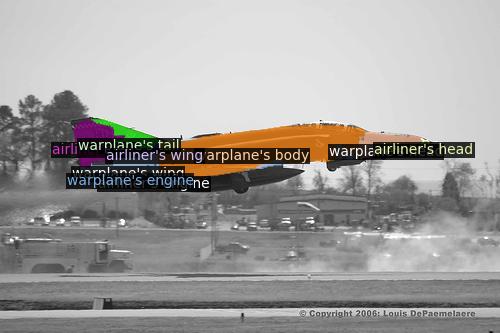} \end{subfigure}
    \begin{subfigure}[t]{0.195\textwidth} \includegraphics[width=\textwidth]{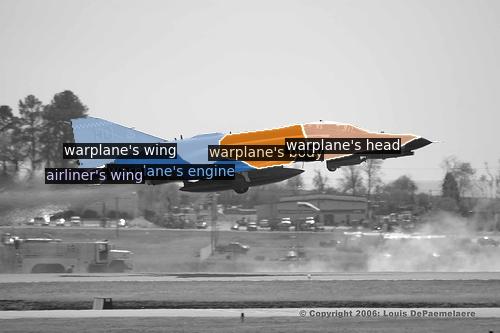} \end{subfigure}
    \begin{subfigure}[t]{0.195\textwidth} \includegraphics[width=\textwidth]{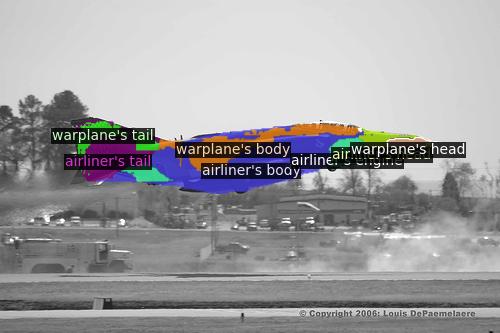} \end{subfigure}
    \begin{subfigure}[t]{0.195\textwidth} \includegraphics[width=\textwidth]{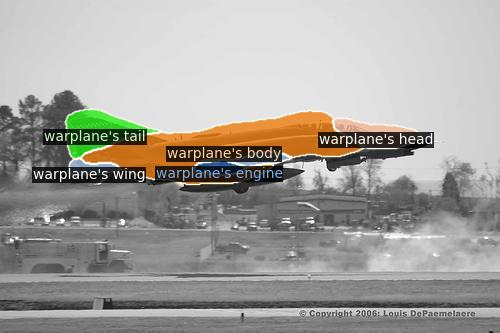} \end{subfigure}
    \begin{subfigure}[t]{0.195\textwidth} \includegraphics[width=\textwidth]{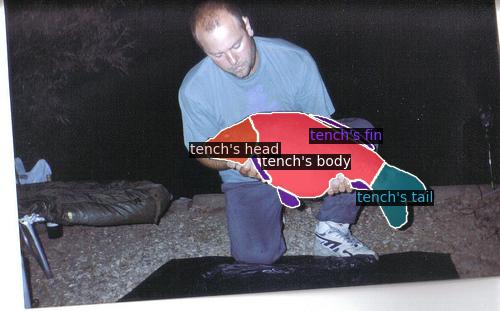} \end{subfigure}
    \begin{subfigure}[t]{0.195\textwidth} \includegraphics[width=\textwidth]{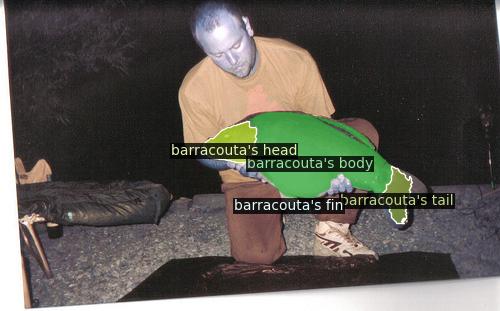} \end{subfigure}
    \begin{subfigure}[t]{0.195\textwidth} \includegraphics[width=\textwidth]{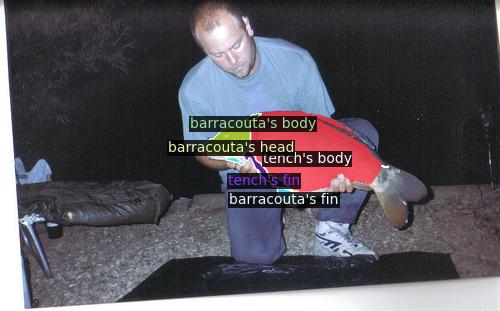} \end{subfigure}
    \begin{subfigure}[t]{0.195\textwidth} \includegraphics[width=\textwidth]{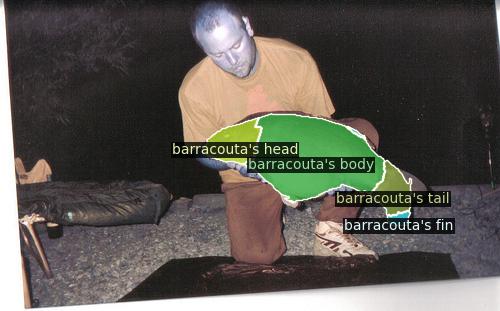} \end{subfigure}
    \begin{subfigure}[t]{0.195\textwidth} \includegraphics[width=\textwidth]{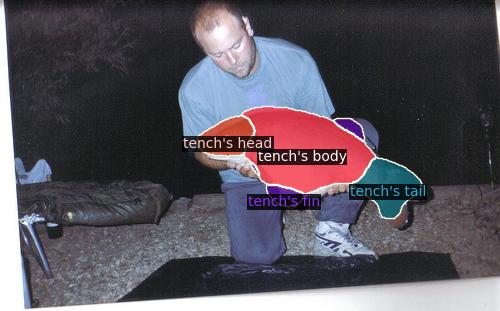} \end{subfigure}
    \begin{subfigure}[t]{0.195\textwidth} \includegraphics[width=\textwidth, trim=0 0 0 20, clip]{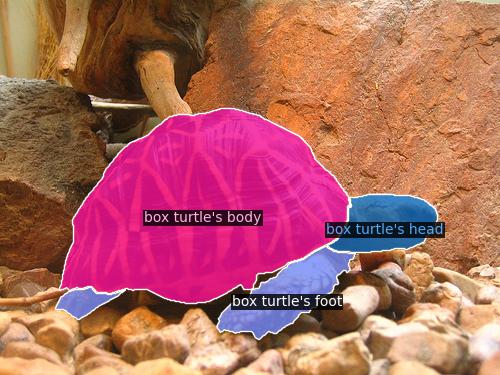} \end{subfigure}
    \begin{subfigure}[t]{0.195\textwidth} \includegraphics[width=\textwidth, trim=0 0 0 20, clip]{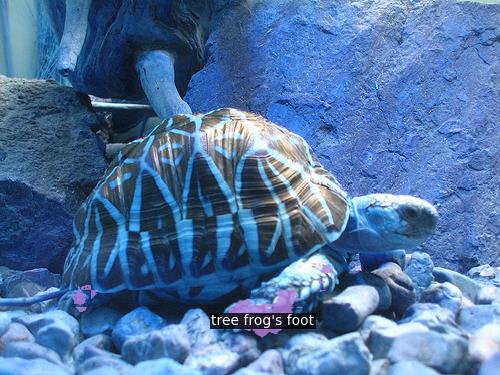}
    \end{subfigure}
    \begin{subfigure}[t]{0.195\textwidth} \includegraphics[width=\textwidth, trim=0 0 0 20, clip]{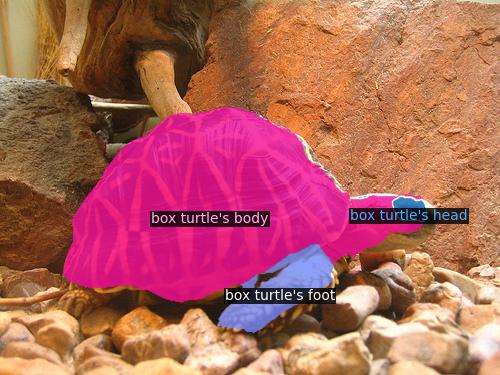} \end{subfigure}
    \begin{subfigure}[t]{0.195\textwidth} \includegraphics[width=\textwidth, trim=0 0 0 20, clip]{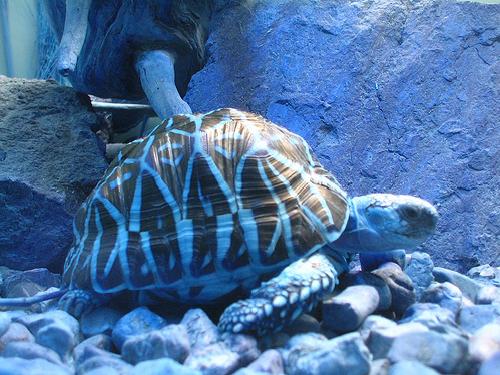} \end{subfigure}
    \begin{subfigure}[t]{0.195\textwidth} \includegraphics[width=\textwidth, trim=0 0 0 20, clip]{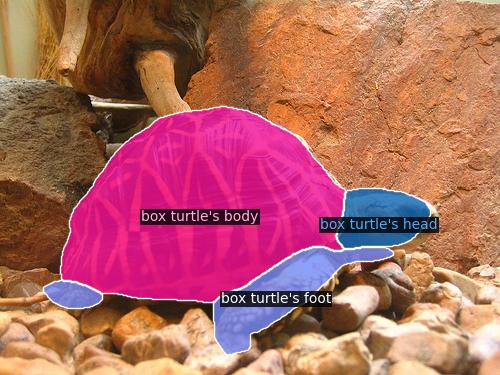} \end{subfigure}
    \begin{subfigure}[t]{0.195\textwidth} \includegraphics[width=\textwidth]{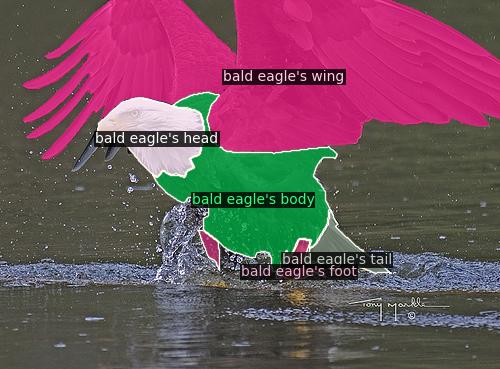} \end{subfigure}
    \begin{subfigure}[t]{0.195\textwidth} \includegraphics[width=\textwidth]{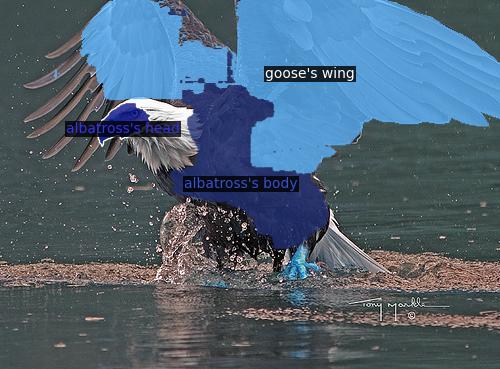} \end{subfigure}
    \begin{subfigure}[t]{0.195\textwidth} \includegraphics[width=\textwidth]{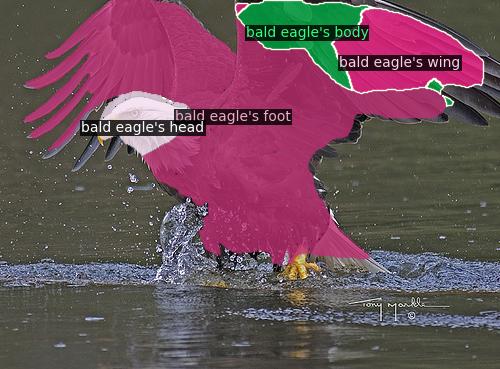} \end{subfigure}
    \begin{subfigure}[t]{0.195\textwidth} \includegraphics[width=\textwidth]{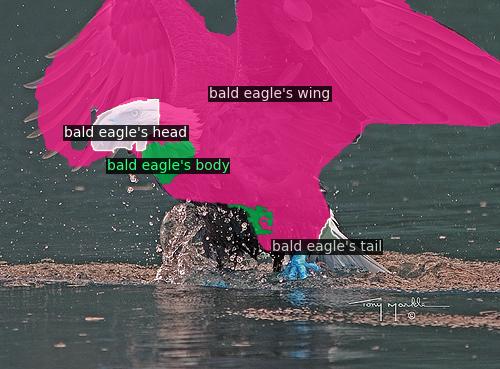} \end{subfigure}
    \begin{subfigure}[t]{0.195\textwidth} \includegraphics[width=\textwidth]{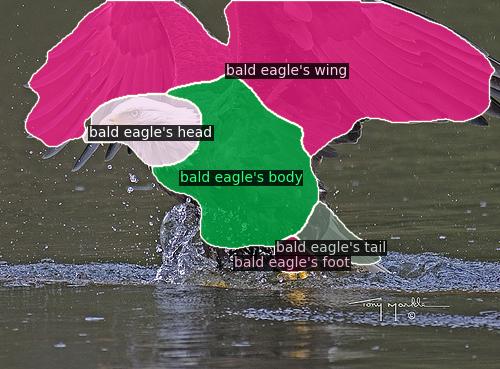} \end{subfigure}
    \begin{subfigure}[t]{0.195\textwidth} \includegraphics[width=\textwidth, trim=0 0 0 50, clip]{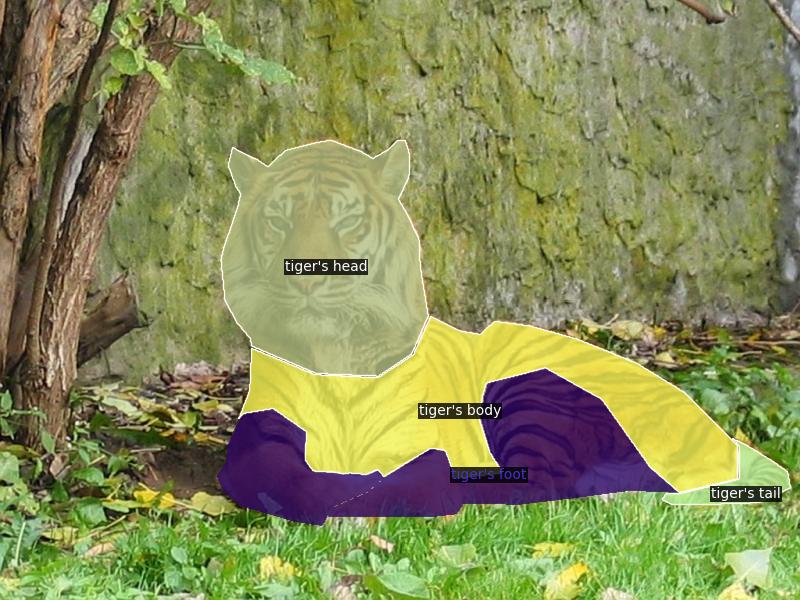} \end{subfigure}
    \begin{subfigure}[t]{0.195\textwidth} \includegraphics[width=\textwidth, trim=0 0 0 50, clip]{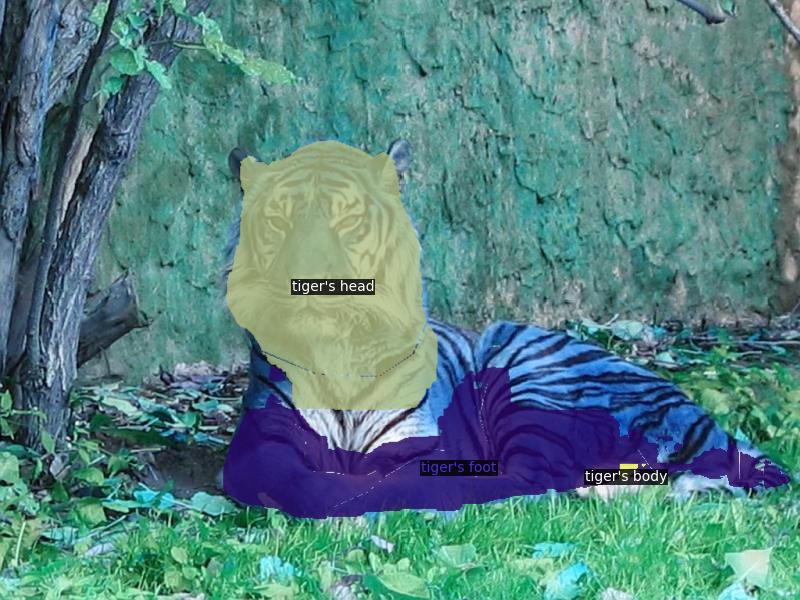} \end{subfigure}
    \begin{subfigure}[t]{0.195\textwidth} \includegraphics[width=\textwidth, trim=0 0 0 50, clip]{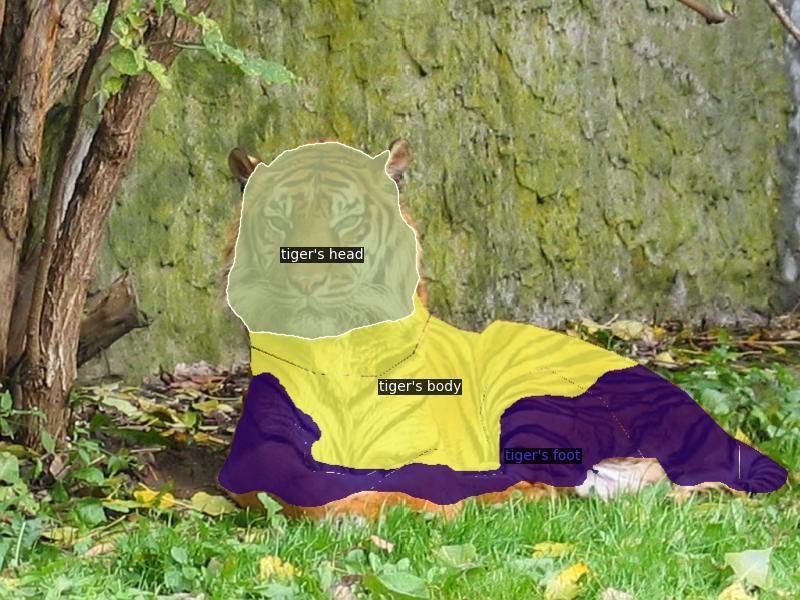} \end{subfigure}
    \begin{subfigure}[t]{0.195\textwidth} \includegraphics[width=\textwidth, trim=0 0 0 50, clip]{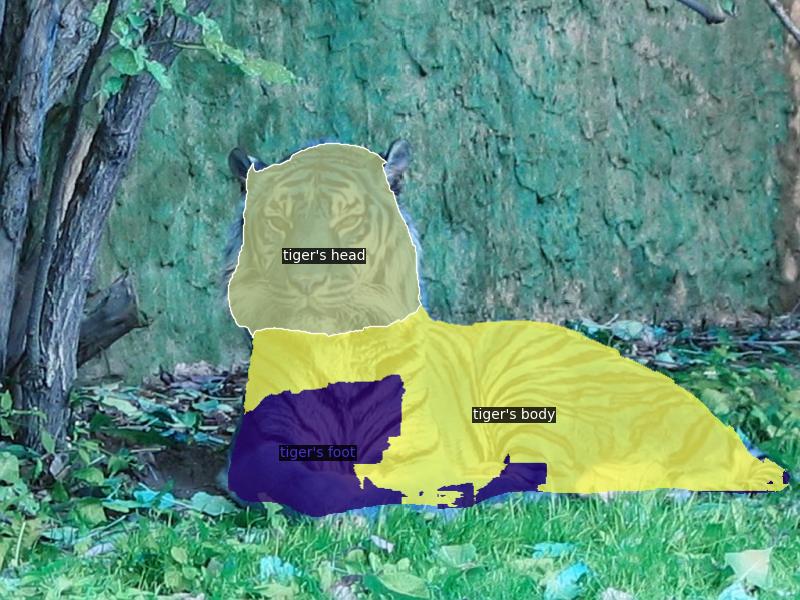} \end{subfigure}
    \begin{subfigure}[t]{0.195\textwidth} \includegraphics[width=\textwidth, trim=0 0 0 50, clip]{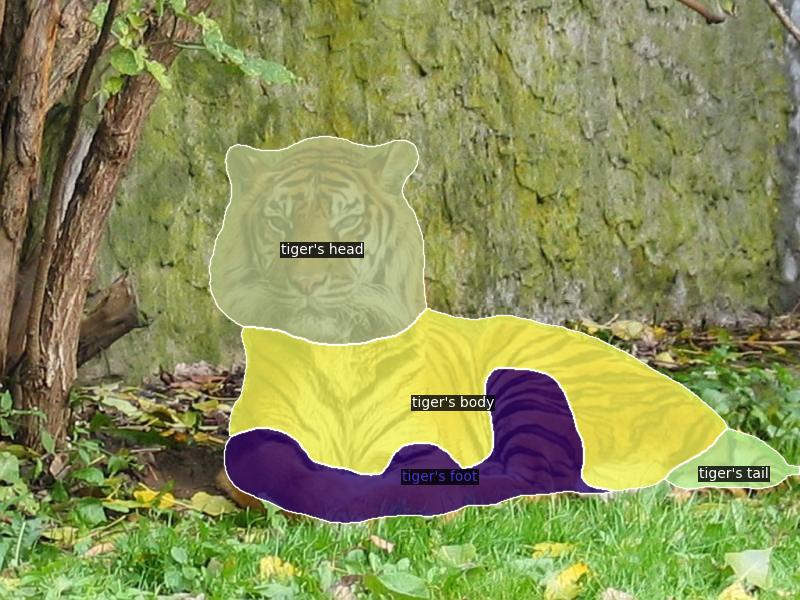} \end{subfigure}
    \vspace{-1em}
    \begin{subfigure}[t]{0.195\textwidth} \includegraphics[width=\textwidth]{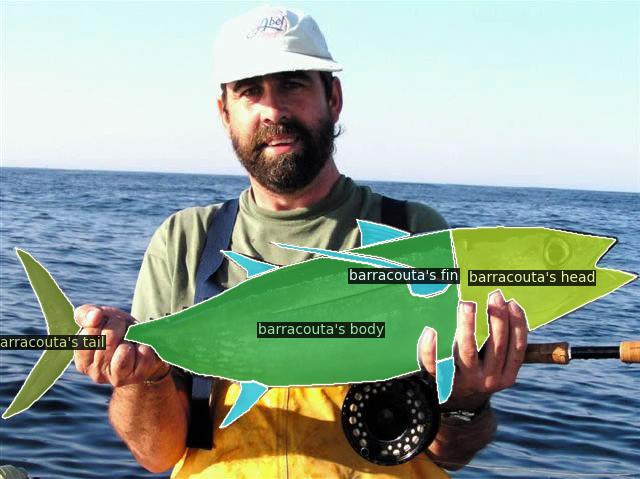} \end{subfigure}
    \begin{subfigure}[t]{0.195\textwidth} \includegraphics[width=\textwidth]{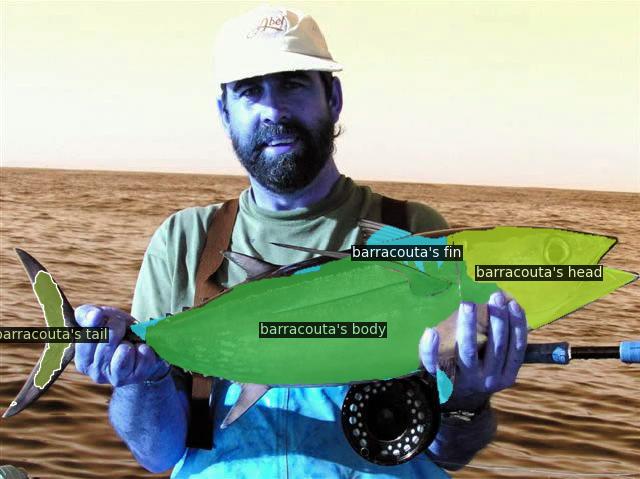} \end{subfigure}
    \begin{subfigure}[t]{0.195\textwidth} \includegraphics[width=\textwidth]{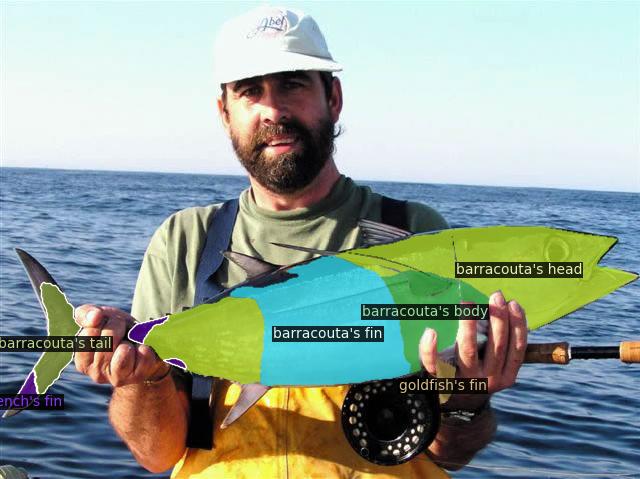} \end{subfigure}
    \begin{subfigure}[t]{0.195\textwidth} \includegraphics[width=\textwidth]{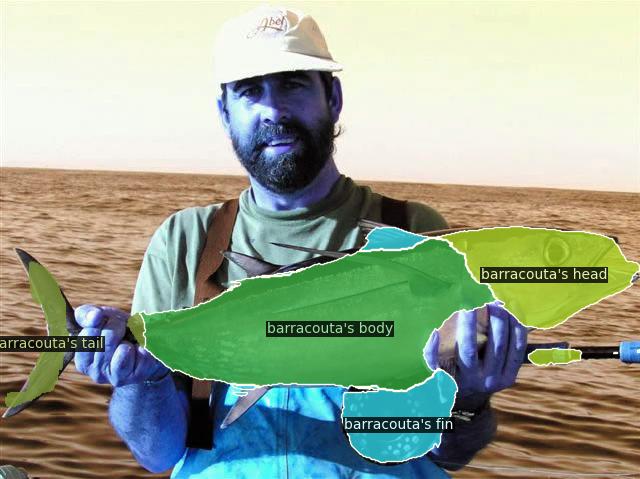} \end{subfigure}
    \begin{subfigure}[t]{0.195\textwidth} \includegraphics[width=\textwidth]{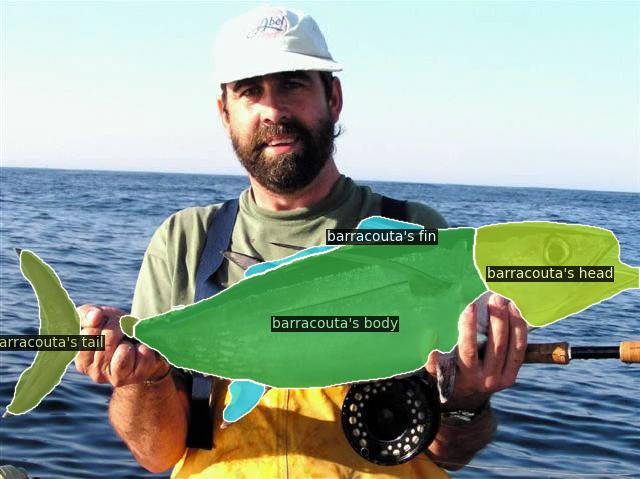} \end{subfigure}
    \vspace{-1em}
    \begin{subfigure}[t]{0.195\textwidth} \caption{Ground-truth} \end{subfigure}
    \begin{subfigure}[t]{0.195\textwidth} \caption{CLIPSeg~\cite{radford2021learning_CLIP,wei2024ov_OV_PARTS}} \end{subfigure}
    \begin{subfigure}[t]{0.195\textwidth} \caption{CAT-Seg~\cite{cho2023cat_CATSeg,wei2024ov_OV_PARTS}} \end{subfigure}
    \begin{subfigure}[t]{0.195\textwidth} \caption{PartCLIPSeg~\cite{PartCLIPSeg2024}} \end{subfigure}
    \begin{subfigure}[t]{0.195\textwidth} \caption{PartCATSeg} \end{subfigure}
    \vspace{1em}
    \caption{
        Qualitative evaluation of zero-shot part segmentation on \textbf{PartImageNet} in the \textbf{Pred-All} configuration.
    }
    \label{suppl_fig:vis_pred_qualitative_partimagenet}
\end{figure*}

The following~\Cref{suppl_fig:vis_pred_qualitative_partimagenet} illustrates the PartImageNet prediction results of PartCATSeg in Pred-All setup.
PartCATSeg demonstrates superior performance compared to other baselines, accurately predicting both appropriate classes and boundaries.

\subsection{PartImageNet (Oracle-Obj Setting)}

\begin{figure*}[ht]
    \centering
    \begin{subfigure}[t]{0.195\textwidth} \includegraphics[width=\textwidth]{assets/supplementary/partimagenet/GT/n04552348_241_gt.jpg} \end{subfigure}
    \begin{subfigure}[t]{0.195\textwidth} \includegraphics[width=\textwidth]{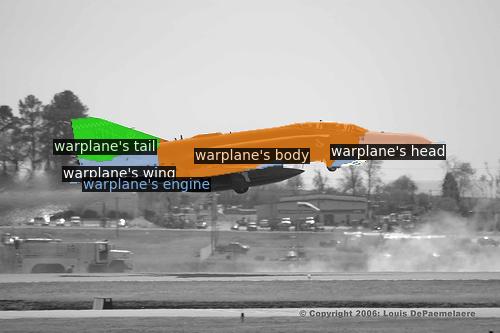} \end{subfigure}
    \begin{subfigure}[t]{0.195\textwidth} \includegraphics[width=\textwidth]{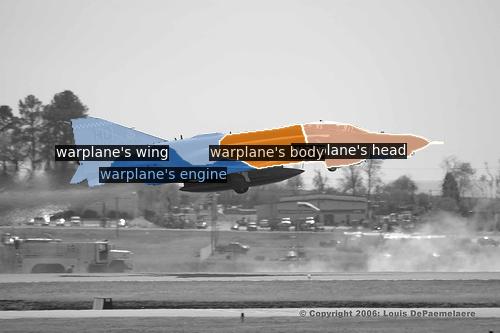} \end{subfigure}
    \begin{subfigure}[t]{0.195\textwidth} \includegraphics[width=\textwidth]{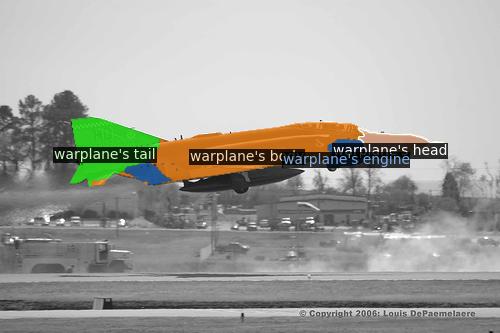} \end{subfigure}
    \begin{subfigure}[t]{0.195\textwidth} \includegraphics[width=\textwidth]{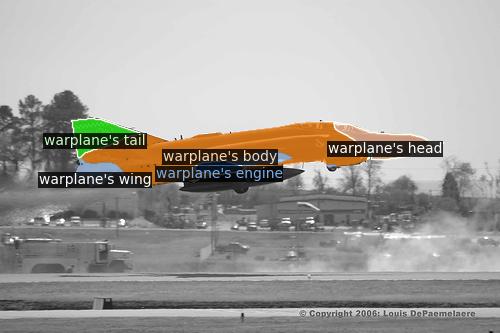} \end{subfigure}
    \begin{subfigure}[t]{0.195\textwidth} \includegraphics[width=\textwidth]{assets/supplementary/partimagenet/GT/n01440764_1302_gt.jpg} \end{subfigure}
    \begin{subfigure}[t]{0.195\textwidth} \includegraphics[width=\textwidth]{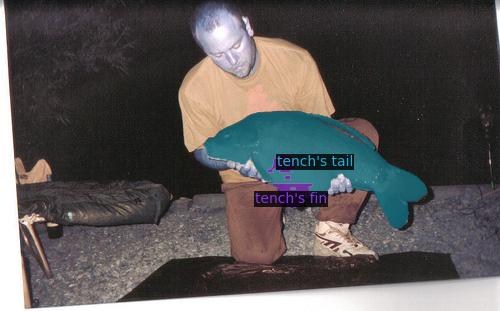} \end{subfigure}
    \begin{subfigure}[t]{0.195\textwidth} \includegraphics[width=\textwidth]{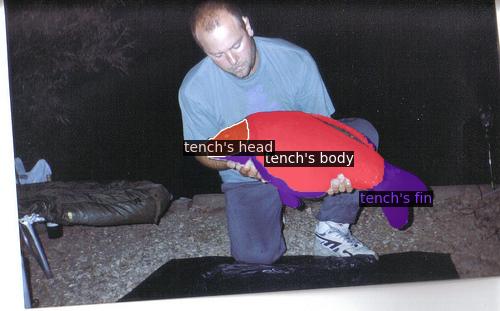} \end{subfigure}
    \begin{subfigure}[t]{0.195\textwidth} \includegraphics[width=\textwidth]{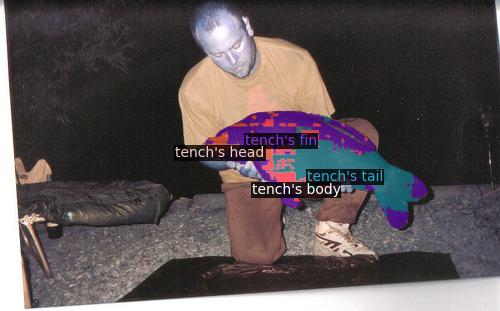} \end{subfigure}
    \begin{subfigure}[t]{0.195\textwidth} \includegraphics[width=\textwidth]{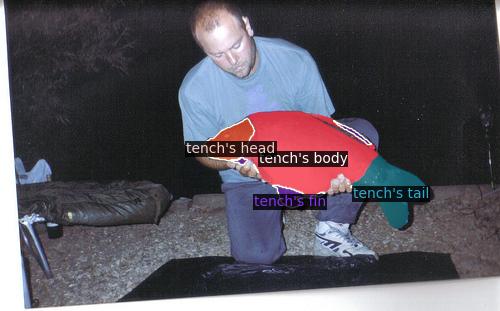} \end{subfigure}
    \begin{subfigure}[t]{0.195\textwidth} \includegraphics[width=\textwidth, trim=0 0 0 20, clip]{assets/supplementary/partimagenet/GT/n01669191_5933_gt.jpg} \end{subfigure}
    \begin{subfigure}[t]{0.195\textwidth} \includegraphics[width=\textwidth, trim=0 0 0 20, clip]{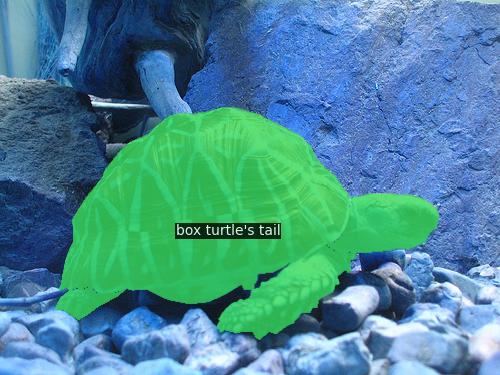}
    \end{subfigure}
    \begin{subfigure}[t]{0.195\textwidth} \includegraphics[width=\textwidth, trim=0 0 0 20, clip]{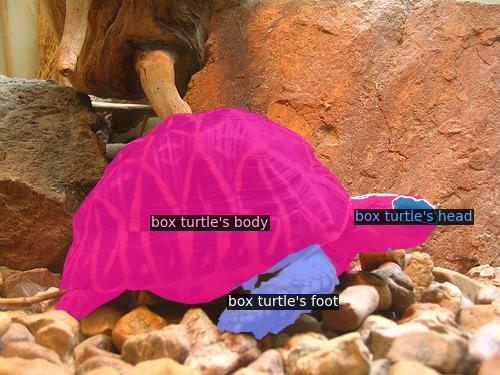} \end{subfigure}
    \begin{subfigure}[t]{0.195\textwidth} \includegraphics[width=\textwidth, trim=0 0 0 20, clip]{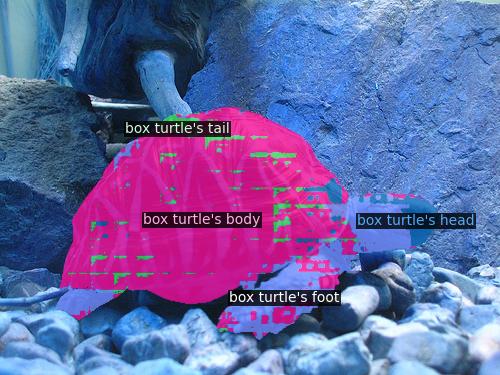} \end{subfigure}
    \begin{subfigure}[t]{0.195\textwidth} \includegraphics[width=\textwidth, trim=0 0 0 20, clip]{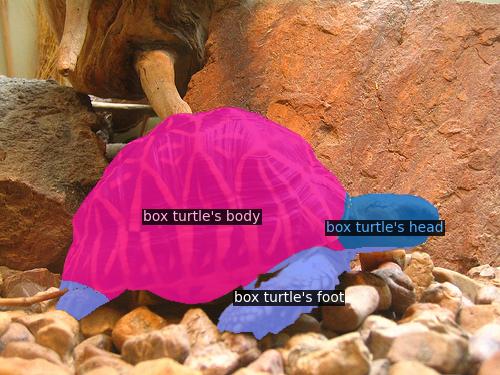} \end{subfigure}
    \begin{subfigure}[t]{0.195\textwidth} \includegraphics[width=\textwidth]{assets/supplementary/partimagenet/GT/n01614925_9631_gt.jpg} \end{subfigure}
    \begin{subfigure}[t]{0.195\textwidth} \includegraphics[width=\textwidth]{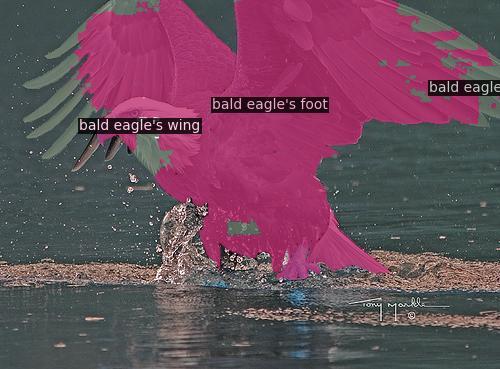} \end{subfigure}
    \begin{subfigure}[t]{0.195\textwidth} \includegraphics[width=\textwidth]{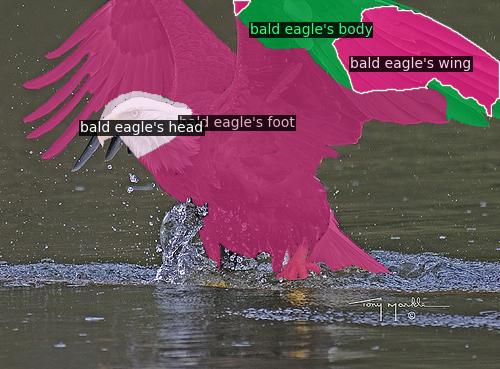} \end{subfigure}
    \begin{subfigure}[t]{0.195\textwidth} \includegraphics[width=\textwidth]{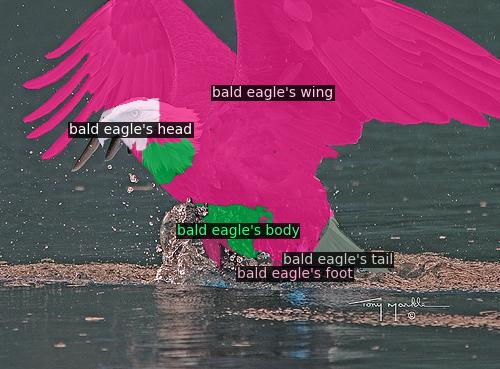} \end{subfigure}
    \begin{subfigure}[t]{0.195\textwidth} \includegraphics[width=\textwidth]{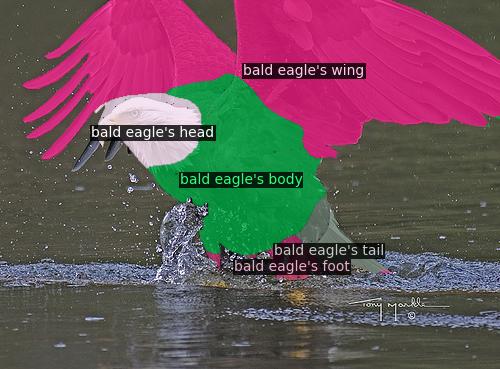} \end{subfigure}
    \begin{subfigure}[t]{0.195\textwidth} \includegraphics[width=\textwidth, trim=0 0 0 50, clip]{assets/supplementary/partimagenet/GT/n02129604_22876_gt.jpg} \end{subfigure}
    \begin{subfigure}[t]{0.195\textwidth} \includegraphics[width=\textwidth, trim=0 0 0 50, clip]{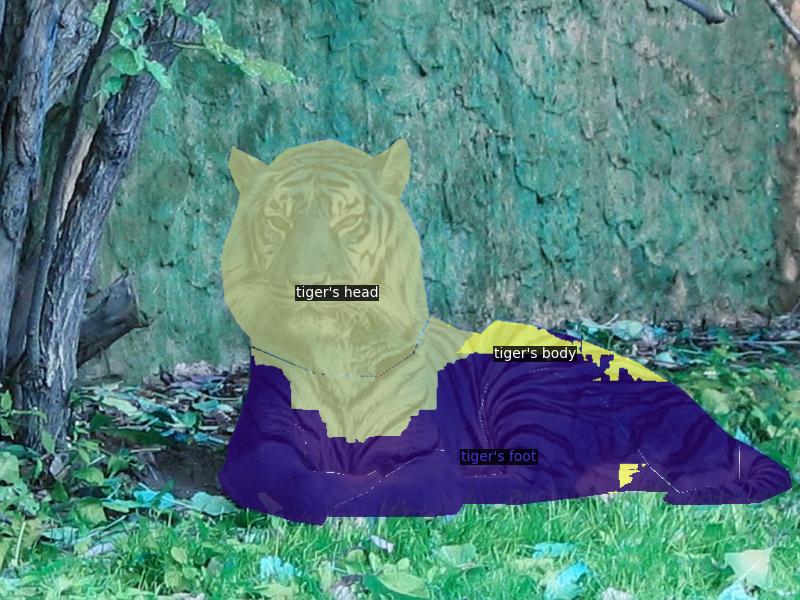} \end{subfigure}
    \begin{subfigure}[t]{0.195\textwidth} \includegraphics[width=\textwidth, trim=0 0 0 50, clip]{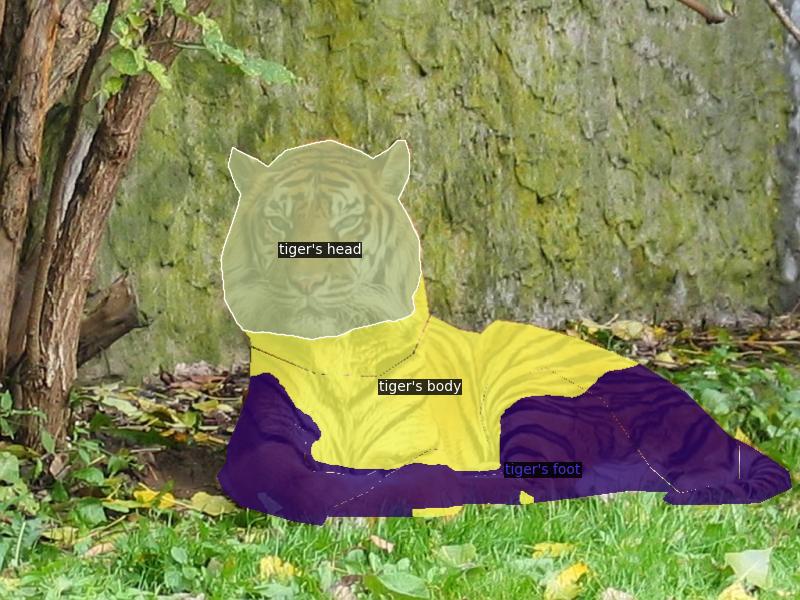} \end{subfigure}
    \begin{subfigure}[t]{0.195\textwidth} \includegraphics[width=\textwidth, trim=0 0 0 50, clip]{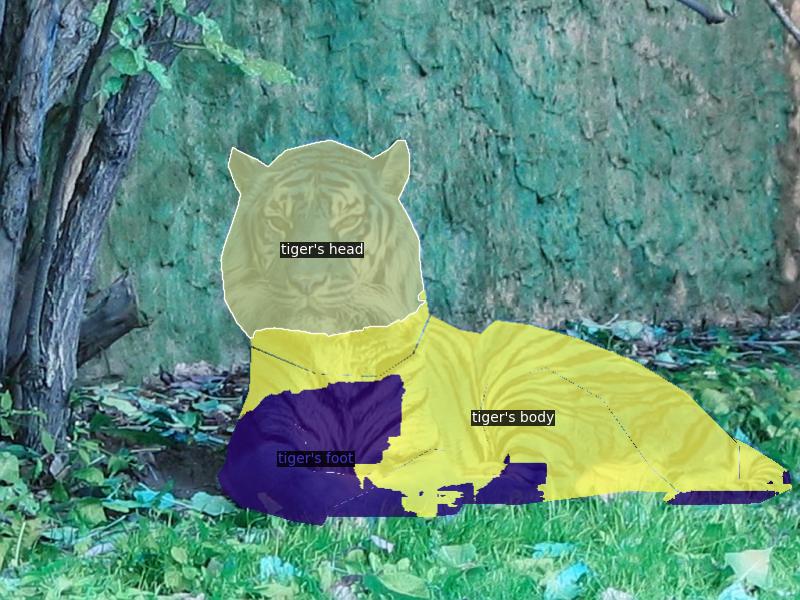} \end{subfigure}
    \begin{subfigure}[t]{0.195\textwidth} \includegraphics[width=\textwidth, trim=0 0 0 50, clip]{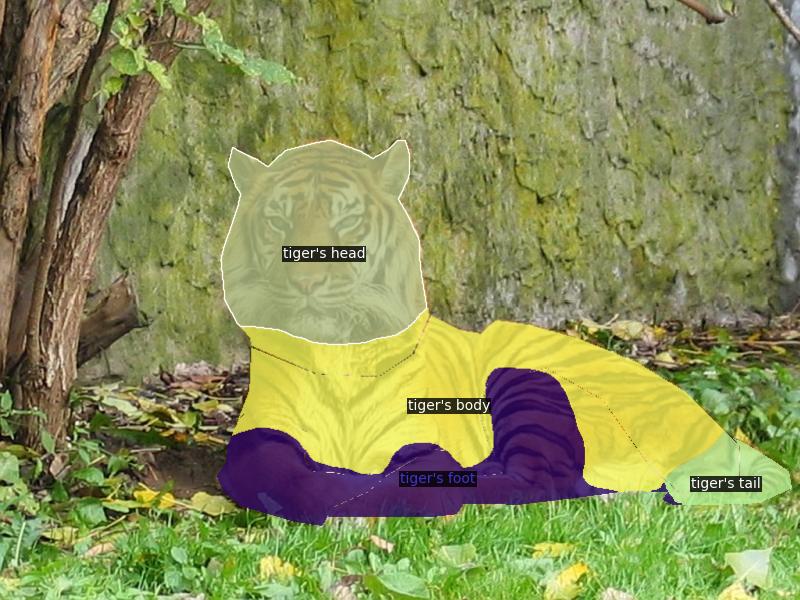} \end{subfigure}
    \vspace{-1em}
    \begin{subfigure}[t]{0.195\textwidth} \includegraphics[width=\textwidth]{assets/supplementary/partimagenet/GT/n02514041_2642_gt.jpg} \end{subfigure}
    \begin{subfigure}[t]{0.195\textwidth} \includegraphics[width=\textwidth]{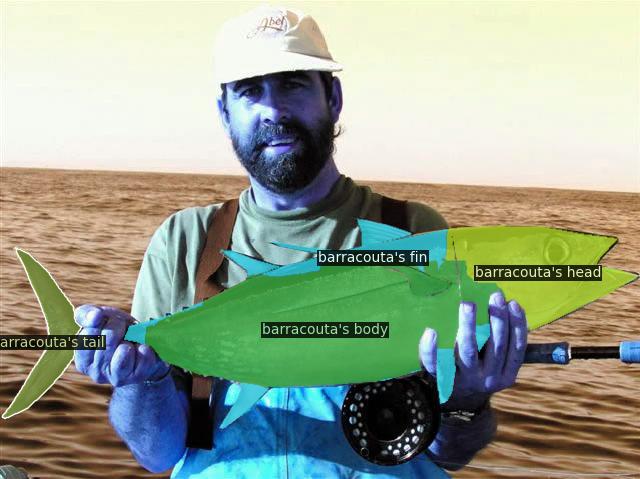} \end{subfigure}
    \begin{subfigure}[t]{0.195\textwidth} \includegraphics[width=\textwidth]{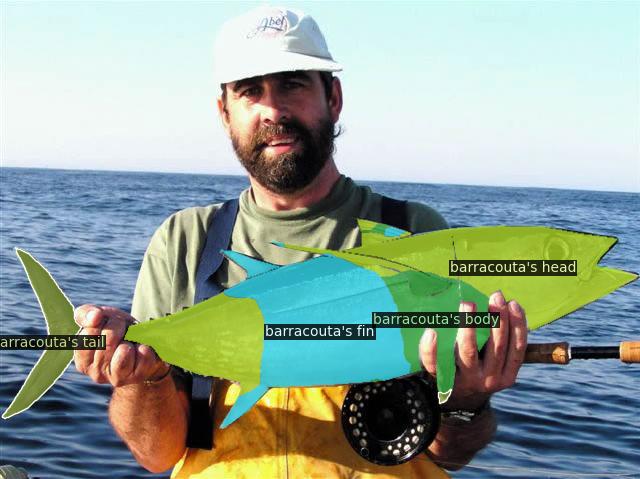} \end{subfigure}
    \begin{subfigure}[t]{0.195\textwidth} \includegraphics[width=\textwidth]{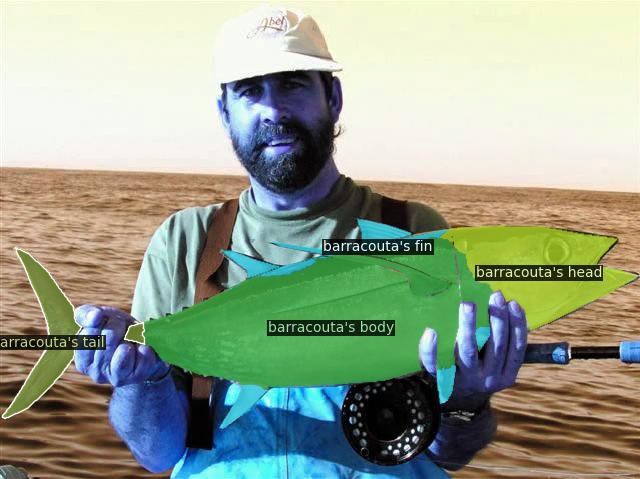} \end{subfigure}
    \begin{subfigure}[t]{0.195\textwidth} \includegraphics[width=\textwidth]{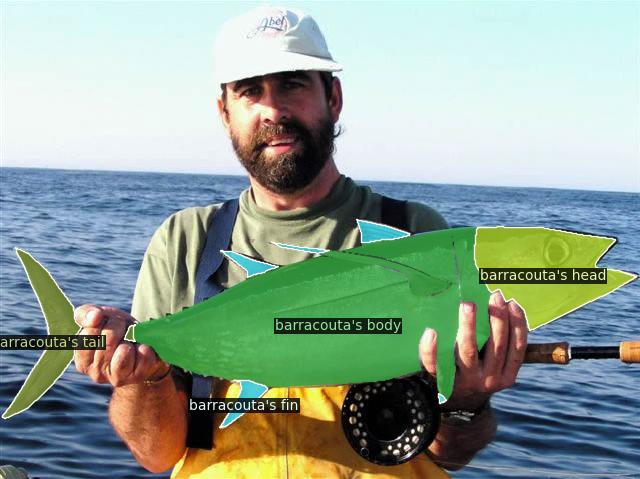} \end{subfigure}
    \vspace{-1em}
    \begin{subfigure}[t]{0.195\textwidth} \caption{Ground-truth} \end{subfigure}
    \begin{subfigure}[t]{0.195\textwidth} \caption{CLIPSeg~\cite{radford2021learning_CLIP,wei2024ov_OV_PARTS}} \end{subfigure}
    \begin{subfigure}[t]{0.195\textwidth} \caption{CAT-Seg~\cite{cho2023cat_CATSeg,wei2024ov_OV_PARTS}} \end{subfigure}
    \begin{subfigure}[t]{0.195\textwidth} \caption{PartCLIPSeg~\cite{PartCLIPSeg2024}} \end{subfigure}
    \begin{subfigure}[t]{0.195\textwidth} \caption{PartCATSeg} \end{subfigure}
    \vspace{1em}
    \caption{
        Qualitative evaluation of zero-shot part segmentation on PartImageNet in the \textbf{Oracle-Obj} configuration.
    }
    \label{suppl_fig:vis_pred_qualitative_partimagenet_oracle}
\end{figure*}

The following~\Cref{suppl_fig:vis_pred_qualitative_partimagenet_oracle} illustrates the PartImageNet prediction results of PartCATSeg in Oracle-Obj setup.
PartCATSeg demonstrates superior performance compared to other baselines, accurately predicting appropriate boundaries.

\subsection{Cost Visualization}

\begin{figure*}[ht]
    \centering
    \small
    \begin{subfigure}{0.162\textwidth} \centering
        \includegraphics[width=1\linewidth]{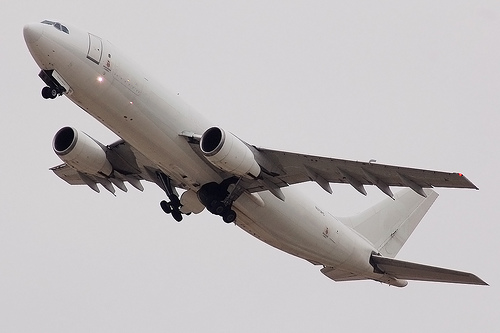}
        \caption{Original Image}
    \end{subfigure}
    \hfill
    \begin{subfigure}{0.162\textwidth} \centering
        \includegraphics[width=1\linewidth]{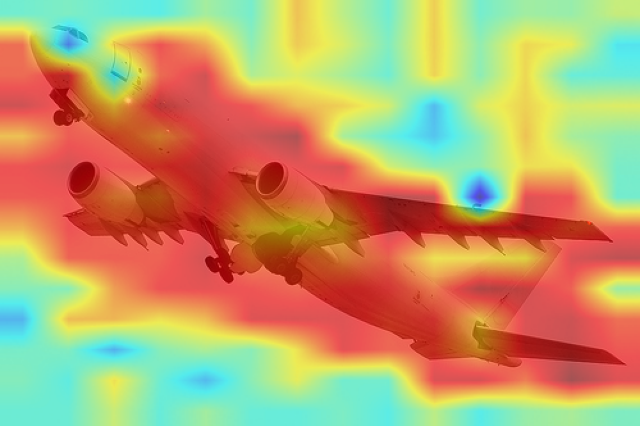}
        \caption{``wheel''}
    \end{subfigure}
    \hfill
    \begin{subfigure}{0.162\textwidth} \centering
        \includegraphics[width=1\linewidth]{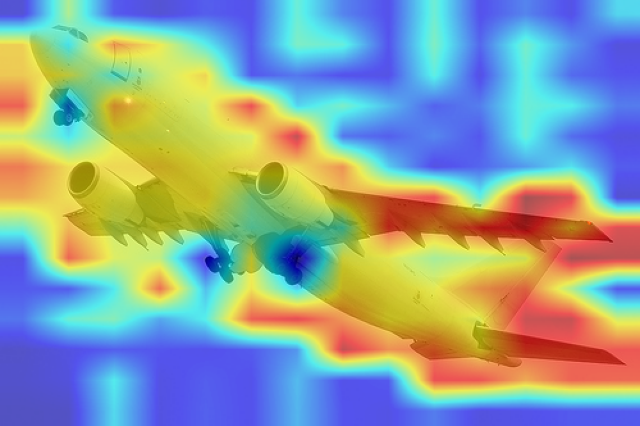}
        \caption{``wing''}
    \end{subfigure}
    \hfill
    \begin{subfigure}{0.162\textwidth} \centering
        \includegraphics[width=1\linewidth]{assets/intro/2009_001385.jpg}
        \caption{Original Image}
    \end{subfigure}
    \hfill
    \begin{subfigure}{0.162\textwidth} \centering
        \includegraphics[width=1\linewidth]{assets/intro/clip_surgery/bird_224/head.png}
        \caption{``head''}
    \end{subfigure}
    \hfill
    \begin{subfigure}{0.162\textwidth} \centering
        \includegraphics[width=1\linewidth]{assets/intro/clip_surgery/bird_224/wing.png}
        \caption{``wing''}
    \end{subfigure}
    \hfill
    \begin{subfigure}{0.162\textwidth} \centering
        \includegraphics[width=1\linewidth]{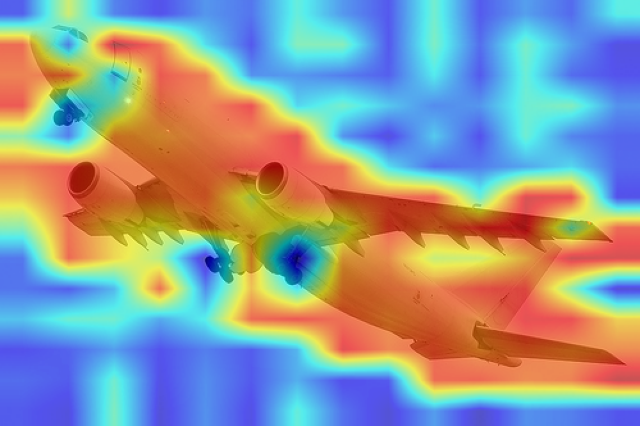}
        \caption{``aeroplane''}
    \end{subfigure}
    \hfill
    \begin{subfigure}{0.162\textwidth} \centering
        \includegraphics[width=1\linewidth]{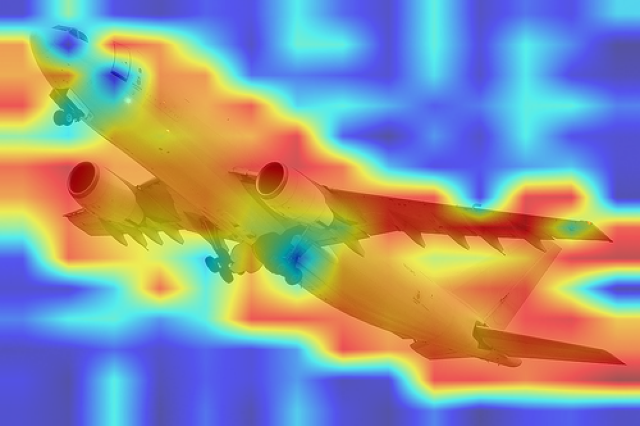}
        \caption{``aeroplane's wheel''}
    \end{subfigure}
    \hfill
    \begin{subfigure}{0.162\textwidth} \centering
        \includegraphics[width=1\linewidth]{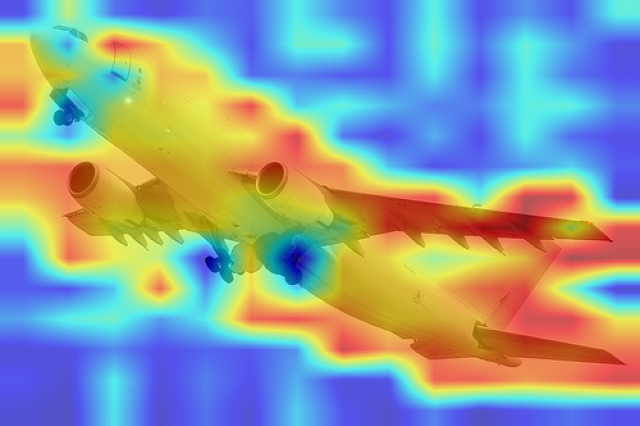}
        \caption{``aeroplane's wing''}
    \end{subfigure}
    \hfill
    \begin{subfigure}{0.162\textwidth} \centering
        \includegraphics[width=1\linewidth]{assets/intro/clip_surgery/bird_224/bird.png}
        \caption{``bird''}
    \end{subfigure}
    \hfill
    \begin{subfigure}{0.162\textwidth} \centering
        \includegraphics[width=1\linewidth]{assets/intro/clip_surgery/bird_224/bird_head.png}
        \caption{``bird's head''}
    \end{subfigure}
    \hfill
    \begin{subfigure}{0.162\textwidth} \centering
        \includegraphics[width=1\linewidth]{assets/intro/clip_surgery/bird_224/bird_wing.png}
        \caption{``bird's wing''}
    \end{subfigure}
    \hfill
    \caption{
        \textbf{CLIP Image-Text Similarity Visualization for Object-Level and Part-Level Text.}
        The visualization compares the frozen CLIP image-text similarity between object-level and part-level text descriptions.
        (a), (d) show the original images; (b), (c), (e), (f) depict the part-level similarities for terms such as "wheel" and "wing" while (g)-(l) show object-specific parts.
        The stronger activation for object-level text suggests a dominant focus on the entire object rather than individual parts in the image-text correspondence.
    }
    \label{suppl_fig:imitations_clip_sim}
\end{figure*}

\begin{figure}[ht]
    \centering
    \small
    \begin{subfigure}{0.192\linewidth} \centering \includegraphics[width=1\linewidth]{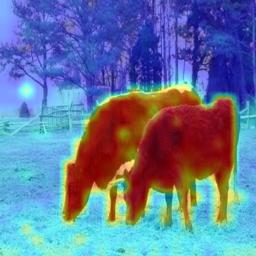} \caption{\scriptsize cow} \end{subfigure} 
    \begin{subfigure}{0.192\linewidth} \centering \includegraphics[width=1\linewidth]{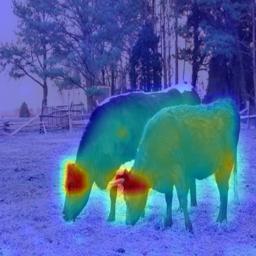} \caption{\scriptsize cow's ear} \end{subfigure}
    \begin{subfigure}{0.192\linewidth} \centering \includegraphics[width=1\linewidth]{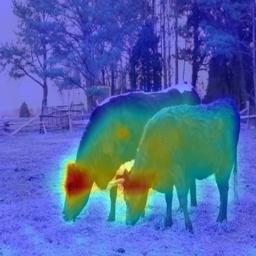} \caption{\scriptsize cow's eye} \end{subfigure}
    \begin{subfigure}{0.192\linewidth} \centering \includegraphics[width=1\linewidth]{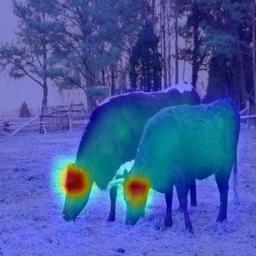} \caption{\scriptsize cow's head} \end{subfigure}
    \begin{subfigure}{0.192\linewidth} \centering \includegraphics[width=1\linewidth]{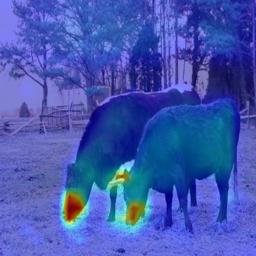} \caption{\scriptsize cow's horn} \end{subfigure}
    \\
    \begin{subfigure}{0.192\linewidth} \centering \includegraphics[width=1\linewidth]{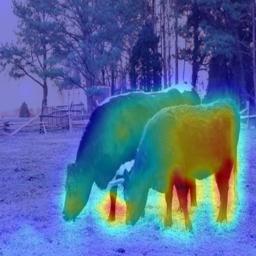} \caption{\scriptsize cow's leg} \end{subfigure}
    \begin{subfigure}{0.192\linewidth} \centering \includegraphics[width=1\linewidth]{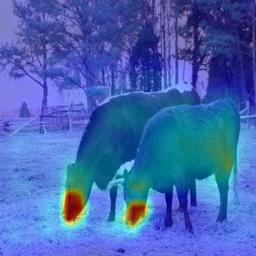} \caption{\tiny cow's muzzle} \end{subfigure}
    \begin{subfigure}{0.192\linewidth} \centering \includegraphics[width=1\linewidth]{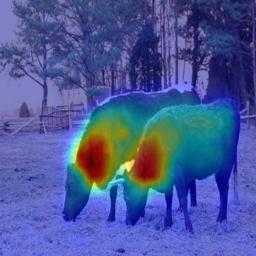} \caption{\scriptsize cow's neck} \end{subfigure}
    \begin{subfigure}{0.192\linewidth} \centering \includegraphics[width=1\linewidth]{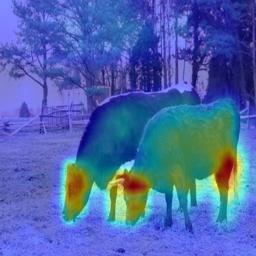} \caption{\scriptsize cow's tail} \end{subfigure}
    \begin{subfigure}{0.192\linewidth} \centering \includegraphics[width=1\linewidth]{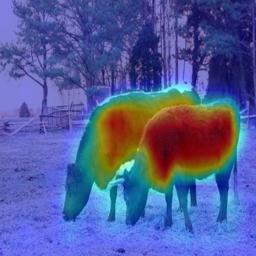} \caption{\scriptsize cow's torso} \end{subfigure}

    \begin{subfigure}{0.192\linewidth} \centering \includegraphics[width=1\linewidth]{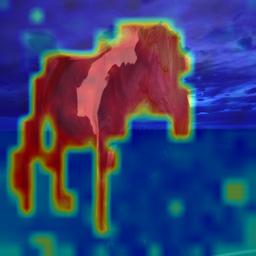}\caption{\scriptsize horse} \end{subfigure}
    \begin{subfigure}{0.192\linewidth} \centering \includegraphics[width=1\linewidth]{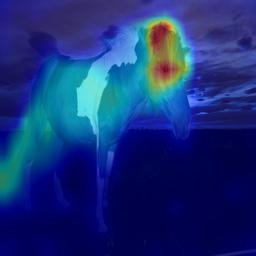}\caption{\scriptsize horse's ear} \end{subfigure}
    \begin{subfigure}{0.192\linewidth} \centering \includegraphics[width=1\linewidth]{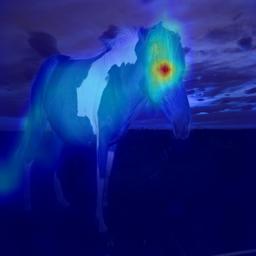}\caption{\scriptsize horse's eye} \end{subfigure}
    \begin{subfigure}{0.192\linewidth} \centering \includegraphics[width=1\linewidth]{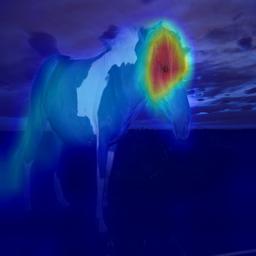}\caption{\scriptsize horse's head} \end{subfigure}
    \begin{subfigure}{0.192\linewidth} \centering \includegraphics[width=1\linewidth]{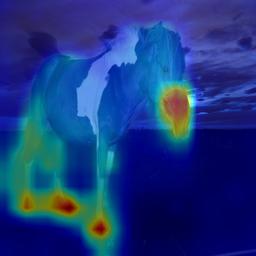}\caption{\scriptsize horse's hoof} \end{subfigure}
    \\
    \begin{subfigure}{0.192\linewidth} \centering \includegraphics[width=1\linewidth]{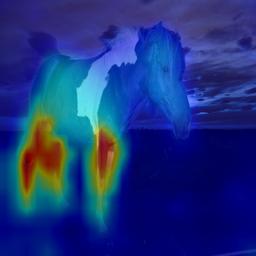}\caption{\scriptsize horse's leg} \end{subfigure}
    \begin{subfigure}{0.192\linewidth} \centering \includegraphics[width=1\linewidth]{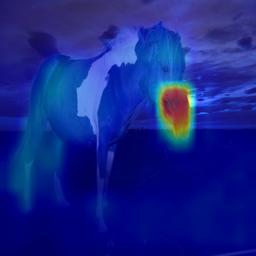}\caption{\tiny horse's muzzle} \end{subfigure}
    \begin{subfigure}{0.192\linewidth} \centering \includegraphics[width=1\linewidth]{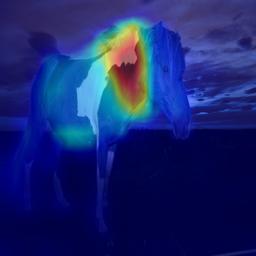}\caption{\scriptsize horse's neck} \end{subfigure}
    \begin{subfigure}{0.192\linewidth} \centering \includegraphics[width=1\linewidth]{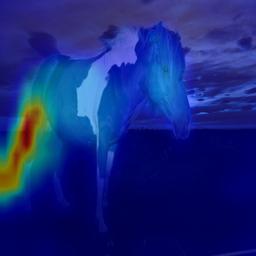}\caption{\scriptsize horse's tail} \end{subfigure}
    \begin{subfigure}{0.192\linewidth} \centering \includegraphics[width=1\linewidth]{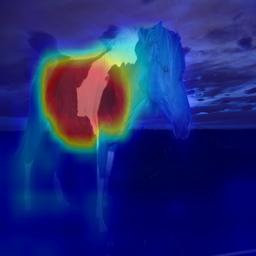}\caption{\scriptsize horse's torso} \end{subfigure}

    \caption{
        \textbf{Cost Volume Visualization After Additional Training.}
        Cost volume visualization showed that PartCATSeg significantly enhanced fine-grained alignment.
    }
    \label{suppl_fig:cost_volume_2_cow}
\end{figure}

The following~\Cref{suppl_fig:cost_volume_1_person} and ~\Cref{suppl_fig:cost_volume_2_cow} present the cost (correspondence) visualization after training.
It visualizes the object-specific part correspondence between the caption text and the image.
Unlike (pretrained) CLIP Image-Text Similarity Visualization (\Cref{fig:intro_limitations_clip_sim}) discussed in~\Cref{sec:intro} and~\Cref{suppl_fig:cost_volume_1_person_before_512}, as well as ~\Cref{suppl_fig:imitations_clip_sim}, the cost volume demonstrates significant improvement in fine-grained alignment after training, as illustrated in~\Cref{suppl_fig:cost_volume_1_person} and~\Cref{suppl_fig:cost_volume_2_cow}.

\section{Additional Ablation Study}

\subsection{Compositional Loss}

\begin{figure}[ht]
    \begin{subfigure}[b]{0.48\linewidth}
        \centering
        \begin{tikzpicture}
            \node[anchor=south west,inner sep=0] (img1) {%
                \includegraphics[width=\linewidth]{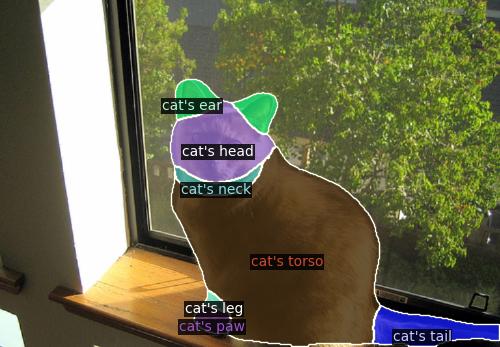}%
            };
            \draw[red, line width=1pt] 
                ($(img1.south west)+(35pt,30pt)$) rectangle ($(img1.north east)+(-47pt,-27pt)$);
        \end{tikzpicture}
         \caption{w/o $\mathcal{L}_{\texttt{comp}}$}
         \label{suppl_subfig:ablation_wo_comp_loss}
    \end{subfigure}
    \hfill
    \begin{subfigure}[b]{0.48\linewidth}
        \centering
        \begin{tikzpicture}
            \node[anchor=south west,inner sep=0] (img2) {%
                \includegraphics[width=\linewidth]{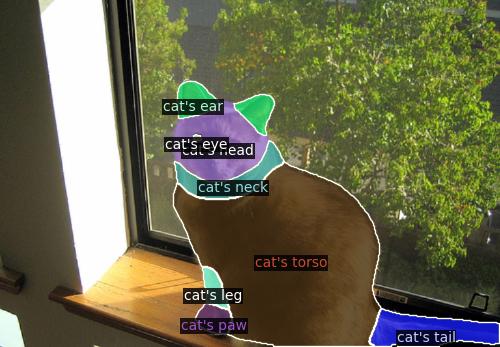}%
            };
            \draw[red, line width=1pt] 
                ($(img2.south west)+(35pt,30pt)$) rectangle ($(img2.north east)+(-47pt,-27pt)$);
        \end{tikzpicture}
        \caption{w/ $\mathcal{L}_{\texttt{comp}}$}
        \label{suppl_subfig:ablation_w_comp_loss}
    \end{subfigure}
    \\
    \begin{subfigure}[b]{0.23\linewidth}
        \includegraphics[width=\linewidth]{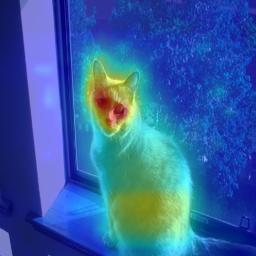}
        \caption{``cat's eye''}
        \label{suppl_subfig:ablation_wo_comp_loss_cat_eye_cost}
    \end{subfigure}
    \begin{subfigure}[b]{0.23\linewidth}
        \includegraphics[width=\linewidth]{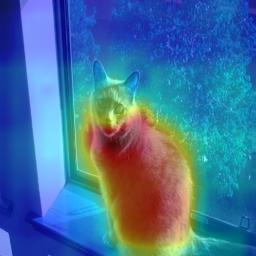}
        \caption{``cat's neck''}
        \label{suppl_subfig:ablation_wo_comp_loss_cat_neck_cost}
    \end{subfigure}
    \hfill
    \begin{subfigure}[b]{0.23\linewidth}
        \includegraphics[width=\linewidth]{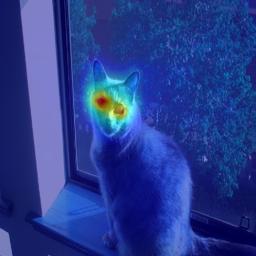}
        \caption{``cat's eye''}
        \label{suppl_subfig:ablation_w_comp_loss_cat_eye_cost}
    \end{subfigure}
    \hfill
    \begin{subfigure}[b]{0.23\linewidth}
        \includegraphics[width=\linewidth]{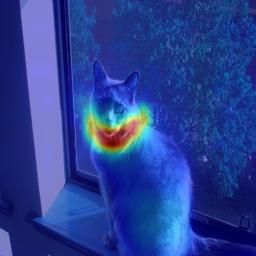}
        \caption{``cat's neck''}
        \label{suppl_subfig:ablation_w_comp_loss_cat_neck_cost}
    \end{subfigure}
    \caption{
        \textbf{Ablation on Compositional Loss.} (a) and (b) show segmentation results without and with $\mathcal{L}_{\texttt{comp}}$, respectively.
        (c) and (d) show less defined cost volumes without $\mathcal{L}_{\texttt{comp}}$, while (e) and (f) reveal more exclusive similarities in the cost volumes with $\mathcal{L}_{\texttt{comp}}$. Notably, ``cat's eye'' is successfully segmented with the inclusion of $\mathcal{L}_{\texttt{comp}}$.
    }
    \label{suppl_fig:compositional_loss}
    \vspace{-1em}
\end{figure}

As detailed in~\Cref{subsec:Cost as Compositional Components}, parts are not only compositional components that constitute an object but also maintain relationships with adjacent parts.
Previous methodologies have proposed learning strategies that consider granularity at two levels—the object and its parts.
However, they have not focused on the composition of object-specific parts within the object and the relationships between these parts.
This limitation often results in small parts, such as ``cat's eye'' and ``cat's neck'', being undetected within the context of a larger object, as shown in~\Cref{suppl_subfig:ablation_wo_comp_loss}.
We hypothesize that this issue arises because certain parts become excessively dominant relative to other parts they encompass, leading to a failure to recognize their spatial and compositional relationships.
By introducing compositional loss, our model better identifies and segments smaller or more discriminative parts.
As illustrated in~\Cref{suppl_fig:compositional_loss}, the inclusion of compositional loss resolves issues of overlapping or diffuse cost volumes, as visualized in~\Cref{suppl_subfig:ablation_wo_comp_loss_cat_eye_cost} and~\Cref{suppl_subfig:ablation_wo_comp_loss_cat_neck_cost}.
These visualizations highlight how compositional loss sharpens the focus on specific parts, mitigating the spread of cost volume and ensuring better part-level segmentation.

\begin{figure}[ht]
    \centering
    \small
    \begin{subfigure}{0.48\linewidth}
        \centering
        \begin{tikzpicture}
            \node[anchor=south west,inner sep=0] (img1) 
              {\includegraphics[width=1\linewidth]{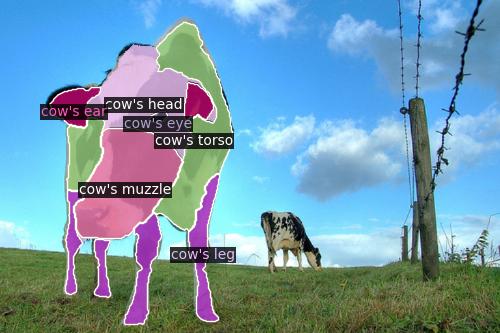}};
            \draw[red, line width=1pt] 
                ($(img1.south west)+(23pt,40pt)$) rectangle ($(img1.north east)+(-67pt,-20pt)$);
        \end{tikzpicture}
        \caption{\scriptsize Ground-truth}
    \end{subfigure} 
    \begin{subfigure}{0.48\linewidth}
        \centering
        \begin{tikzpicture}
            \node[anchor=south west,inner sep=0] (img2) 
              {\includegraphics[width=1\linewidth]{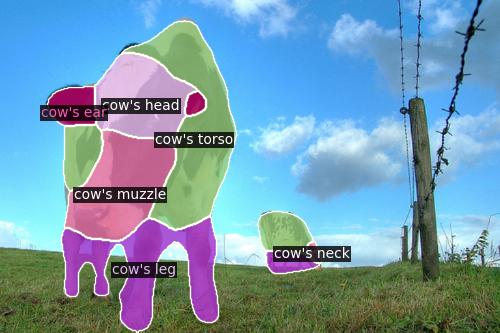}};
            \draw[red, line width=1pt] 
                ($(img2.south west)+(23pt,40pt)$) rectangle ($(img2.north east)+(-67pt,-20pt)$);
        \end{tikzpicture}
        \caption{\scriptsize w/o $\mathcal{L}_{\texttt{comp}}$}
    \end{subfigure}\\[1ex]
    \begin{subfigure}{0.48\linewidth}
        \centering
        \begin{tikzpicture}
            \node[anchor=south west,inner sep=0] (img3) 
              {\includegraphics[width=1\linewidth]{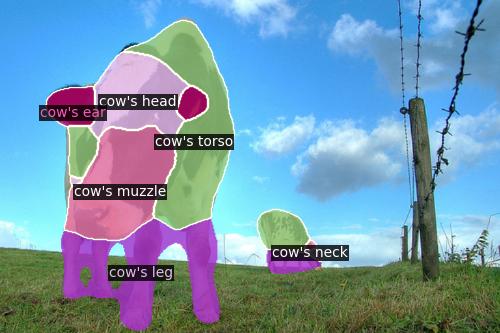}};
            \draw[red, line width=1pt] 
                ($(img3.south west)+(23pt,40pt)$) rectangle ($(img3.north east)+(-67pt,-20pt)$);
        \end{tikzpicture}
        \caption{\scriptsize w/ $\mathcal{L}_{\texttt{comp}}$-L1}
    \end{subfigure}
    \begin{subfigure}{0.48\linewidth}
        \centering
        \begin{tikzpicture}
            \node[anchor=south west,inner sep=0] (img4) 
              {\includegraphics[width=1\linewidth]{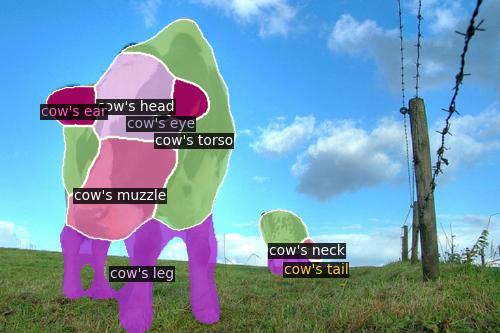}};
            \draw[red, line width=1pt] 
                ($(img4.south west)+(23pt,40pt)$) rectangle ($(img4.north east)+(-67pt,-20pt)$);
        \end{tikzpicture}
        \caption{\scriptsize w/ $\mathcal{L}_{\texttt{comp}}$-SM}
    \end{subfigure}
    \\
    \begin{subfigure}{0.48\linewidth}
        \centering
        \begin{tikzpicture}
            \node[anchor=south west,inner sep=0] (img5) 
              {\includegraphics[width=1\linewidth]{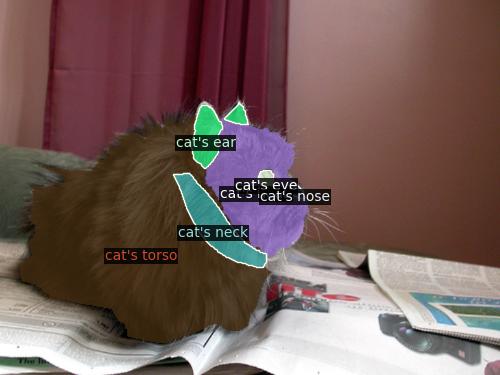}};
            \draw[red, line width=1pt] 
                ($(img5.south west)+(38pt,25pt)$) rectangle ($(img5.north east)+(-42pt,-30pt)$);
        \end{tikzpicture}
        \caption{\scriptsize Ground-truth}
    \end{subfigure}
    \begin{subfigure}{0.48\linewidth}
        \centering
        \begin{tikzpicture}
            \node[anchor=south west,inner sep=0] (img6) 
              {\includegraphics[width=1\linewidth]{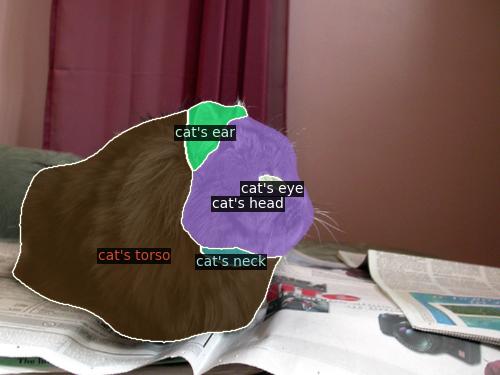}};
            \draw[red, line width=1pt] 
                ($(img6.south west)+(38pt,25pt)$) rectangle ($(img6.north east)+(-42pt,-30pt)$);
        \end{tikzpicture}
        \caption{\scriptsize w/o $\mathcal{L}_{\texttt{comp}}$}
    \end{subfigure}\\[1ex]
    \begin{subfigure}{0.48\linewidth}
        \centering
        \begin{tikzpicture}
            \node[anchor=south west,inner sep=0] (img7) 
              {\includegraphics[width=1\linewidth]{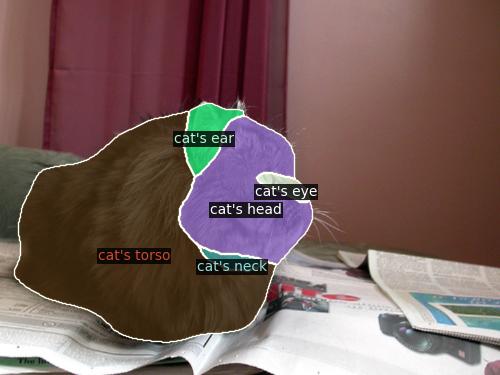}};
            \draw[red, line width=1pt] 
                ($(img7.south west)+(38pt,25pt)$) rectangle ($(img7.north east)+(-42pt,-30pt)$);
        \end{tikzpicture}
        \caption{\scriptsize w/ $\mathcal{L}_{\texttt{comp}}$-L1}
    \end{subfigure}
    \begin{subfigure}{0.48\linewidth}
        \centering
        \begin{tikzpicture}
            \node[anchor=south west,inner sep=0] (img8) 
              {\includegraphics[width=1\linewidth]{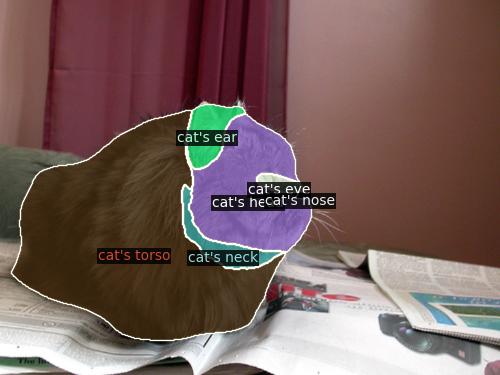}};
            \draw[red, line width=1pt] 
                ($(img8.south west)+(38pt,25pt)$) rectangle ($(img8.north east)+(-42pt,-30pt)$);
        \end{tikzpicture}
        \caption{\scriptsize w/ $\mathcal{L}_{\texttt{comp}}$-SM}
    \end{subfigure}
    \caption{
        \textbf{Qualitative Ablation on Compositional Loss for Pred-All setting in Pascal-Part-116.}
        Applying compositional loss improves segmentation performance, particularly for challenging parts like ``cow's eye'' or ``cat's nose''. Furthermore, using softmax normalization in the compositional loss ($\mathcal{L}_{\texttt{comp}}$-SM) outperforms L1 normalization ($\mathcal{L}_{\texttt{comp}}$-L1) by better capturing these fine-grained parts such as ``cat's neck''.
    }
    \label{suppl_fig:quali_ablation_comp_voc}
    \vspace{-1.0em}
\end{figure}

\begin{figure}[ht]
    \centering
    \small
    \begin{subfigure}{0.325\linewidth}
        \centering
        \begin{tikzpicture}
            \node[anchor=south west,inner sep=0] (img1) 
              {\includegraphics[width=1\linewidth]{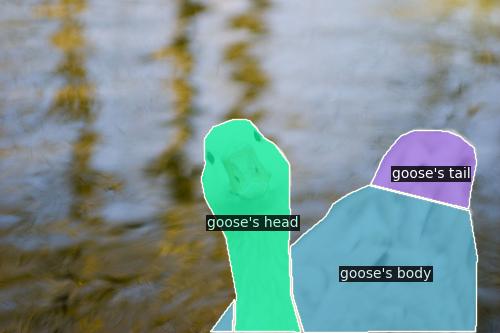}};
            \draw[red, line width=1pt] 
              ($(img1.south west)+(0.75\linewidth,0.25\linewidth)$) rectangle 
              ($(img1.north east)+(-0.05\linewidth,-0.25\linewidth)$);
        \end{tikzpicture}
    \end{subfigure} 
    \begin{subfigure}{0.325\linewidth}
        \centering
        \begin{tikzpicture}
            \node[anchor=south west,inner sep=0] (img2) 
              {\includegraphics[width=1\linewidth]{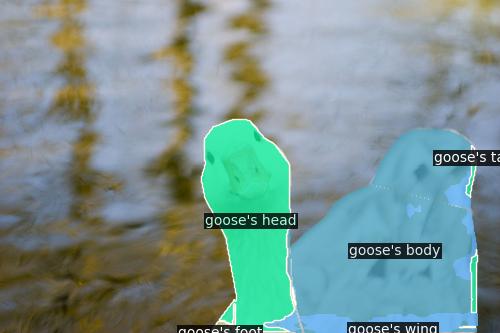}};
            \draw[red, line width=1pt] 
              ($(img2.south west)+(0.75\linewidth,0.25\linewidth)$) rectangle 
              ($(img2.north east)+(-0.05\linewidth,-0.25\linewidth)$);
        \end{tikzpicture}
    \end{subfigure}
    \begin{subfigure}{0.325\linewidth}
        \centering
        \begin{tikzpicture}
            \node[anchor=south west,inner sep=0] (img3) 
              {\includegraphics[width=1\linewidth]{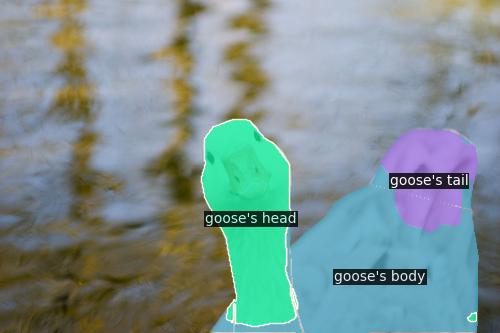}};
            \draw[red, line width=1pt] 
              ($(img3.south west)+(0.75\linewidth,0.25\linewidth)$) rectangle 
              ($(img3.north east)+(-0.05\linewidth,-0.25\linewidth)$);
        \end{tikzpicture}
    \end{subfigure}\\[1ex]
    \begin{subfigure}{0.325\linewidth}
        \centering
        \begin{tikzpicture}
            \node[anchor=south west,inner sep=0] (img4) 
              {\includegraphics[width=1\linewidth]{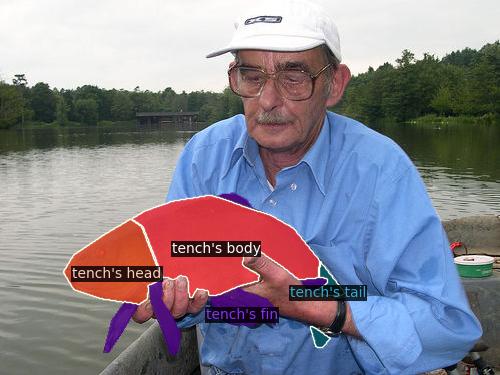}};
            \draw[red, line width=1pt] 
              ($(img4.south west)+(0.56\linewidth,0.07\linewidth)$) rectangle 
              ($(img4.north east)+(-0.25\linewidth,-0.52\linewidth)$);
        \end{tikzpicture}
    \end{subfigure}
    \begin{subfigure}{0.325\linewidth}
        \centering
        \begin{tikzpicture}
            \node[anchor=south west,inner sep=0] (img5) 
              {\includegraphics[width=1\linewidth]{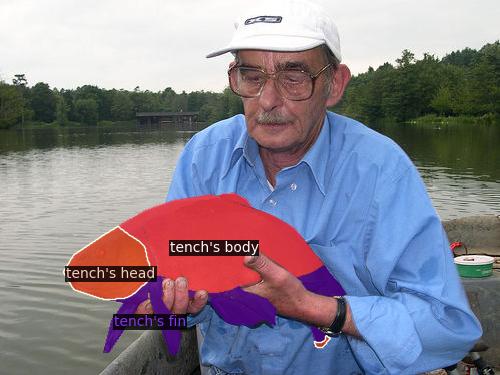}};
            \draw[red, line width=1pt] 
              ($(img5.south west)+(0.56\linewidth,0.07\linewidth)$) rectangle 
              ($(img5.north east)+(-0.25\linewidth,-0.52\linewidth)$);
        \end{tikzpicture}
    \end{subfigure}
    \begin{subfigure}{0.325\linewidth}
        \centering
        \begin{tikzpicture}
            \node[anchor=south west,inner sep=0] (img6) 
              {\includegraphics[width=1\linewidth]{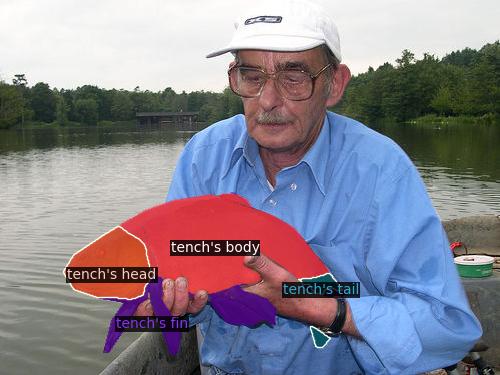}};
            \draw[red, line width=1pt] 
              ($(img6.south west)+(0.56\linewidth,0.07\linewidth)$) rectangle 
              ($(img6.north east)+(-0.25\linewidth,-0.52\linewidth)$);
        \end{tikzpicture}
    \end{subfigure}\\[1ex]
    \begin{subfigure}{0.325\linewidth}
        \centering
        \begin{tikzpicture}
            \node[anchor=south west,inner sep=0] (img7) 
              {\includegraphics[width=1\linewidth]{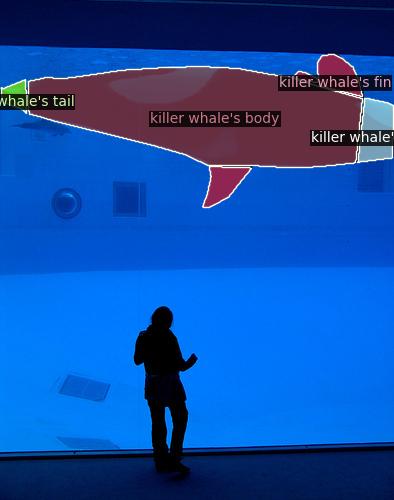}};
            \draw[red, line width=1pt] 
              ($(img7.south west)+(0.85\linewidth,0.82\linewidth)$) rectangle 
              ($(img7.north east)+(-0.01\linewidth,-0.17\linewidth)$);
        \end{tikzpicture}
    \end{subfigure}
    \begin{subfigure}{0.325\linewidth}
        \centering
        \begin{tikzpicture}
            \node[anchor=south west,inner sep=0] (img8) 
              {\includegraphics[width=1\linewidth]{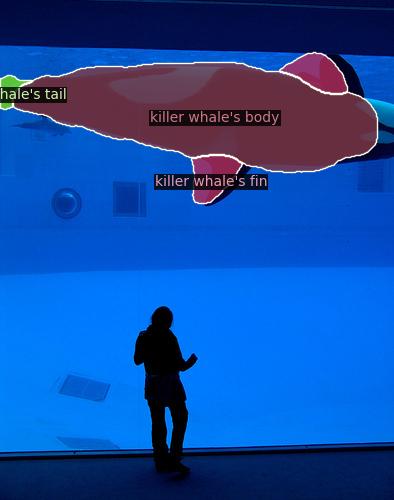}};
            \draw[red, line width=1pt] 
              ($(img8.south west)+(0.85\linewidth,0.82\linewidth)$) rectangle 
              ($(img8.north east)+(-0.01\linewidth,-0.17\linewidth)$);
        \end{tikzpicture}
    \end{subfigure}
    \begin{subfigure}{0.325\linewidth}
        \centering
        \begin{tikzpicture}
            \node[anchor=south west,inner sep=0] (img9) 
              {\includegraphics[width=1\linewidth]{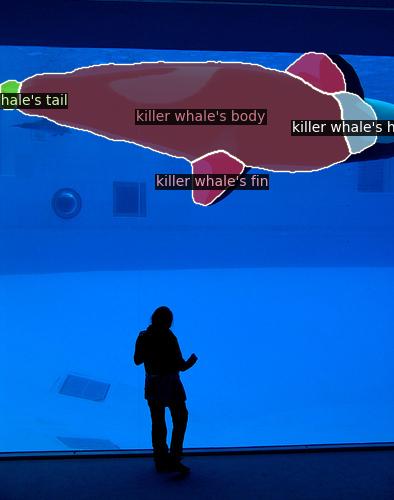}};
            \draw[red, line width=1pt] 
              ($(img9.south west)+(0.85\linewidth,0.82\linewidth)$) rectangle 
              ($(img9.north east)+(-0.01\linewidth,-0.17\linewidth)$);
        \end{tikzpicture}
    \end{subfigure}\\[1ex]
    \begin{subfigure}{0.325\linewidth}
        \centering
        \begin{tikzpicture}
            \node[anchor=south west,inner sep=0] (img10) 
              {\includegraphics[width=1\linewidth]{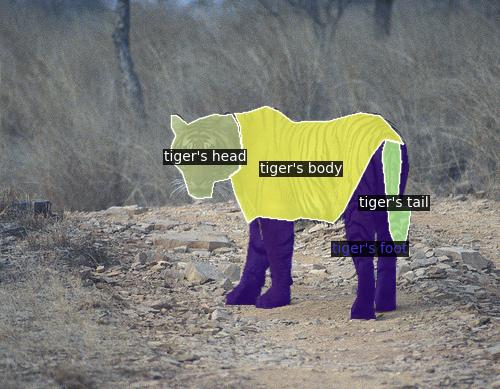}};
            \draw[red, line width=1pt] 
              ($(img10.south west)+(0.7\linewidth,0.2\linewidth)$) rectangle 
              ($(img10.north east)+(-0.15\linewidth,-0.22\linewidth)$);
        \end{tikzpicture}
    \end{subfigure}
    \begin{subfigure}{0.325\linewidth}
        \centering
        \begin{tikzpicture}
            \node[anchor=south west,inner sep=0] (img11) 
              {\includegraphics[width=1\linewidth]{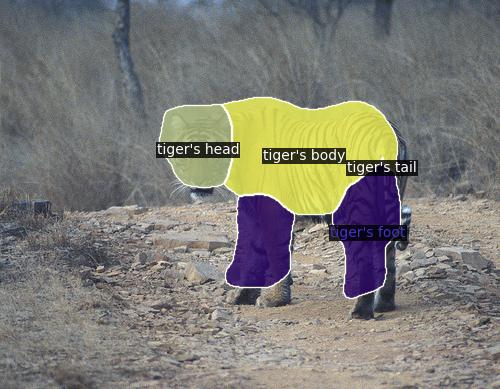}};
            \draw[red, line width=1pt] 
              ($(img11.south west)+(0.7\linewidth,0.2\linewidth)$) rectangle 
              ($(img11.north east)+(-0.15\linewidth,-0.22\linewidth)$);
        \end{tikzpicture}
    \end{subfigure}
    \begin{subfigure}{0.325\linewidth}
        \centering
        \begin{tikzpicture}
            \node[anchor=south west,inner sep=0] (img12) 
              {\includegraphics[width=1\linewidth]{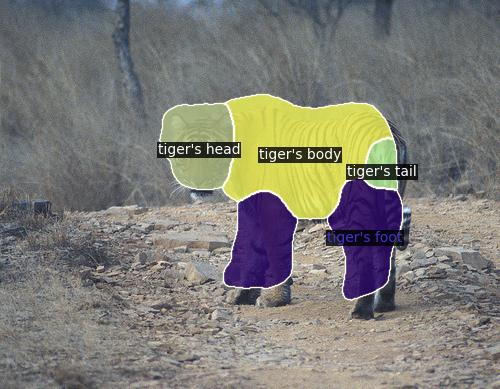}};
            \draw[red, line width=1pt] 
              ($(img12.south west)+(0.7\linewidth,0.2\linewidth)$) rectangle 
              ($(img12.north east)+(-0.15\linewidth,-0.22\linewidth)$);
        \end{tikzpicture}
    \end{subfigure}\\[1ex]
    \begin{subfigure}{0.325\linewidth}
        \centering
        \begin{tikzpicture}
            \node[anchor=south west,inner sep=0] (img13) 
              {\includegraphics[width=1\linewidth]{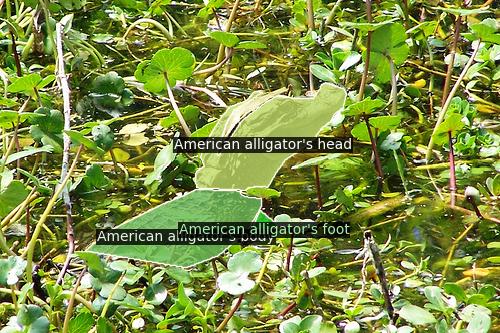}};
            \draw[red, line width=1pt] 
              ($(img13.south west)+(0.4\linewidth,0.13\linewidth)$) rectangle 
              ($(img13.north east)+(-0.26\linewidth,-0.38\linewidth)$);
        \end{tikzpicture}
        \caption{\scriptsize Ground-truth}
    \end{subfigure}
    \begin{subfigure}{0.325\linewidth}
        \centering
        \begin{tikzpicture}
            \node[anchor=south west,inner sep=0] (img14) 
              {\includegraphics[width=1\linewidth]{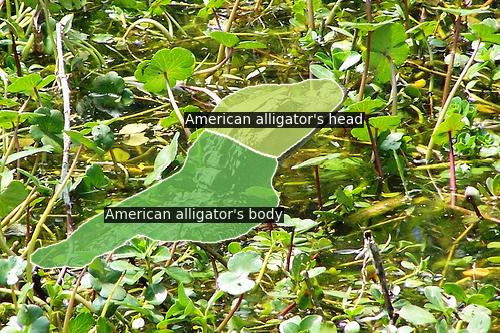}};
            \draw[red, line width=1pt] 
              ($(img14.south west)+(0.4\linewidth,0.13\linewidth)$) rectangle 
              ($(img14.north east)+(-0.26\linewidth,-0.38\linewidth)$);
        \end{tikzpicture}
        \caption{\scriptsize w/o $\mathcal{L}_{\texttt{comp}}$}
    \end{subfigure}
    \begin{subfigure}{0.325\linewidth}
        \centering
        \begin{tikzpicture}
            \node[anchor=south west,inner sep=0] (img15) 
              {\includegraphics[width=1\linewidth]{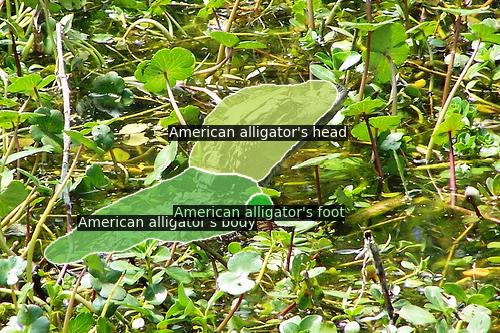}};
            \draw[red, line width=1pt] 
              ($(img15.south west)+(0.4\linewidth,0.13\linewidth)$) rectangle 
              ($(img15.north east)+(-0.26\linewidth,-0.38\linewidth)$);
        \end{tikzpicture}
        \caption{\scriptsize w/ $\mathcal{L}_{\texttt{comp}}$-SM}
    \end{subfigure}

    \caption{
        \textbf{Qualitative Ablation on Compositional Loss in PartImageNet.} The first two rows show results in the Oracle-Obj setting, while the bottom three rows are in the Pred-All setting. Segmentation with $\mathcal{L}_{\texttt{comp}}$ captures finer part relationships, such as ``goose's tail,'' ``tench's tail,'' ``killer whale's head,'' ``tiger's tail,'' and ``American alligator's foot,'' compared to results without $\mathcal{L}_{\texttt{comp}}$. 
    }
    \vspace{-1.0em}
    \label{suppl_fig:quali_ablation_comp_partimagenet}
\end{figure}

Furthermore, as shown in~\Cref{suppl_fig:quali_ablation_comp_voc}, compositional loss consistently improves segmentation performance across datasets. For example, in Pascal-Part-116, it enhances the segmentation of challenging parts such as cow's eye'' or cat's nose,'' which are often difficult to detect. Specifically, the comparison demonstrates that softmax normalization in the compositional loss ($\mathcal{L}_{\texttt{comp}}$-SM) outperforms L1 normalization ($\mathcal{L}_{\texttt{comp}}$-L1). Similarly, as shown in ~\Cref{suppl_fig:quali_ablation_comp_partimagenet}, compositional loss demonstrates its effectiveness in capturing fine-grained part relationships in PartImageNet, such as ``goose's tail,'' ``tench's tail,'' and ``killer whale's head.'' This improvement underscores the importance of explicitly modeling compositional relationships in part segmentation, particularly for smaller or less distinct parts.

The effectiveness of compositional loss is further validated through quantitative results, as shown in~\Cref{suppl_tab:ablation_comp_loss}. The inclusion of $\mathcal{L}_{\texttt{comp}}$ improves performance across both the Pred-All and Oracle-Obj settings on PartImageNet. Notably, it enhances the harmonic IoU for unseen parts, demonstrating its ability to better capture fine-grained compositional relationships and improve segmentation consistency for challenging parts.

\begin{table}[ht]
\centering
\small
\resizebox{\linewidth}{!}
{
\begin{tabular}{lcccccc}
\toprule
\multicolumn{1}{c}{\multirow{2}{*}{\makecell{Compositional\\Loss}}} & \multicolumn{3}{c}{Pred-All} & \multicolumn{3}{c}{Oracle-Obj} \\ \cmidrule(l){2-4} \cmidrule(l){5-7} 
                        & Seen & Unseen & h-IoU       & Seen  & Unseen & h-IoU       \\ \midrule
w/o $\mathcal{L}_{\texttt{comp}}$  & \textbf{59.21} & 50.75 & 54.66      & 72.17 & 68.42  & 70.24     \\
w/ $\mathcal{L}_{\texttt{comp}}$                       & 57.33 & \textbf{53.07} & \textbf{55.12} & \textbf{73.83} & \textbf{71.52}  & \textbf{72.66}   \\ 
\bottomrule
\end{tabular}
}
\caption{Impact of Compositional Loss on PartImageNet}
\label{suppl_tab:ablation_comp_loss}
\end{table}

\section{Datasets Details}
\label{suppl_sec:datasets_details}

\subsection{Pascal-Part-116}

\noindent
Pascal-Part-116~\cite{wei2024ov_OV_PARTS} is a modified version of the Pascal-Part~\cite{chen2014detect_PascalPart} dataset specifically designed for Open-Vocabulary Part Segmentation tasks.
The dataset includes a mix of base and novel categories, focusing on diverse object-level classes.
It features novel classes for the object categories ``bird'', ``car'', ``dog'', ``sheep'', and ``motorbike''.
Additionally, based on object-specific part categories, the dataset comprises 74 base categories and 42 novel categories, offering a comprehensive benchmark for evaluating segmentation models.
Detailed information about the base and novel classes can be found in~\Cref{suppl_tab:pascal-part-116}.

\begin{table}[ht]
    \vspace{0.5em}
    \centering
    \begin{small}
    \resizebox{\linewidth}{!}
    {
       \begin{tabular}{lllll}
            \toprule
            \multicolumn{5}{c}{Pascal-Part-116 Object-specific Part Categories} \\
            \cmidrule(l){1-5}
            \multicolumn{5}{c}{\textcolor{teal}{Base Categories (74)}} \\
            \cmidrule(l){1-5}
                aeroplane's body & aeroplane's stern & aeroplane's wing & aeroplane's tail & aeroplane's engine \\
                aeroplane's wheel & bicycle's wheel & bicycle's saddle & bicycle's handlebar & bicycle's chainwheel \\
                bicycle's headlight & bottle's body & bottle's cap & bus's wheel & bus's headlight \\
                bus's front & bus's side & bus's back & bus's roof & bus's mirror \\
                bus's license plate & bus's door & bus's window & cat's tail & cat's head \\
                cat's eye & cat's torso & cat's neck & cat's leg & cat's nose \\
                cat's paw & cat's ear & cow's tail & cow's head & cow's eye \\
                cow's torso & cow's neck & cow's leg & cow's ear & cow's muzzle \\
                cow's horn & horse's tail & horse's head & horse's eye & horse's torso \\
                horse's neck & horse's leg & horse's ear & horse's muzzle & horse's hoof \\
                person's head & person's eye & person's torso & person's neck & person's leg \\
                person's foot & person's nose & person's ear & person's eyebrow & person's mouth \\
                person's hair & person's lower arm & person's upper arm & person's hand & pottedplant's pot \\
                pottedplant's plant & train's headlight & train's head & train's front & train's side \\
                train's back & train's roof & train's coach & tvmonitor's screen &  \\
            \cmidrule(l){1-5}
            \multicolumn{5}{c}{\textcolor{violet}{Novel Categories (42)}} \\
            \cmidrule(l){1-5}
                bird's wing & bird's tail & bird's head & bird's eye & bird's beak \\
                bird's torso & bird's neck & bird's leg & bird's foot & car's wheel \\
                car's headlight & car's front & car's side & car's back & car's roof \\
                car's mirror & car's license plate & car's door & car's window & dog's tail \\
                dog's head & dog's eye & dog's torso & dog's neck & dog's leg \\
                dog's nose & dog's paw & dog's ear & dog's muzzle & motorbike's wheel \\
                motorbike's saddle & motorbike's handlebar & motorbike's headlight & sheep's tail & sheep's head \\
                sheep's eye & sheep's torso & sheep's neck & sheep's leg & sheep's ear \\
                sheep's muzzle & sheep's horn &  \\
            \bottomrule
            \end{tabular}
    }
    \caption{List of object-specific classes in \textbf{Pascal-Part-116}.}
    \vspace{-1.0em}
    \label{suppl_tab:pascal-part-116}
\end{small}
\end{table}

\noindent


\subsection{ADE20K-Part-234}

ADE20K-Part-234~\cite{wei2024ov_OV_PARTS} is an adapted version of the ADE20K dataset~\cite{zhou2017scene_ADE20K} tailored for Open-Vocabulary Part Segmentation tasks. The dataset includes a mix of base and novel categories, with a focus on a diverse range of object-level classes.
It features novel classes for the object categories ``bench'', ``bus'', ``fan'', ``desk'', ``stool'', ``truck'', ``van'', ``swivel chair'', ``oven'', ``ottoman," and ``kitchen island''.
Additionally, based on object-specific part categories, the dataset comprises 176 base categories and 58 novel categories, providing a robust benchmark for evaluating segmentation models. The dataset presents additional challenges due to its diverse categories and the frequent appearance of small parts, which require precise part segmentation.

\begin{table}[ht]
    \vspace{0.5em}
    \centering
    \begin{small}
    \resizebox{\linewidth}{!}
    {
       \begin{tabular}{lllll}
            \toprule
            \multicolumn{5}{c}{ADE20K-Part-234 Object-specific Part Categories} \\
            \cmidrule(l){1-5}
            \multicolumn{5}{c}{\textcolor{teal}{Base Categories (176)}} \\
            \cmidrule(l){1-5}
                person's arm & person's back & person's foot & person's gaze & person's hand \\
                person's head & person's leg & person's neck & person's torso & door's door frame \\
                door's handle & door's knob & door's panel & clock's face & clock's frame \\
                toilet's bowl & toilet's cistern & toilet's lid & cabinet's door & cabinet's drawer \\
                cabinet's front & cabinet's shelf & cabinet's side & cabinet's skirt & cabinet's top \\
                sink's bowl & sink's faucet & sink's pedestal & sink's tap & sink's top \\
                lamp's arm & lamp's base & lamp's canopy & lamp's column & lamp's cord \\
                lamp's highlight & lamp's light source & lamp's shade & lamp's tube & sconce's arm \\
                sconce's backplate & sconce's highlight & sconce's light source & sconce's shade & chair's apron \\
                chair's arm & chair's back & chair's base & chair's leg & chair's seat \\
                chair's seat cushion & chair's skirt & chair's stretcher & chest of drawers's apron & chest of drawers's door \\
                chest of drawers's drawer & chest of drawers's front & chest of drawers's leg & chandelier's arm & chandelier's bulb \\
                chandelier's canopy & chandelier's chain & chandelier's cord & chandelier's highlight & chandelier's light source \\
                chandelier's shade & bed's footboard & bed's headboard & bed's leg & bed's side rail \\
                table's apron & table's drawer & table's leg & table's shelf & table's top \\
                table's wheel & armchair's apron & armchair's arm & armchair's back & armchair's back pillow \\
                armchair's leg & armchair's seat & armchair's seat base & armchair's seat cushion & shelf's door \\
                shelf's drawer & shelf's front & shelf's shelf & coffee table's leg & coffee table's top \\
                sofa's arm & sofa's back & sofa's back pillow & sofa's leg & sofa's seat base \\
                sofa's seat cushion & sofa's skirt & computer's computer case & computer's keyboard & computer's monitor \\
                computer's mouse & wardrobe's door & wardrobe's drawer & wardrobe's front & wardrobe's leg \\
                wardrobe's mirror & wardrobe's top & car's bumper & car's door & car's headlight \\
                car's hood & car's license plate & car's logo & car's mirror & car's wheel \\
                car's window & car's wiper & cooking stove's burner & cooking stove's button panel & cooking stove's door \\
                cooking stove's drawer & cooking stove's oven & cooking stove's stove & microwave's button panel & microwave's door \\
                microwave's front & microwave's side & microwave's top & microwave's window & refrigerator's button panel \\
                refrigerator's door & refrigerator's drawer & refrigerator's side & dishwasher's button panel & dishwasher's handle \\
                dishwasher's skirt & bookcase's door & bookcase's drawer & bookcase's front & bookcase's side \\
                television receiver's base & television receiver's buttons & television receiver's frame & television receiver's keys & television receiver's screen \\
                television receiver's speaker & glass's base & glass's bowl & glass's opening & glass's stem \\
                pool table's bed & pool table's leg & pool table's pocket & airplane's door & airplane's fuselage \\
                airplane's landing gear & airplane's propeller & airplane's stabilizer & airplane's turbine engine & airplane's wing \\
                minibike's license plate & minibike's mirror & minibike's seat & minibike's wheel & washer's button panel \\
                washer's door & washer's front & washer's side & traffic light's housing & traffic light's pole \\
                light's aperture & light's canopy & light's diffusor & light's highlight & light's light source \\
                light's shade &  &  &  &  \\
            \cmidrule(l){1-5}
            \multicolumn{5}{c}{\textcolor{violet}{Novel Categories (58)}} \\
            \cmidrule(l){1-5}
                ottoman's back & ottoman's leg & ottoman's seat & swivel chair's back & swivel chair's base \\
                swivel chair's seat & swivel chair's wheel & fan's blade & fan's canopy & fan's tube \\
                stool's leg & stool's seat & desk's apron & desk's door & desk's drawer \\
                desk's leg & desk's shelf & desk's top & bus's bumper & bus's door \\
                bus's headlight & bus's license plate & bus's logo & bus's mirror & bus's wheel \\
                bus's window & bus's wiper & oven's button panel & oven's door & oven's drawer \\
                oven's top & kitchen island's door & kitchen island's drawer & kitchen island's front & kitchen island's side \\
                kitchen island's top & van's bumper & van's door & van's headlight & van's license plate \\
                van's logo & van's mirror & van's taillight & van's wheel & van's window \\
                van's wiper & truck's bumper & truck's door & truck's headlight & truck's license plate \\
                truck's logo & truck's mirror & truck's wheel & truck's window & bench's arm \\
                bench's back & bench's leg & bench's seat &  &  \\
            \bottomrule
            \end{tabular}
    }
    \caption{List of object-specific classes in \textbf{ADE20K-Part-234}.}
    \vspace{-1.0em}
    \label{suppl_tab:ade20k-part-234}
\end{small}
\end{table}

\subsection{PartImageNet}
\label{suppl_sec:partimagenet}

PartImageNet~\cite{he2022partimagenet_PartImageNet} is a dataset adapted from ImageNet~\cite{deng2009imagenet}, comprising approximately 24,000 images across 158 categories, each annotated with detailed part information. These categories are grouped into 11 superclasses, based on the hierarchical taxonomy provided by WordNet~\cite{miller1995wordnet}. 
In cross-dataset evaluation settings, PartImageNet categories are not pre-divided into base and novel classes. Therefore, for zero-shot evaluation, we select 40 representative object classes from the dataset, which we further split into 25 base object classes and 15 novel object classes, just as in PartCLIPSeg \cite{PartCLIPSeg2024}. Each superclass contains corresponding part classes, allowing us to construct part categories by associating these object classes with their respective part classes within each superclass. This selective splitting ensures that the novel classes remain unseen during training, thereby providing a robust evaluation of zero-shot capabilities. Detailed information about the base and novel classes, along with their corresponding superclasses, is presented in \Cref{suppl_tab:partimagenet_seg}.

\begin{table}[ht]
    \vspace{1.0em}
    \centering
    \begin{small}
        \resizebox{\linewidth}{!} {
           \begin{tabular}{llll}
                \toprule
                Superclass & \textcolor{teal}{Base Object Categories (25)} & \textcolor{violet}{Novel Object Categories (15)} & Part Classes \\
                \midrule
                Quadruped & tiger, giant panda, leopard, gazelle & ice bear, impala, golden retriever & Head, Body, Foot, Tail \\
                Snake & green mamba & Indian cobra & Head, Body \\
                Reptile & green lizard, Komodo dragon, tree frog & box turtle, American alligator & Head, Body, Foot, Tail \\
                Boat & yawl, pirate & schooner & Body, Sail \\
                Fish & barracouta, goldfish, killer whale & tench & Head, Body, Fin, Tail \\
                Bird & albatross, goose & bald eagle & Head, Body, Wing, Foot, Tail \\
                Car & garbage truck, minibus, ambulance & jeep, school bus & Body, Tier, Side Mirror \\
                Bicycle & mountain bike, moped & motor scooter & Body, Head, Seat, Tier \\
                Biped & gorilla, orangutan & chimpanzee & Head, Body, Hand, Foot, Tail \\
                Bottle & beer bottle, water bottle & wine bottle & Mouth, Body \\
                Aeroplane & warplane & airliner & Head, Body, Engine, Wing, Tail \\
                \bottomrule
            \end{tabular}
        }
    \caption{List of selected object classes and their corresponding part classes per superclass from \textbf{PartImageNet} \cite{he2022partimagenet_PartImageNet}. Object categories are categorized into base and novel object classes, with part classes assigned to each respective superclass.}
    \vspace{-1.0em}
    \label{suppl_tab:partimagenet_seg}
    \end{small}
\end{table}

\subsection{PartImageNet (OOD)}

PartImageNet~\cite{he2022partimagenet_PartImageNet} offers an alternative split designed for few-shot learning, which ensures non-overlapping classes across the training and validation sets. This few-shot split includes 109 base object classes in the training set and 19 novel object classes in the validation set. Detailed information about the base and novel classes, along with their corresponding superclasses, is presented in \Cref{suppl_tab:partimagenet_ood}.

\begin{table*}[ht]
    \vspace{1.0em}
    \centering
    \begin{small}
        \resizebox{\linewidth}{!} {
           \begin{tabular}{l l l}
                \toprule
                Superclass & \multicolumn{1}{l}{\textcolor{teal}{Base Object Categories (109)}} & \multicolumn{1}{l}{\textcolor{violet}{Novel Object Categories (19)}}\\
                \midrule
                Quadruped & impala, Egyptian cat, warthog, otter, Tibetan terrier & golden retriever, cougar, ice bear, mink, Saint Bernard \\
                & timber wolf, polecat, water buffalo, ox, redbone &  \\
                & English springer, tiger, American black bear, leopard, hartebeest &  \\
                & vizsla, Brittany spaniel, giant panda, Boston bull, ram &  \\
                & cairn, Arabian camel, fox squirrel, Eskimo dog, Irish water spaniel &  \\
                & Saluki, Walker hound, cheetah, gazelle, soft-coated wheaten terrier &  \\
                & bighorn, brown bear, chow, weasel &  \\
                
                Snake & night snake, boa constrictor, green mamba, thunder snake, green snake & Indian cobra \\
                & hognose snake, sidewinder, horned viper, diamondback, rock python &  \\
                & garter snake, vine snake &  \\
                
                Reptile & Gila monster, common newt, green lizard, bullfrog, American alligator & whiptail, alligator lizard \\
                & leatherback turtle, spotted salamander, box turtle, tailed frog, African chameleon &  \\
                & Komodo dragon, agama, frilled lizard, loggerhead &  \\
                
                Boat & yawl, pirate & trimaran \\
                
                Fish & goldfish, coho, tench, anemone fish, killer whale & barracouta, great white shark \\
                
                Bird & albatross, spoonbill, black stork, dowitcher, American egret & little blue heron, bald eagle \\
                & goose, ruddy turnstone, bee eater, kite &  \\
                
                Car & garbage truck, minibus, ambulance, snowplow, golfcart & beach wagon, cab \\
                & police van, minivan, convertible, limousine, recreational vehicle &  \\
                & go-kart, tractor, school bus, racer &  \\
                
                Bicycle & motor scooter, tricycle, mountain bike & unicycle \\
                
                Biped & gorilla, gibbon, guenon, macaque, patas & siamang, marmoset \\
                & howler monkey, chimpanzee, proboscis monkey, spider monkey, baboon &  \\
                & colobus, capuchin &  \\
                
                Bottle & beer bottle, pop bottle, pill bottle & water bottle \\
                
                Aeroplane & airliner &  \\

                \bottomrule
            \end{tabular}
        }
    \caption{
        \textbf{PartImageNet (OOD).}
        List of selected object classes and their corresponding part classes per superclass from PartImageNet (OOD) \cite{he2022partimagenet_PartImageNet}. Object categories are divided into base and novel object classes. Detailed associations of part classes with their respective superclasses are provided in~\Cref{suppl_tab:partimagenet_seg}.}
    \vspace{-1.0em}
    \label{suppl_tab:partimagenet_ood}
    \end{small}
\end{table*}

\newpage \thispagestyle{empty} \mbox{}
\newpage \thispagestyle{empty} \mbox{}
\newpage \thispagestyle{empty} \mbox{}
\newpage \thispagestyle{empty} \mbox{}
\newpage \thispagestyle{empty} \mbox{}
\newpage \thispagestyle{empty} \mbox{}
\newpage \thispagestyle{empty} \mbox{}
\newpage \thispagestyle{empty} \mbox{}
\newpage

\end{document}